\documentclass[a4paper,fleqn]{cas-dc}

\usepackage[numbers]{natbib}
\usepackage{subcaption}
\usepackage{xcolor}

\def\tsc#1{\csdef{#1}{\textsc{\lowercase{#1}}\xspace}}
\tsc{WGM}
\tsc{QE}
\tsc{EP}
\tsc{PMS}
\tsc{BEC}
\tsc{DE}

\begin{document}
\let\WriteBookmarks\relax
\def\floatpagepagefraction{1}
\def\textpagefraction{.001}
\shorttitle{Assessing invariance to affine transformations in image quality metrics}
\shortauthors{N. Alabau-Bosque et~al.}


\title [mode = title]{Assessing invariance to affine transformations in image quality metrics}


\author[1,2]{Nuria Alabau-Bsoque}[type=editor,
                        auid=000,bioid=1,
                        orcid=0009-0002-8092-2066]
\cormark[1]
\ead{nuria.alabau@uv.es}
\credit{Conceptualization of this study, Data Curation, Formulation, Investigation, Methodology, Software, Visualization, Writing - Original draft preparation}

\author[1]{Paula Daudén-Oliver}
\credit{Data Curation, Investigation, Software, Visualization}

\author[1]{Jorge Vila-Tomás}
\credit{Investigation, Software, Writing - review and editing}

\author[1]{Valero Laparra}
\credit{Conceptualization of this study, Methodology, Supervision, Writing - Original draft preparation}

\author[1]{Jesús Malo}
\credit{Conceptualization of this study, Formulation, Methodology, Supervision,  Writing - Original draft preparation}


\affiliation[1]{organization={Image Processing Lab, Universitat de València},
                addressline={Carrer del Catedrátic José Beltrán Martinez}, 
                city={Paterna},
                postcode={46980},
                country={Spain}}
\affiliation[2]{organization={ValgrAI: Valencian Grad. School Research Network of AI, València, , Spain},
                addressline={Camí de Vera, s/n, Edificio 3Q}, 
                postcode={46022}, 
                city={Valencia},
                country={Spain}}




\begin{abstract}
Subjective image quality metrics are usually evaluated according to the correlation with human opinion in databases with distortions that may appear in digital media.
However, these oversee affine transformations which may represent better the changes in the images actually happening in natural conditions. Humans can be particularly invariant to these natural transformations, as opposed to the digital ones.\\
In this work, we propose a methodology to evaluate \emph{any image quality metric} by assessing their invariance to affine transformations, specifically: rotation, translation, scaling, and changes in spectral illumination.
Here, \emph{invariance} refers to the fact that certain distances should be neglected 
if their values are below a threshold. 
This is what we call \emph{invisibility threshold} of a metric. 
Our methodology consists of two elements: (1)~the determination of a visibility threshold in a subjective representation common to every metric, and (2)~a transduction from the distance values of the metric and this common representation. This common representation is based on subjective ratings of readily available image quality databases. We determine the threshold in such common representation (the first element) using accurate psychophysics. 
Then, the transduction (the second element) can be trivially fitted for \emph{any metric}: with the provided threshold extension of the method to any metric is straightforward. 
%
We test our methodology with some well-established metrics and find that none of them show human-like invisibility thresholds. 
\\
This means that tuning the models exclusively to predict the visibility of generic distortions may disregard other properties of human vision as for instance invariances or invisibility thresholds.
The data and code is publicly available to test other metrics\tnotemark[1].
\end{abstract}

\tnotetext[1]{https://github.com/Rietta5/InvarianceTestIQA}.

\begin{graphicalabstract}
\includegraphics[width=1\linewidth]{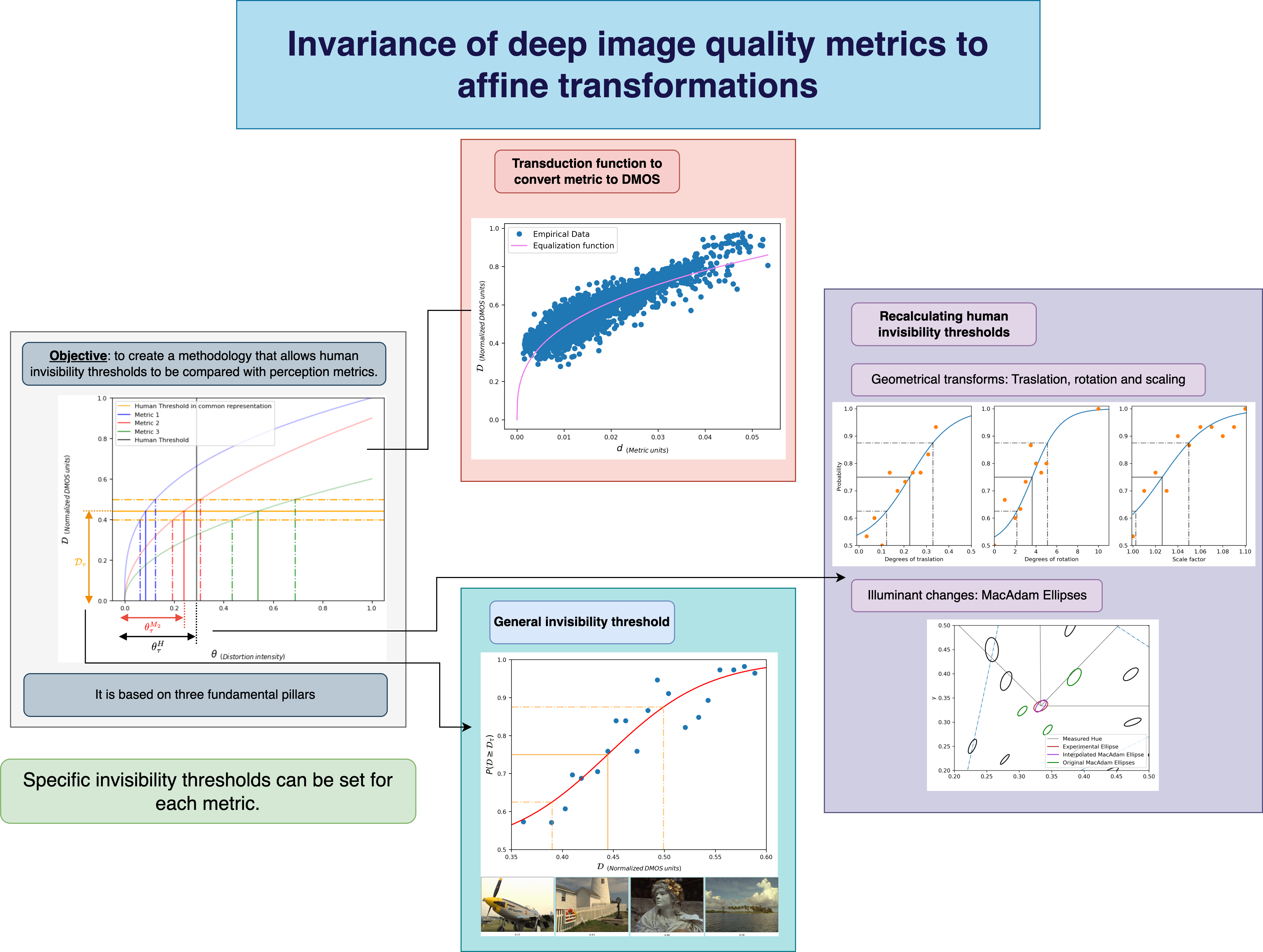}
\end{graphicalabstract}

\begin{highlights}
\item We propose a methodology to assign invisibility thresholds for any perceptual metric.
\item We measure the human invisibility threshold to image distortions in the Difference Mean Opinion Score representation by fitting a psychometric function. 
\item We expand classical literature on human thresholds to affine transformations in physical units using natural stimuli as opposed to synthetic stimuli.
\item The proposed methodology takes the above-mentioned thresholds and straightforward fitting of transduction functions to derive thresholds for any metric in physical units.
\item Alignment between metrics and humans is assessed in two ways: (1) a quantitative comparison of thresholds, and (2) a qualitative comparison of sensitivities.
\end{highlights}


\maketitle

\section{Introduction}


Artificial models used to predict subjective image quality, usually referred to as \emph{perceptual metrics}, are often used as measures to optimize other image processing tasks~\cite{IQA_compression, IQA_denoising,ICV,ICV2} so, assessing their performance is critical. Usually, these models are evaluated (or even tuned) according to their ability to correlate with human opinion in databases including a wide range of generic distortions~\cite{Perceptnet,parecido1,LPIPS}. In fundamental terms, this means trying to predict the visibility of generic distortions. However, focus on generic distortions may be a problem to take into account other relevant phenomena~\cite{Martinez19}, and human vision may also be described in terms of perceptual constancies or invariances~\cite{chrom_adap,contrast_const,Kelly79,color_const,motion_compens,scale_const}.

In this regard, considering the structural analysis of distortions and high-level interpretation of human vision, people have suggested that humans are mainly invariant to transformations which do not change the structure of the scenes~\cite{Wang&Sim}. Affine transformations (rotations, translations, scalings, and changes in spectral illumination) are examples of distortions that do not change the structure of the scene. Therefore, humans should be relatively tolerant to them, and the corresponding models to assess image similarity should be invariant to these transformations too.

In fact, the spirit of the influential SSIM was focused on measuring changes of structure \cite{SSIM} to achieve invariances to irrelevant transformations~\cite{MSE_Loveit}. Moreover, Wang and Simoncelli \cite{Wang&Sim} decomposed generic distortions into \emph{structural} and \emph{non-structural} components so that the part not affecting the structure (e.g. affine transformations) could be processed, and weighted, differently. 
On the other hand, metrics with bio-inspired, explainable architecture~\cite{Perceptnet,Laparra16_NLPD,LaparraJOSA17,Laparra10,Malo97,Mantiuk11,Martinez19,Martinez18} work in multi-scale / multi-orientation representations where invariances can be introduced by means of appropriate poolings in the representations \cite{Bruna13_scatter}. This sort of poolings are thought to happen in the visual brain leading to invariances and texture metamers~\cite{FreemanSimoncelli11_NN}. 

However, current state-of-the-art deep-architectures for image quality~\cite{LPIPS} do not address the invariance problem in any way, while examples that try to apply the SSIM concept in deep-nets~\cite{DISTS} do not use invariances in simple or explicit ways. 
As a result, the analysis of invariance in deep image quality metrics remains an open question.

In this work, we propose a methodology to evaluate the metrics from the point of view of human detection thresholds: by assessing the invariance of the metrics to image transformations that are irrelevant (or invisible) to human observers. 
In particular, we compare the ability of metrics to be invariant to affine transformations in the same way as humans are. 
For example, the classical literature on visual thresholds determines the intensity of certain affine transformations which is invisible for humans \cite{umbral_traslacion,MacAdam,umbral_rotacion,umbral_escala}.
The sizes of invisibility thresholds are related to the more general concept of \emph{invariance} to transformations. 
By definition, transformations whose intensity is below the threshold are invisible to the observer. Then one can say that, in this region, the observer is invariant to the transform. 
Here, we propose a methodology that allows us to measure the invisibility thresholds for any given metric. 
The proposed methodology has two elements: (1)~the determination of a visibility threshold in a subjective representation common to every metric, and (2)~a transduction function from the raw distance values of the metric and this common representation. This common representation is based on subjective ratings of readily available image quality databases. Here we present accurate psychophysical measurements of the threshold in such subjective representation (the first element), and then, the transduction function (second element) can be trivially fitted for \emph{any metric} using available image quality databases. In this way the proposed methodology is general and applicable to any metric \emph{any metric}.

Then, we can (1)~assess if the thresholds for the metrics are comparable to those found for human observers, and (2)~assess if the sensitivities of a metric for different distortions follow the same order as the human sensitivities (for instance, humans are more sensible to rotation changes than to illumination changes).
This proposed evaluation of invariance to distortions is a necessary complement to the conventional evaluation of the visibility of distortions because, as we will see, none of the studied metrics presents human-like thresholds nor sensitivities for all the transformations considered.

The structure of the paper is as follows: Section~2 describes the proposed methodology to assess the human-like behavior of metric invariances, consisting of comparing the detection thresholds for humans and metrics. 
This proposal depends on several concepts (transduction functions, psychophysical thresholds for humans, theoretical thresholds for metrics... ) that will be detailed in this section too. Section~3 describes the experimental setting, and Section~4 considers the results on the thresholds and the sensitivities. The discussion and conclusions are presented in Section~5 and Section~6, respectively.

\section{Proposed Methodology: Thresholds and Sensitivities}

Given an original image, $i$, it can be distorted through certain transform whose intensity depends on a parameter~$\theta$, $i' = T_\theta(i)$. 
Image quality metrics are models that try to reproduce the subjective sensation of distance between the original image and the distorted image, $d(i,i')$.
Human observers are unable to distinguish between $i$ and $i'$, i.e. they are invariant to transform~$T_\theta$, if its intensity is below a (human) threshold, $\theta_{\tau}^{\textrm{H}}$.

\begin{figure}[!h]
			\centering
   \includegraphics[width=1\linewidth]{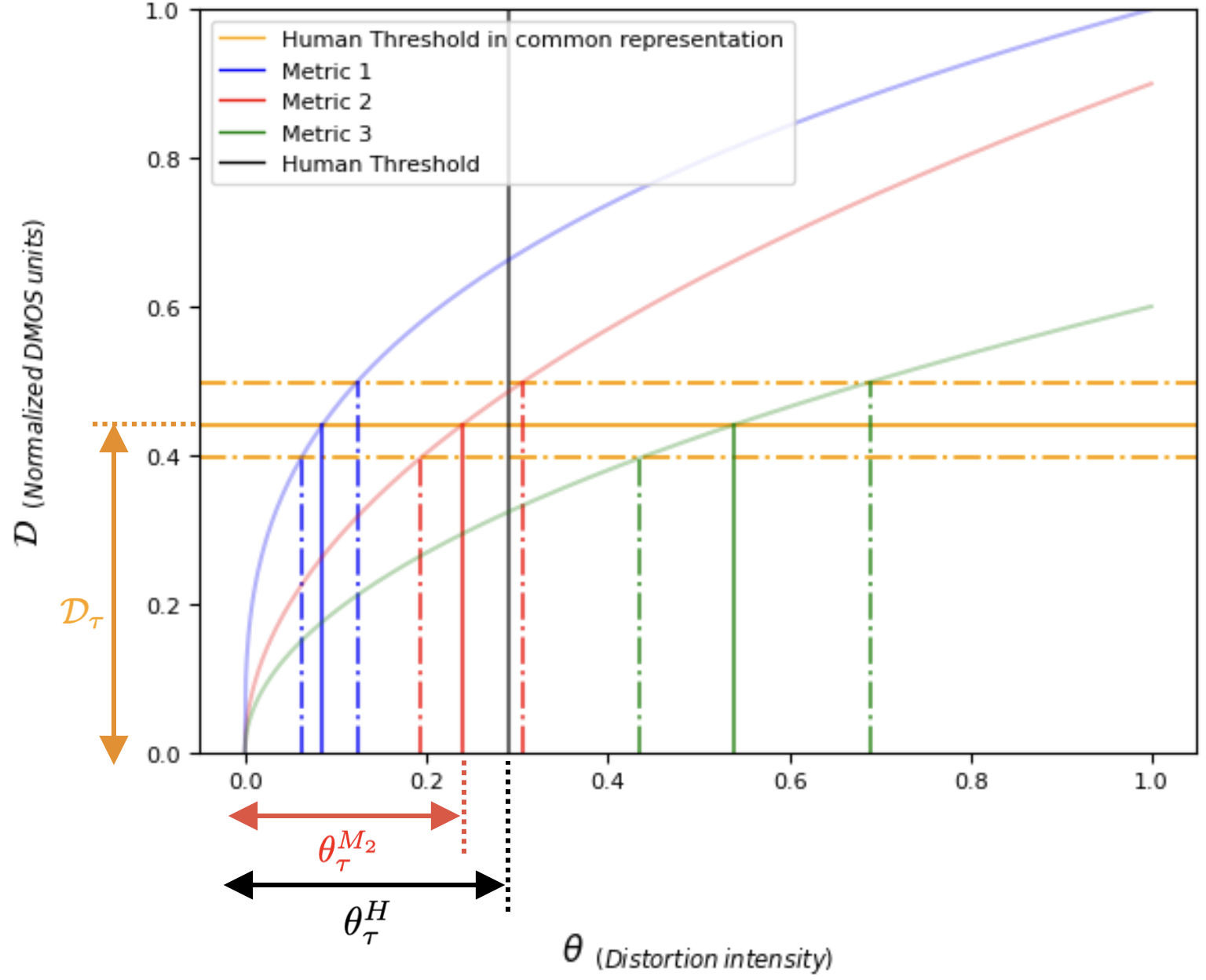}	
   \caption{\textbf{Illustration of the methodology: comparison of metric vs human thresholds}. For each distortion, parametrized with $\theta$, we evaluate three different metrics (continuous curves in blue, red, and green) in common units of subjective distance ($\mathcal{D}$, the normalized Differential Mean Opinion Score). 
   Given $N$ competing metrics, these curves are the \emph{transduction functions} that map the physical description of the distortion intensity, $\theta$ into a common scale of subjective distance, $\mathcal{D}_M = g_M(\theta)$, here with $M=1,\ldots,3$. 
   The yellow lines show the human \emph{threshold in the internal distance representation}, $\mathcal{D}_\tau$, experimentally measured in this work with a psychometric function. 
   We assign the \emph{metric thresholds} from the generic human threshold using the inverse of the transduction functions for the different metrics: $\theta_\tau^{M} = g_M^{-1}(\mathcal{D}_\tau)$. 
   Given the fact that $\mathcal{D}_\tau$ has an experimental uncertainty, the inverse gives us an interval - dotted lines- on the x-axis where the human threshold should fall for each metric and this particular distortion. 
   For any number $N$ of metrics we propose to compare the \emph{metric thresholds}, $\theta_\tau^{M}$, for $M=1,\ldots,N$, with the actual \emph{human threshold} in intensity units, $\theta_\tau^H$. 
   In this case, $\theta_\tau^H$ (the black line), falls within the red interval. Therefore, the metric~2 is the one with the human-like behavior.}
		\label{fig:Fig1}
   \end{figure}

While invariance thresholds in human observers, $\theta_{\tau}^{\textrm{H}}$, are easy to understand and measure~\cite{umbral_traslacion,MacAdam,umbral_rotacion,umbral_escala}, they are not obvious to define in artificial models of perceptual distance. 
The reason is simple: for the usual image quality models, any non-zero image distortion leads to non-zero variations in the distance. 
In this situation, where artificial distances are real-valued functions, one should define a value, \emph{the threshold distance}, $\mathcal{D}_\tau$, below which the difference between the images could be disregarded (or taken as zero).
Once this \emph{threshold distance} is available (eventually in a common scale for all metrics) one can translate this threshold to the axis that measures the distortion intensity, $\theta$, and hence obtain the threshold of the metric, $\theta_\tau^{M}$, in the same units that has been measured for humans, $\theta_\tau^{H}$.  
See an illustration of this concept in Figure~\ref{fig:Fig1}.

The diagram in Figure~\ref{fig:Fig1} displays the relation between 
the physical description of the intensity of certain distortions in abscissas 
(the parameter $\theta$, e.g. the angle in a rotation transform), and the common internal description of the perceived distance in ordinates (the distance $\mathcal{D}$, e.g. the normalized Differential Mean Opinion Score units explained below).
This relation between the physical description of the intensity and common perceptual distance is what we call \emph{transduction function}. 
The example displays three transduction functions, $\mathcal{D}_M = g_M(\theta)$, with $M=1,2,3$, in blue, red, and green,  for the three corresponding metrics.
Transduction functions are monotonic: the application of a progressively increasing distortion along the abscissas leads to monotonic increments in distances in ordinates. 
The \emph{threshold distance}, $\mathcal{D}_\tau$, in the internal representation is plotted in orange in Figure~\ref{fig:Fig1}. This threshold may be uncertain (as represented by the central value, solid line, and the quartile limits, in dashed lines), but the empirical transduction functions can be used to put this internal threshold back in the axis that describes the distortion intensity: 
$\theta_\tau^{M} = g_M^{-1}(\mathcal{D}_\tau)$.
In this way, one can check if the actual invisibility threshold measured for humans (black line) is consistent with the threshold interval deduced for each metric. In our illustration, the metric~2 (red lines) is the only one compatible with human behavior.

Using the above, on the one hand, we propose to evaluate the alignment between the metric models and human observers by comparing $\theta_\tau^{M}$ versus $\theta_\tau^{H}$.
This is a strict comparison that only depends on the
experimental uncertainty of the thresholds. 
On the other hand, we can define an alternative (less strict) comparison by considering the order among the sensitivities for the different distortions in humans and models. In particular, one can define the sensitivity of a metric for a distortion as the variation of the transduction function in terms of the energy (or Mean Squared Error) of the distortion introduced by the transform~\cite{Hepburn22}. The human sensitivity in detection is classically defined to be proportional the inverse of the energy required to see the distortion~\cite{Campbell68}. 


The proposed comparisons of metric vs human thresholds and metric vs human sensitivities are general as long as one can address the following issues:

\begin{enumerate}
     \item[\textbf{(a)}] 
     \textbf{The transduction function} (red, green, and blue curves in the illustration).
     For the $M$-th metric the relation between the physical description of the image transform and a common metric-independent distance domain has two components: 

     \begin{enumerate}

     \item[(a.1)] The (non-scaled) \emph{response function} can be empirically computed by generating images distorted (transformed) with different intensities,~$\theta$, and using the metric expression to compute the corresponding distances from the original image. This leads to the distances
     $d_M(\theta) = d_M(i,i') = d_M(i,T_\theta(i))$.

     \item[(a.2)] A metric \emph{equalization function} transforming the previous (non-scaled) distance values into the common scale of the internal distance representation (what we called normalized DMOS units in the illustration). 
     Here we propose to use auxiliary empirical data (e.g. certain subjectively rated databases) 
     to scale the range of the different metrics: $\mathcal{D}_M = f_M(d_M)$. This makes the different $\mathcal{D}_M$ comparable.

     \end{enumerate}

     Then, the final (scaled) transduction is the composition of response and equalization: $\mathcal{D}(\theta) = g_M(\theta) = f_M(d_M(\theta))$.\\

     \item[\textbf{(b)}] \textbf{The human thresholds}, 
     that can be defined in different domains:
     
     \begin{enumerate}

     \item[(b.1)] \emph{The human threshold in the common internal representation}, $\mathcal{D}_\tau$, orange line in the illustration of Fig~\ref{fig:Fig1}.
     In principle, this value is unknown.      
     Here we propose a standard measurement of this threshold through a psychometric function~\cite{Kingdom13} using distorted images of the selected subjectively rated database.

     \item[(b.2)] \emph{The human threshold in the input physical representation}, $\theta_\tau^{H}$, black line in the illustration of Fig~\ref{fig:Fig1}. In this work, we explore two options: 
     (i)~take the values from the classical literature, which in general uses substantially different stimuli (synthetic as opposed to natural), and 
     (ii)~re-determine the thresholds in humans by using comparable natural stimuli and a separate psychometric function for each distortion.

     \end{enumerate}     
     
     \item[\textbf{(c)}] \textbf{The metric threshold}, which can be expressed in intuitive physical units, or in the own units of the metric:

     \begin{enumerate}

     \item[(c.1)] In physical units $\theta_\tau^{M}$. Here, the blue, red, and green points in the x-axis of the illustration of Fig~\ref{fig:Fig1}, computed as $\theta_\tau^{M} = g_M^{-1}(\mathcal{D}_\tau)$.
     These units are useful to compare with the equivalent human values.

     \item[(c.2)] In the units of the metric, $d_\tau^{M} = f_M^{-1}(\mathcal{D}_\tau)$.
     This value is particularly interesting, as it indicates a variation in the distance of each metric for which humans see no difference between the original and the distorted image. Distortions leading to distances below this value are invisible to humans and hence should be neglected.

     \end{enumerate}   
          
     \item[\textbf(d)] \textbf{The sensitivities of humans and metrics for the different distortions}. The sensitivity for a small distortion is usually defined as the inverse of the energy required to be above the invisibility threshold~\cite{Campbell68}. 
     In the case of metrics, this general definition reduces to the derivative of the transduction function with regard to the energy of the distortion~\cite{Hepburn22}.

\end{enumerate}

Below, we elaborate on each of these factors in turn.

\subsection{Transduction: response and equalization}

The response function of a metric, $d_M$, to a certain image transform, $T_\theta$, is just the average over a set of images, $\{i^s\}_{s=1}^S$, of the distances for the distorted images for different transform intensities:

\begin{equation}
d_M(\theta) = \frac{1}{S} \sum_{s=1}^{S} d_M(i^s,T_\theta(i^s))
\label{eq:response}
\end{equation}

\noindent which can be empirically computed from controlled distortions of the images of the dataset. 
Of course, when considering $M$ different metrics, their response functions, $d_M$, are not given in a common scale.\\

In this work, we propose to use auxiliary empirical data to determine a common distance scale, $\mathcal{D}$, for any metric. 
In particular, subjectively rated image quality databases 
(e.g. TID~\cite{TID13}) consist of pairs of original and distorted images, 
$\{(i^p,{i^p}')\}_{p=1}^P$, 
with associated subjective scores, the so-called Differential Mean Opinion Scores, 
$\{\textrm{DMOS}^p\}_{p=1}^{P}$.
The databases contain a wide set of generic distortions of different intensities thus ranging from invisible distortions to highly noticeable distortions. In this setting, DMOS values in the database can be normalized to be in the [0,1] range. 
In this way, the lower and higher values of the normalized DMOS represent an invisible distortion and the biggest subjective distortion in the database respectively. 
Therefore, if the range of distortions in the database is wide, the variations induced by $T_\theta$ will be within the limits of the normalized DMOS, and hence, it can be used to set the common distance scale $\mathcal{D} = \textrm{norm DMOS} \in [0,1]$.

\begin{figure}[!h]
			\centering
			\includegraphics[width=0.8\linewidth]{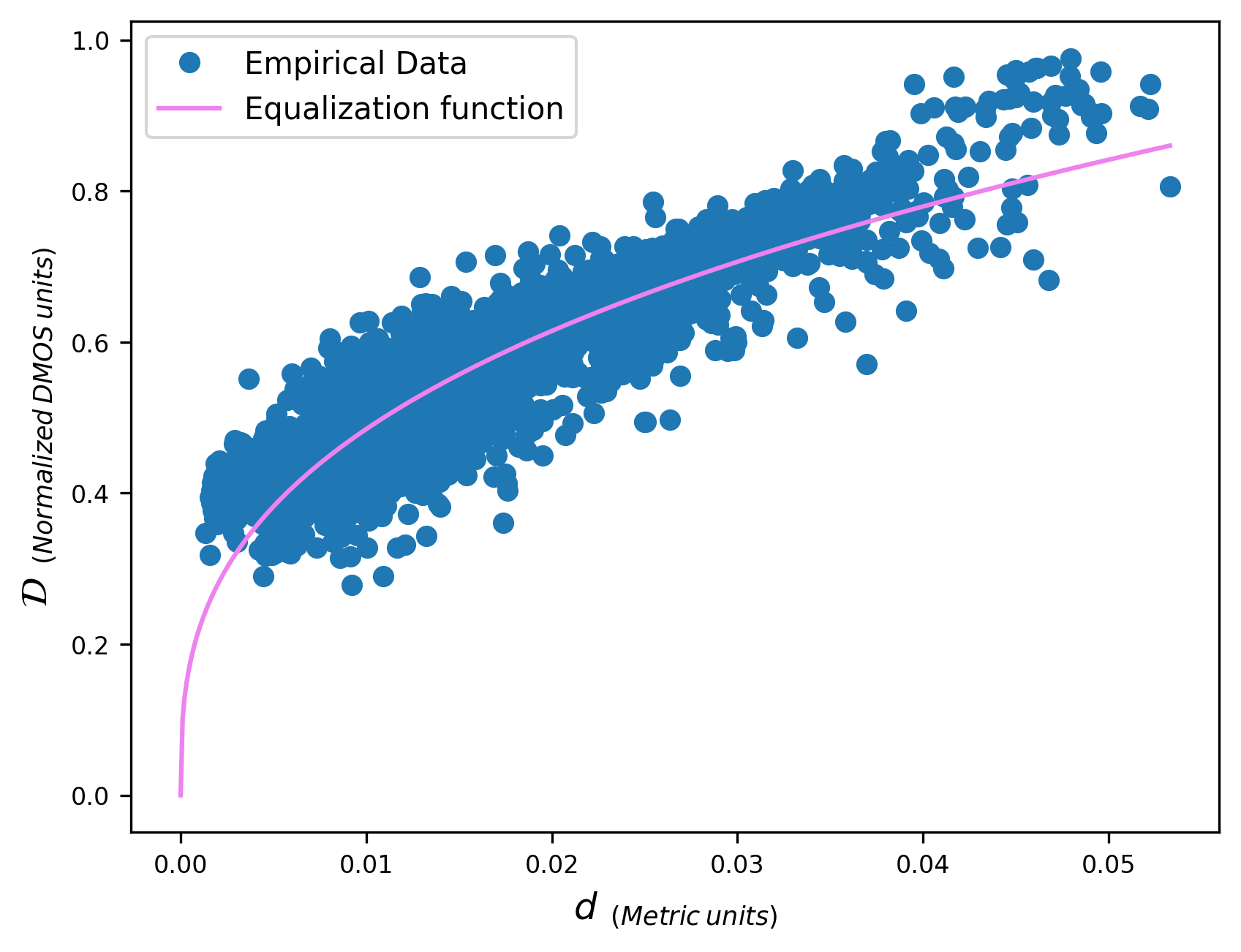}	
   \caption{Example of equalization function fitted on TID13 data, for the metric LPIPS, following Eq.~\ref{eq:equaliz}.}
		\label{fig:ex_equalization}
   \end{figure}

Using the normalized DMOS values and the image pairs of a large subjectively rated database one can fit an equalization function, $f_M$, for each metric to transform the non-scaled response $D_M$ into the common scale $\mathcal{D}$:
%
%
\begin{equation}
   \mathcal{D} = f_M( d_M(i,i') ) = a_M\cdot  d_M(i,i')\,^{b_M}     \label{eq:equaliz}
\end{equation}
\noindent where an exponential function with $a_n > 0$ and $ 0 < b_n \neq 1$ is chosen because of the nature of the DMOS, which changes rapidly for low distortion intensities, low values of $\theta$, and saturates for bigger distortions, big values of $\theta$. An example of an equalization function is shown in Figure~\ref{fig:ex_equalization} as an example.

In summary, (a.1) an application of the distance metric to a controlled set of transformed images, and (a.2) a fit of an equalization function to transform the arbitrary scale of the distance to a common scale given by the set of distortions in a wide subjectively rated database, gives us the transduction function of the metric made of Eqs.~\ref{eq:response} and~\ref{eq:equaliz}:
\begin{equation}
    \mathcal{D} = g_n(\theta) = f_n \circ d_n (\theta)  
\end{equation}

In our work, as means of showcasing our proposed methodology, we compute these transduction functions for:
\begin{itemize}
\item Six distance metrics: four state-of-the-art deep-learning metrics~\cite{PIM,DISTS,Perceptnet,LPIPS}, and two convenient references (the Euclidean metric, RMSE, and the classical SSIM~\cite{SSIM}).
\item Four affine transformations: translation, rotation, scale, and change of spectral illumination.
\item Four datasets to transform the images and compute the response functions: MNIST~\cite{MNIST}, CIFAR10~\cite{cifar10}, ImageNet~\cite{Imagenet}, and TID2013~\cite{TID13}.
\item One subjectively rated dataset, TID2013~\cite{TID13}, to define the equalization functions to the common internal distance representation.
\end{itemize}

\subsection{Human thresholds}

In this section, we measure two kinds of human thresholds. First, we use distorted images from a subjectively rated database to measure from the internal common representation, $\mathcal{D}_\tau$. Then, we measure the thresholds in physichal units, $\theta_\tau$, using natural stimuli.

\subsubsection{Human thresholds in the common internal representation}
\label{subsec:human_internal}

$\mathcal{D}_\tau$ is the distance in normalized DMOS units from which humans can't tell the difference between $i$ and $i'$ for low distortion intensities $\theta$. For that, we need a database with ratings and opinions of observers, for which we can assign a threshold value of invisibility. In this case, we use the TID13 database~\cite{TID13}.

The value of $\mathcal{D}_\tau$ could be roughly estimated by visual inspection of the images presented in Appendix~A (example in Figure \ref{fig:ExAppendixA}): images with low values of normalized DMOS (below 0.3) cannot be discriminated from the original, while images with big normalized DMOS (above 0.6) are clearly distinct from the original.
However, the more accurate estimation of such a threshold is obtained from the psychometric function in a constant stimulus experiment~\cite{Kingdom13} applied to other fields within computer vision and image processing~\cite{pf_segmentation, pf_computervision, pf_artifacts, pf_naturalscenes, pf_NR}.

\begin{figure*}[!h]
			\centering
			\includegraphics[width=0.9\linewidth]{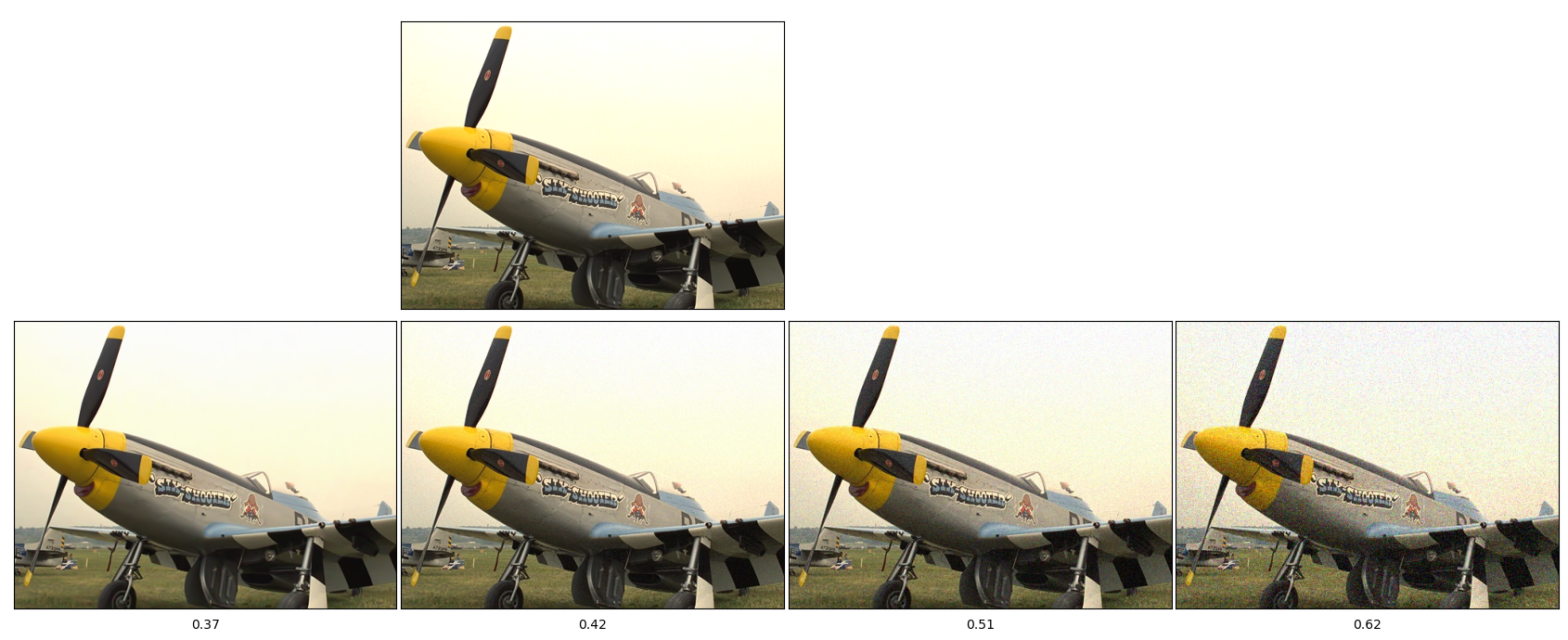}	
   \caption{Example image from Appendix A, showing the relationship between the DMOS of the images and their visibility. An image with a low DMOS (around 0.3) is indistinguishable from the original, while a high DMOS (around 0.6) is clearly different.}
		\label{fig:ExAppendixA}
   \end{figure*}

In this kind of experiments, given a set of distorted images one computes the probability that an observer sees some distorted image as different from the original, i.e. $P(\mathcal{D} \geq \mathcal{D}_\tau)$. This is done by using the two-alternative forced choice paradigm (2AFC): by randomly presenting the observer each distorted image together with the original so that the observer is forced to choose the distorted one. This is repeated $R$ times, so the probability, $P(\mathcal{D} \geq \mathcal{D}_\tau)$, is given by the number of correct responses over $R$. Note that if $\mathcal{D} \ll \mathcal{D}_\tau$ the observer will not see the difference and the probability of correct answer will be 0.5. On the other extreme, if $\mathcal{D} \gg \mathcal{D}_\tau$ the answer will be obvious for the observer, and the probability of a correct answer will be~1. As a result, the threshold can be defined as the point where $P(\mathcal{D} \geq \mathcal{D}_\tau) = 0.75$.

We look for the optimal threshold, $\mathcal{D}_\tau$, and slope, $k$, that better fit the experimental data using the following sigmoid~\cite{Kingdom13}:
\begin{equation}
    p(x) = \frac{1}{2} + \frac{1}{2(1 + e^{-k (\mathcal{D}-\mathcal{D}_\tau)}) }
    \label{eq:psycho}
\end{equation}
Note that this expression enforces that for extremely distorted images $\lim\limits_{\mathcal{D}\to\infty} p(\mathcal{D} \geq \mathcal{D}_\tau) = 1$, and that the probability at the threshold, $\mathcal{D} = \mathcal{D}_\tau$ is 0.75, as required. 

\begin{figure}[!h]
			\centering
			\includegraphics[width=0.9\linewidth]{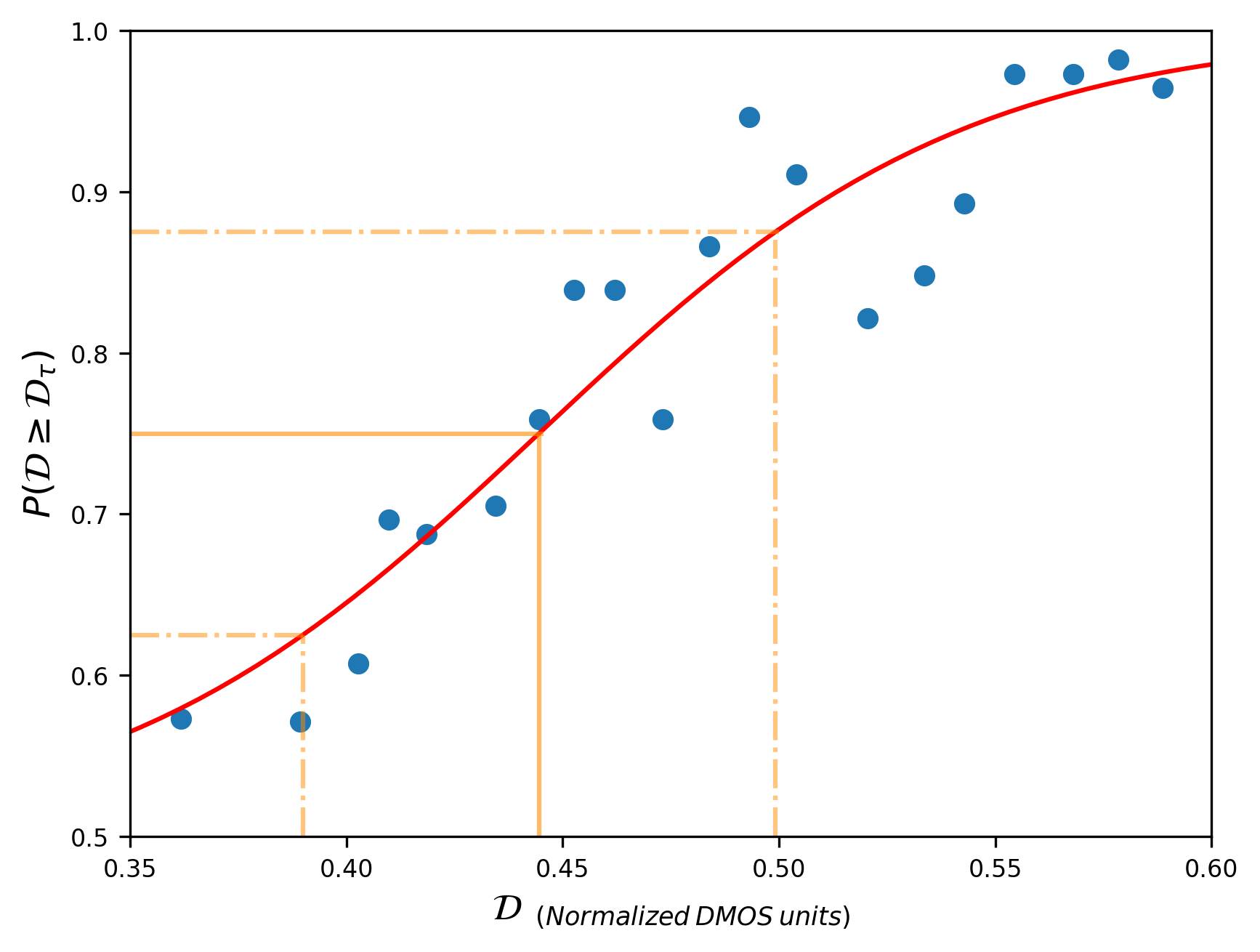}	
   \caption{Human threshold in the internal distance representation, $\mathcal{D}_\tau$, from the psychometric function of a constant stimuli experiment (see details in the text). The detection threshold is $\mathcal{D}_\tau = 0.44 \pm 0.05$ in normalized DMOS units.}
		\label{fig:psycometric}
   \end{figure}

In our experiment, the psychometric function has been evaluated in 20 different values of $\mathcal{D}$, repeated 15 times for 11 observers, resulting in $20 \times 15 \times 11 = 3300$ forced choices in total. The fitted psychometric function is shown in Figure~\ref{fig:psycometric} and shows that the value of the threshold for humans in the common internal distance representation is: $$\mathcal{D}_\tau = 0.44 \pm 0.05$$ where the corresponding quartiles give the uncertainty. This threshold corresponds to the orange lines in Figure~\ref{fig:Fig1}. 


\subsubsection{Human threshold in physical units}
\label{subsec:Humans}


In principle, one could take the thresholds
$\theta_\tau^H$ from the classical literature~\cite{umbral_traslacion,MacAdam,umbral_rotacion,umbral_escala} but, considering the critical effect of the stimuli used in the experiments and the fact that classical thresholds were measured with synthetic stimuli, 
we decided to measure these thresholds for natural images to ensure consistency with the kind of images used by the metrics. We measure the thresholds according to the constant stimuli setting, as in Section \ref{subsec:human_internal}, to derive the classical psychometric function~\cite{Kingdom13}. 
In particular, images from ImageNet have been used to find the invisibility thresholds in a more realistic context.

For each transform, this experimental procedure requires the variation of the stimuli in the parameter space $\theta$, which is trivially one-dimensional for 
scale, translation, rotation, and intrinsically two-dimensional for chromatic shifts (despite they coming from arbitrary variations of a higher dimensional object such as the spectrum of the illuminant~\cite{Stiles00}).

\begin{figure}[!h]
			\centering
	\includegraphics[width=1\linewidth]{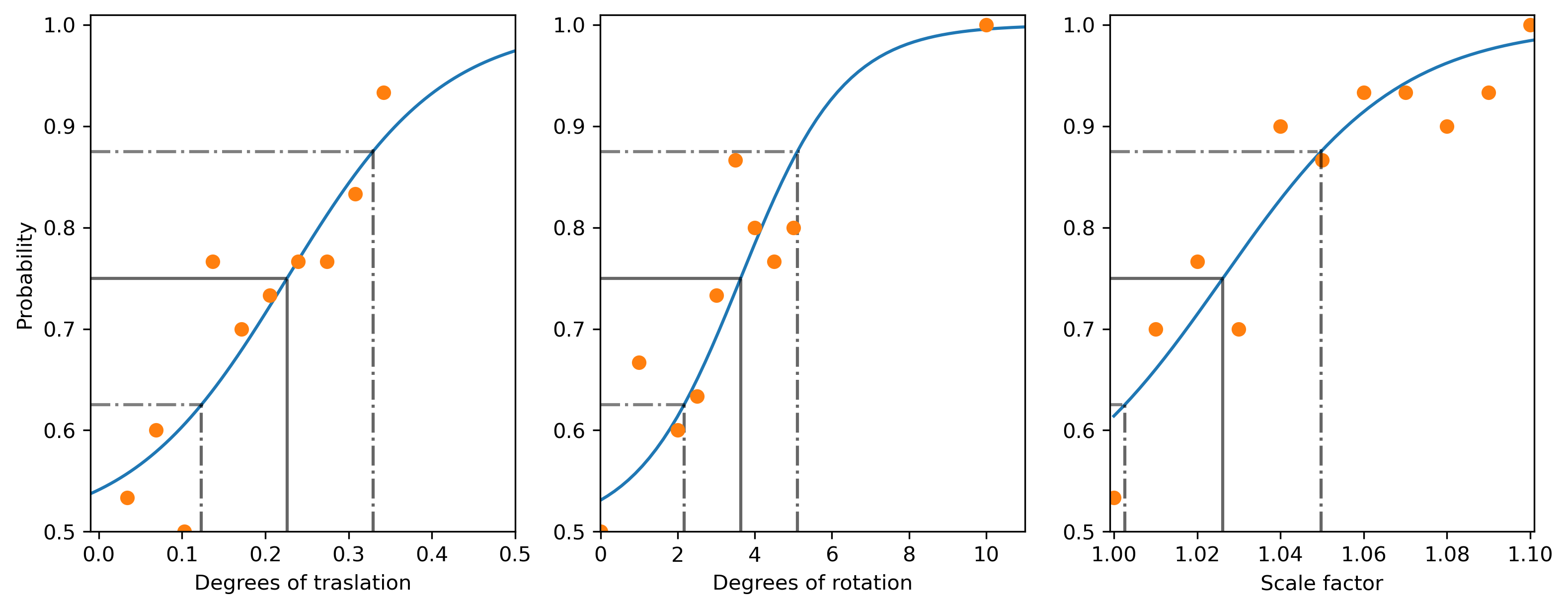}
            \label{fig:leaky}
		        \caption{Experimental (human) psychometric functions measured with natural images. According to them, the human thresholds for the geometric affine transformations are 0.23 $\pm$ 0.10 degrees for translation, 3.6 $\pm 1.5$ degrees for rotation, and 1.03 $\pm 0.02$ for scale.}
		\label{fig:psycometric_afin}
  \end{figure}

On the one hand, regarding the geometrical transforms with one-dimensional parameter, the results of the thresholds obtained from the psychometric curves are shown in Figure~\ref{fig:psycometric_afin}.

On the other hand, now we address the results of the color thresholds which are two-dimensional in nature.
As a convenient summary, here we describe the chromatic content of natural images using 
a single two-dimensional point in the 
so-called chromatic diagram~\cite{Stiles00}.
In particular, we will use the chromatic coordinates of the spectrum of the illuminant assumed to be the illumination source of the scene. Moreover, we will describe the color-shifts introduced to check the invariance of metrics by the new chromatic coordinates of the new spectrum applied to the scene (as affine transform on the image vector as in  Eq.~\ref{eq:affine2} in  Section~\ref{subsec:trasnformations}). Here we assume that the natural images in the databases are, in principle, illuminated by a generic \emph{white} spectrum. Therefore, we describe the chromaticity of these images using the location of that generic white spectrum in the CIE xy diagram, i.e. in the 2d-point $(1/3,1/3)$~\cite{Stiles00}. 

\begin{figure}[!h]
			\centering
			\includegraphics[width=1\linewidth]{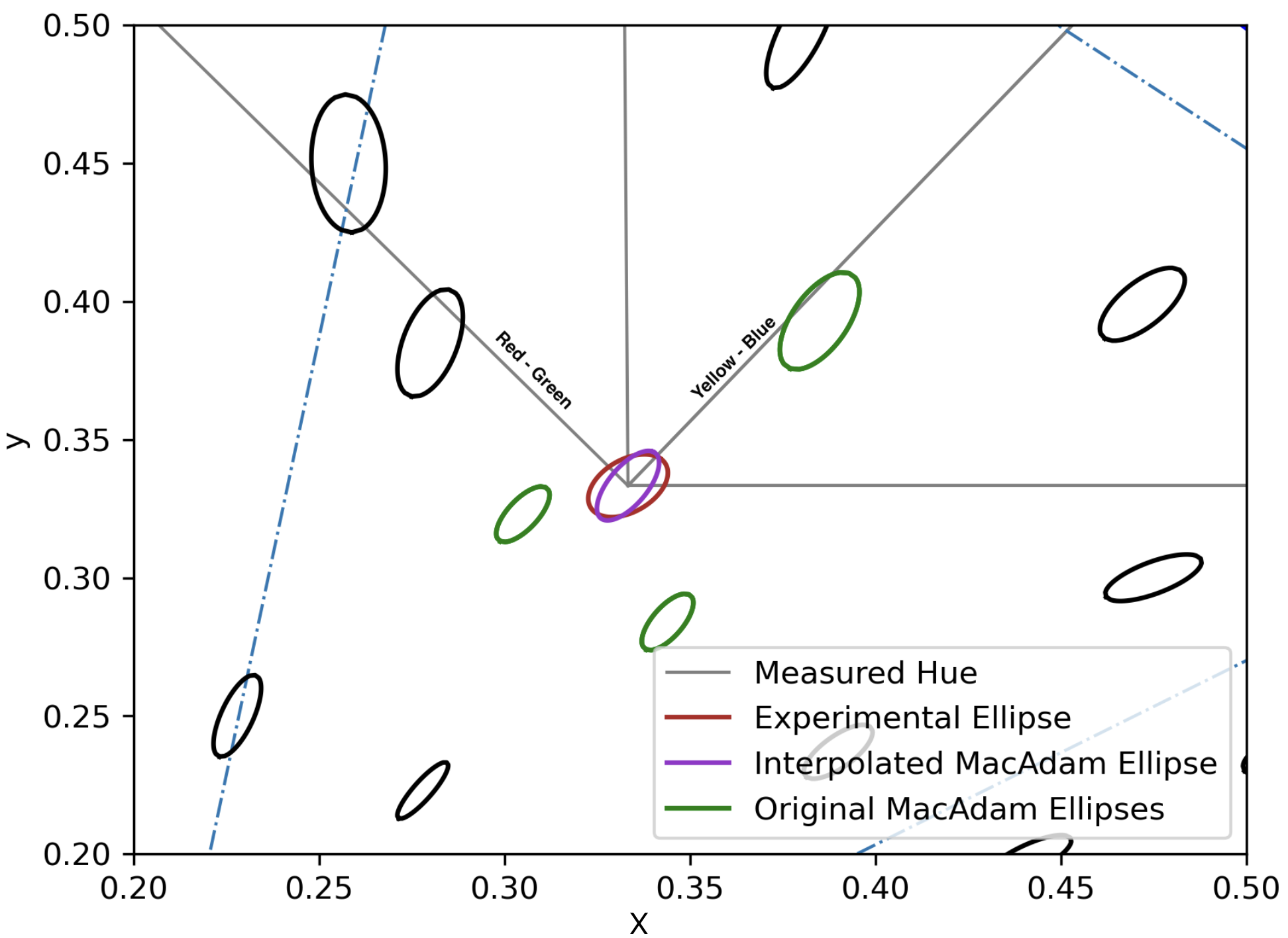}	
   \caption{New fitted ellipses. In purple, the new ellipse centered on the absolute white interpolated from the closest ellipses (in green). In red, the experimental ellipse fitted with our own measured data. The highlighted axes are the interpolated ellipse's major and minor axes and two intermediate hues.}
		\label{fig:fit_elipses}
\end{figure}
\begin{figure}
  		\begin{subfigure}{0.5\textwidth}
			\centering
			\includegraphics[width=0.9\linewidth]{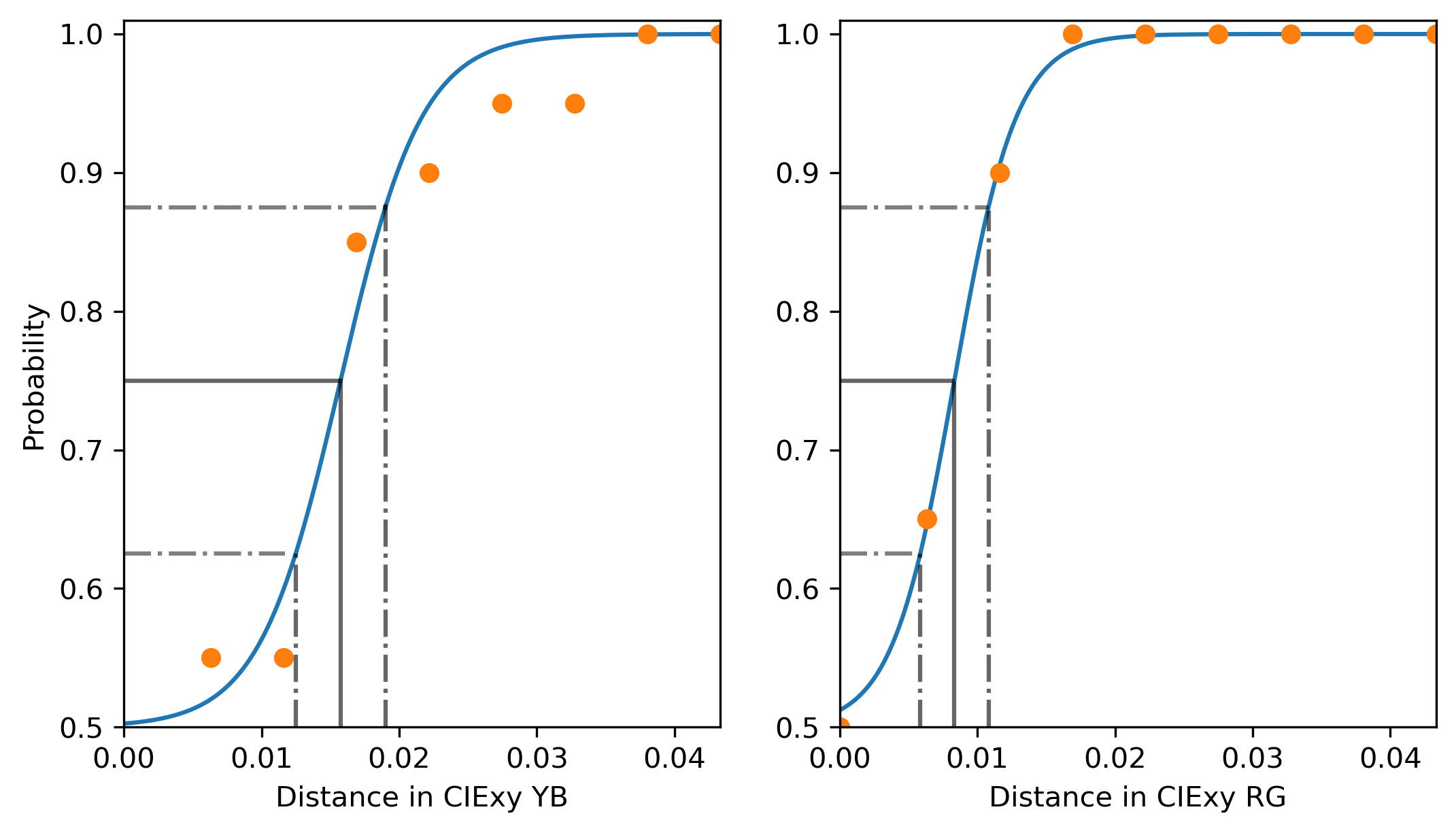}			
		\end{subfigure}

            \begin{subfigure}{0.5\textwidth}
			\centering
			\includegraphics[width=0.9\linewidth]{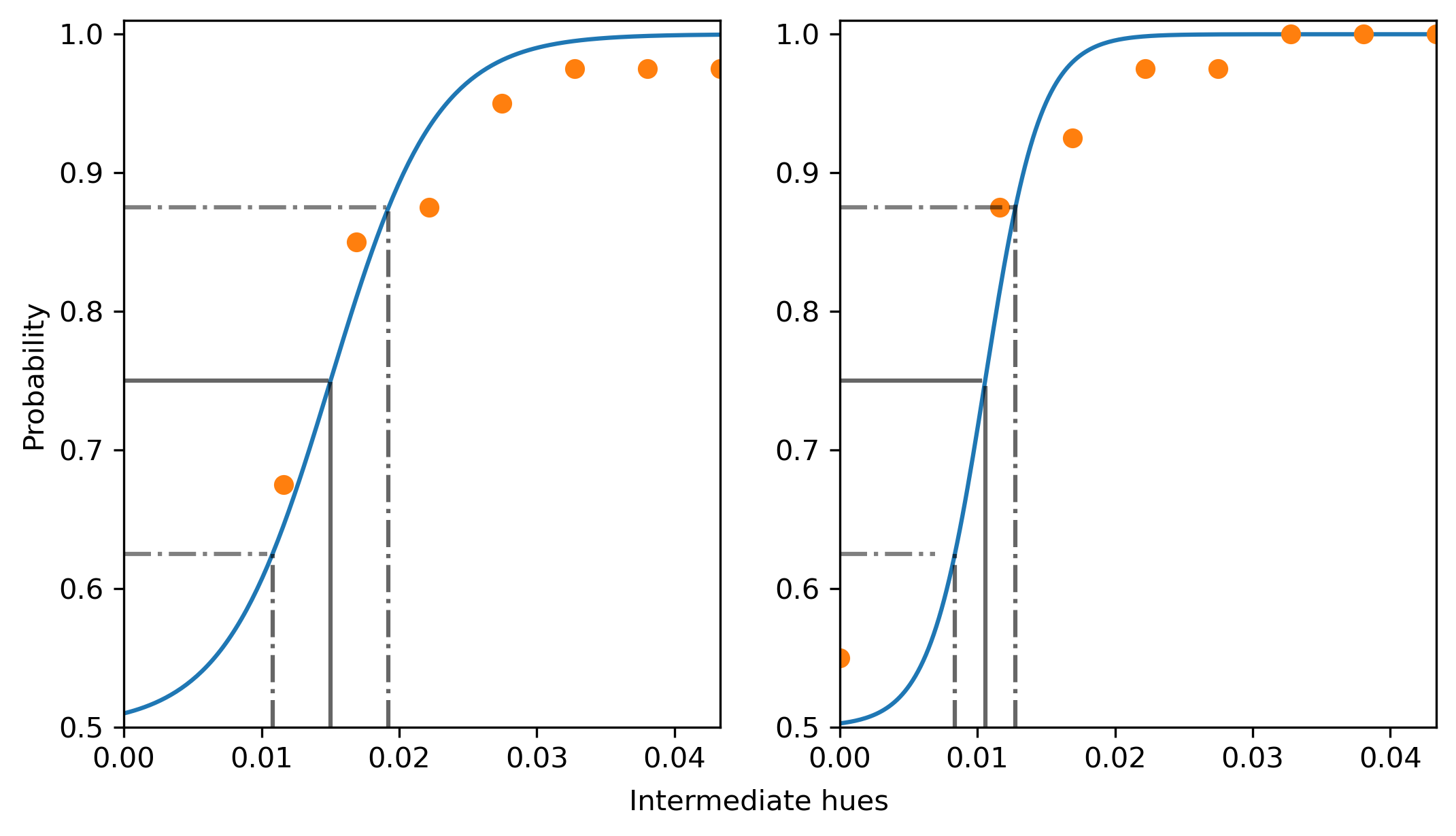}	
		\end{subfigure}
		        \caption{Experimental (human) psychometric functions measured with natural stimuli. For the illuminant changes, we obtained 4 curves where the YB direction is the major axis, and the RB direction is the minor axis of the ellipse. In addition, two intermediate shades are also shown for more precise adjustment.}
		\label{fig:psycometric_color}
	\end{figure}

The two-dimensional nature of chromatic shifts implies that invisibility regions are not intervals, but two-dimensional shapes (e.g. discrimination ellipses~\cite{MacAdam}).
In order to determine the shape of such discrimination region around a certain central chromatic location, e.g. the aforementioned white point, one could measure one-dimensional intervals moving the images in different chromatic directions away from the selected
center. For instance, Figure~\ref{fig:fit_elipses}
represents the white point $(1/3,1/3)$, generic illuminant, and four linear paths indicating color shifts away from this center (radial gray lines).
Measurements along radial lines can be done with the constant-stimulus paradigm mentioned above. In our case, given the ellipsoidal shape of the classical chromatic discrimination thresholds~\cite{MacAdam} (also shown in Figure~\ref{fig:fit_elipses}), measuring only in four directions is enough for a trustable fit of an ellipse~\cite{ajuste_elipses}.
The fitted psychometric functions for these four chromatic directions are shown in Figure~\ref{fig:psycometric_color}.
The thresholds in departure in chromatic coordinates from these psychometric functions determine four two-dimensional experimental thresholds that allow to fit the ellipse in pink in Figure~\ref{fig:fit_elipses}.

Obtaining and assessing the result found (pink ellipse in Figure~\ref{fig:fit_elipses}), and also the assessment of ellipses obtained from the metrics, depends on the measuring RMSE distances between ellipses.
This error is relevant in the fitting process~\cite{ajuste_elipses}, where theoretical points are compared with the experimental points.
Additionally, assessment implies the comparison of the predicted ellipsoid with the corresponding region that could be found in classical literature~\cite{MacAdam,Barbur10}.
As the selected reasonable white was not explicitly measured in the classical literature, we fitted a new ellipse at this point considering the closest existing ellipses (in green in Figure~\ref{fig:fit_elipses}).

Interestingly the ellipse we got by assessing the detection thresholds with color-shifted natural images is quite consistent with previous results that used synthetic stimuli and different experimental techniques~\cite{MacAdam,Barbur10}.
On the one hand, note the similarity in size and orientation between the pink and the purple ellipses.
Note that the ellipse is wider in the yellow-blue (YB) direction than in the red-green (RG) direction.
The authors in~\cite{Barbur10} reported that their threshold method with synthetic textured stimuli led to ellipses $\times 4.5$ than MacAdam ellipses~\cite{MacAdam} measured with the variance of color matching paradigms. 
As Fig~\ref{fig:fit_elipses} shows the MacAdam ellipses with a $\times 5$ visualization factor~\cite{Stiles00}, our experimentally fitted pink ellipse obtained with thresholds from natural images is very consistent with~\cite{Barbur10}.

Table~\ref{tab:tabla_umbrales} summarizes the results with synthetic (obtained from classical literature) and real stimuli (experimentally measured in this work).

\begin{table}[!h]
\centering
\caption{Human thresholds for the different related transformations. In the second column, we have the thresholds according to the classical literature. In the third column, we have the thresholds obtained in experiments with humans carried out in the laboratory.}
\resizebox{8.5cm}{!} {
\begin{tabular}{|c|c|c|}
\cline{1-3} 
 \textbf{Affine Transformations} & \multicolumn{1}{c|}{\textbf{Synthetic stimuli}}&\textbf{Natural stimuli}\\ \hline
\multicolumn{1}{|c|}{Translation} &\begin{tabular}[c]{@{}l@{}}0.024 degrees \cite{umbral_traslacion}\\ 0.12 degrees \cite{Baldwin16}\end{tabular}&  0.23 $\pm$ 0.10 degrees\\ \hline
\multicolumn{1}{|c|}{Rotation} & 3 degrees \cite{umbral_rotacion} & 3.6 $\pm$ 1.5 degrees\\ \hline
\multicolumn{1}{|c|}{Scale} & 1.03 scale factor \cite{umbral_escala} & 1.03 $\pm$ 0.02 scale factor\\ \hline
\multicolumn{1}{|c|}{Color discrimination} & MacAdam ellipse \cite{MacAdam} & Experimental ellipse (Fig. \ref{fig:fit_elipses})\\\hline

\end{tabular}
}
\label{tab:tabla_umbrales}
\end{table}

From Table \ref{tab:tabla_umbrales} we see that the rotation and scale thresholds are practically equivalent, but this is not the case for the translation threshold. This can be attributed to the fact their experiments with synthetic stimuli~\cite{umbral_traslacion} can be considered as being close to a hyperacuity setting, (which returns very low thresholds)~\cite{Strasburger18}. 
Our experiments employ natural images and do not enforce the hyperacuity setting. This results in a considerably higher detection threshold for this transformation. Recent work we could consider is the thresholds measured with gabor filters \cite{Baldwin16}. In this case, the human rotation threshold was considered to be 2.7 degrees of rotation and the human translation threshold the equivalent of 0.12 degrees of translation. These measurements are quite consistent with ours. Taking into account that the methodology of obtaining the threshold is the same as in our case, the difference in the thresholds is given exclusively by the type of stimulus used.
For this reason, the detection threshold for translation from other literature will not be considered in the results because of differences in the procedure for obtaining it and its interpretation in the case of classical literature; and for the type of stimulus used in the recent literature.


\subsection{Metric thresholds in physical units}
\label{sec:metric_threshold}


 Following the methodology in Figure~\ref{fig:Fig1}, we can obtain the metric thresholds in physical units, $\theta_\tau^{M}$. For example, the rotation threshold is expressed in degrees. Through the function $g_M(\theta)$, specific to each metric, a value in physical units is assigned to the threshold in the common internal representation, computed as $\theta_\tau^{M} = g_M^{-1}(\mathcal{D}_\tau)$. These values can be compared numerically with the thresholds obtained for humans, as they are expressed in the same units because they were obtained with the psychometric functions, Section \ref{subsec:Humans}. Therefore, a metric can be said to have human behavior if the invisibility threshold for a certain affine transformation coincides, $\theta_\tau^{M} =  \theta_\tau^{H}$. In particular, since $\mathcal{D}_\tau$ has an associated uncertainty, $\theta_\tau^{M}$ also has a confidence interval. In this case, we will check whether the human threshold falls within the confidence interval of each metric.
 
This test is particularly demanding because it is strictly quantitative. It is to be expected, that the metrics will not be able to reproduce these thresholds. This is why an alternative, less demanding, test is proposed, reflecting the qualitative behavior of the metrics in the face of these distortions.

\subsection{Sensitivities of metrics and humans for the different distortions}
\label{sec:sensitivities}

Taking into account that the methodology proposed above is particularly demanding, we propose another test. Having calculated the thresholds for metrics and humans, we can define sensitivities for both and compare them. However, this comparison will be qualitative, not quantitative, so this test is less demanding for metrics.

The sensitivity for a small distortion is usually defined as the inverse of the energy required to be above the invisibility threshold, i.e, expressing the distortion $\theta_\tau^{H}$ in RMSE units. In the case of metrics, this general definition reduces to the derivative of the transduction function concerning the energy of the distortion, i.e. its slope~\cite{Hepburn22}. 

\begin{equation}
\label{eq:sens}
S = \frac{1}{|i-T_{\theta^{H,M}_\tau}(i)|_2}
\end{equation}

In this way, we can obtain the sensitivity of a metric for each affine transform, order them and check if the order matches the human ones. A real example of this can be found in Figure~\ref{fig:ejemplo_combinadas} (the rest of the figures can be found in Appendix~D), where the vertical lines represent the human thresholds expressed in terms of energy and the different curves describe the behavior of that metric for the various transformations, where both are calculated as the mean over the whole database. Given two curves, the most sensitive will be the one with a higher slope.

In Figure~\ref{fig:ejemplo_combinadas}, the human thresholds (vertical lines) indicate that the human sensitivity order is: scale, translation, rotation, RG, and YB illuminants, where a lower threshold implies a higher sensitivity. However, the metric shown returns the following order given by the slopes of the curves (slope is directly related to the sensitivity): scale, rotation, translation, YB, and RG illuminants. Even if the ordering doesn't strictly match, it is to be noted that this metric maintains more sensitivity to geometric than chromatic distortions. 

   \begin{figure}[!h]
			\centering
			\includegraphics[width=0.8\linewidth]{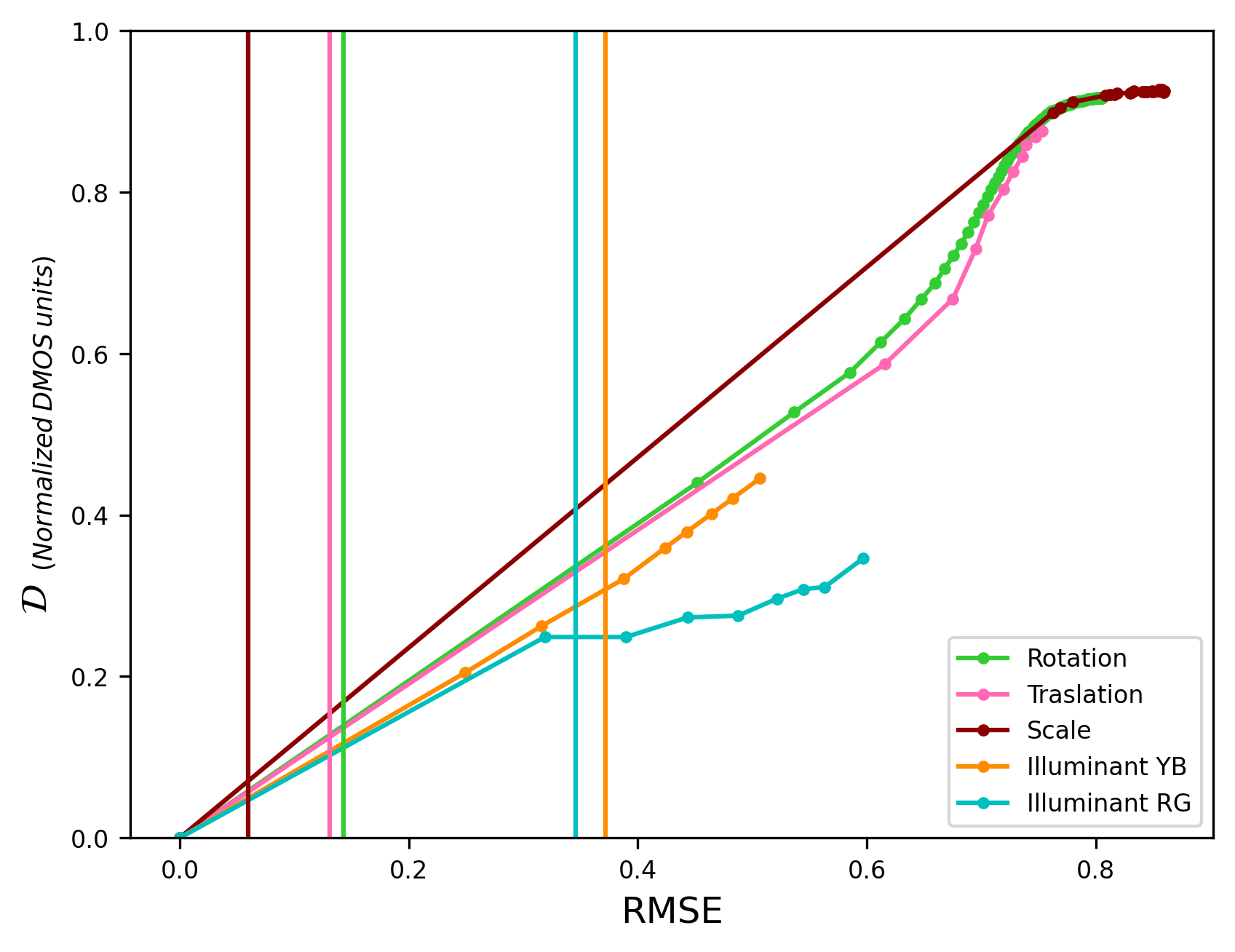}	
   \caption{Example of results comparing the sensitivities. Here, humans are more sensitive in the following order: scale, translation, rotation, illuminant RG, and YB. However, this metric, in this case, PerceptNet, returns the following order: scale, rotation, translation, illuminant YB, and RG.}
		\label{fig:ejemplo_combinadas}
   \end{figure}

\section{Experimental Settings}
In this section, we review the selected affine transformations, models, and databases that we will use in the experiments. Code available to test other metrics is available publicly\footnote{https://github.com/Rietta5/InvarianceTestIQA}.

\subsection{Affine Transformations}
\label{subsec:trasnformations}

An affine transformation or affine application (also called affinity) between two affine spaces is a transformation that satisfies:

\begin{equation}
    F: v \to Mv + b
    \label{eq:affine}
\end{equation}

Where $v$ can be any vector and the affine transformation is represented by a matrix $\mathbf{M}$ and a vector $\mathbf{b}$ satisfying the following properties: first, it maintains the collinearity (and coplanarity) relations between points and, second, it maintains the ratios between distances along a line.

Here, we apply affine transformations in two cases: domain (Equation~\ref{eq:affine1}) and image samples (Equation~\ref{eq:affine2}), i.e. modifications within the image vector:

\begin{equation}
    i'(x) = i(Mx + b)
    \label{eq:affine1}
\end{equation}

\begin{equation}
    i'(x) = Mi(x) + b
    \label{eq:affine2}
\end{equation}

Some examples of affine transformations that follow Equation~\ref{eq:affine1} are geometric contraction, expansion, dilation, reflection, rotation, or shear; and some examples following Equation~\ref{eq:affine2} are changes in contrast, luminance, and illuminant. In this work, we are going to focus specifically on \textbf{translation, rotation, scale, and illuminant changes}. For each tested affine transformation, the original images are modified in the following ways:
\begin{itemize}
    \item Translation: Displacements on the vertical and horizontal axis (and the combination between them) with an amplitude of 0.3 degrees of translation in each direction (left, right, up, and down). Given the symmetry in both displacement directions, we calculate the average and show only the displacements to the right in the graphs.
    \item Rotation: Rotations from -10 to 10 degrees, in steps of 0.1 degree. Again, given the symmetry in both displacement directions, only positive rotations will be shown in the graphs.
    \item Scale: Scale factors range from 0.1 to 2, but do not have a fixed step. Only scale factors that return images of even size are used. This ensures that only scales that do not force a translation are applied. Only scale factors bigger than 1 will be shown.
    \item Illuminant changes: We have desaturated the original images and modified the illuminant for 20 hues in 8 saturations, i.e. angle with respect to the $x$ axis, and distance with respect to the central white point, respectively. Following the distribution at the chromatic diagram as indicated in Figure~\ref{fig:hues}.

    \begin{center}
\begin{figure}[!h]
\centering
			\includegraphics[width=0.9\linewidth]{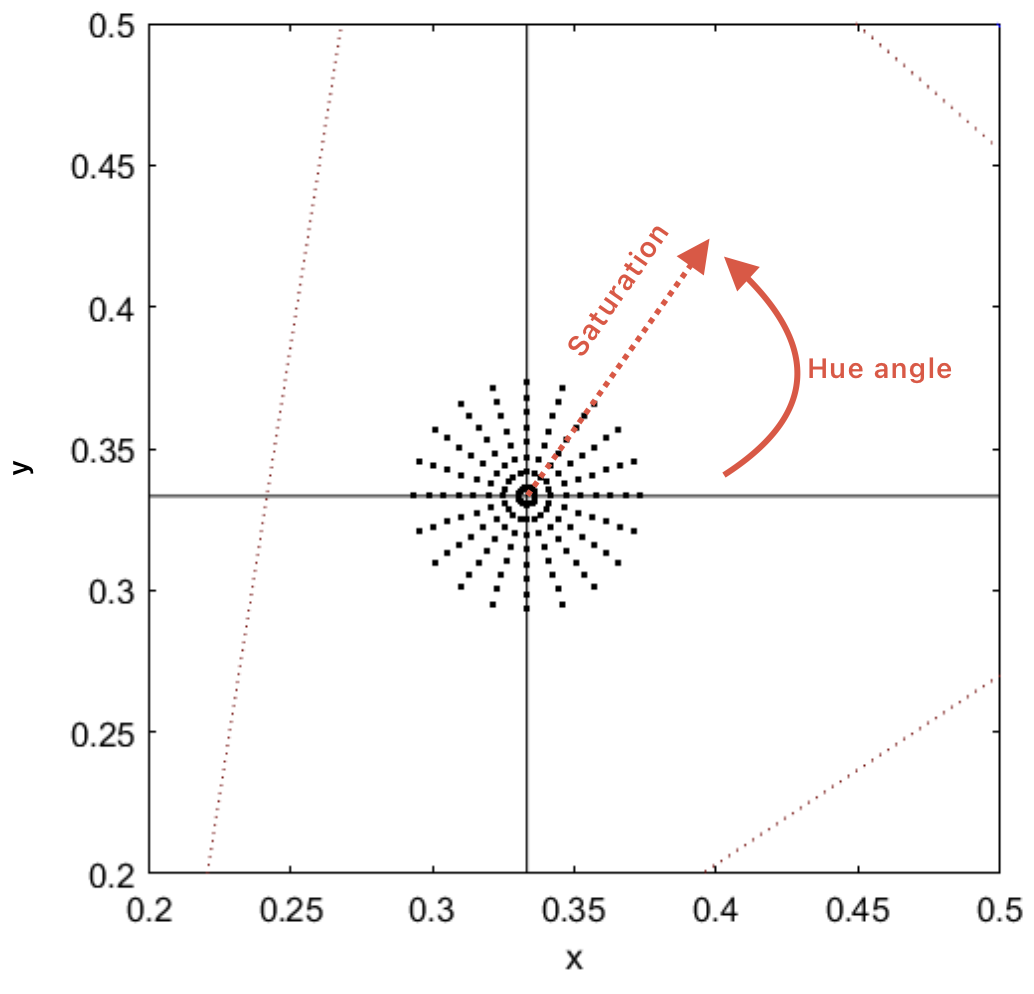}	
		
		\caption{The considered illuminants are organized as a function of hue and saturation, i.e. angle concerning the x-axis, and distance concerning the white point respectively. We consider 20 hues, i.e. 20 different angles, and 8 intensities for each hue.}
		\label{fig:hues}
	\end{figure}
\end{center}
    
\end{itemize}

\begin{center}
\begin{figure}[!h]
\centering
			\includegraphics[width=1\linewidth]{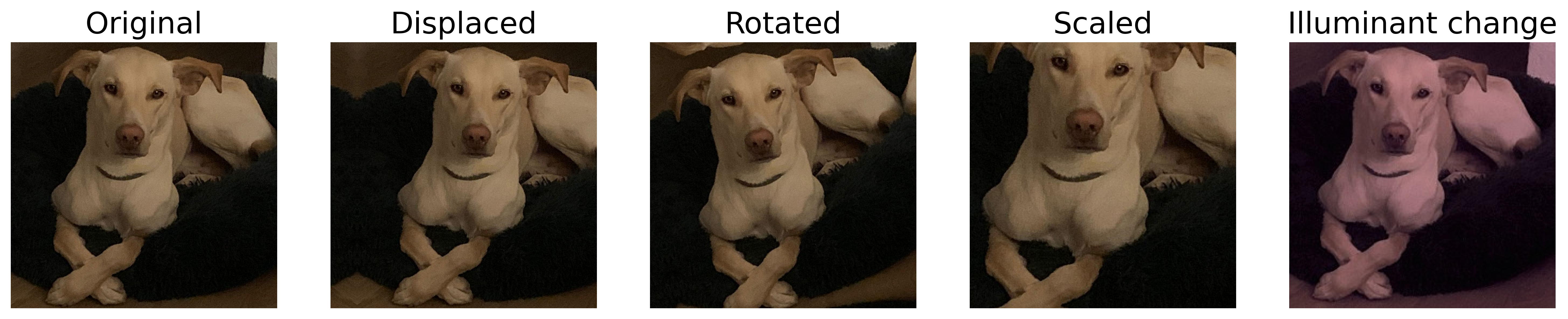}	
		
		\caption{Examples of affine transformations applied to the original images. From left to right it shows the original image, translation (100 pixels right), rotation (15 degrees), scale (factor of 1.3), and illuminant change (hue 17, purple, with maximum intensity).}
		\label{fig:distorsiones}
	\end{figure}
\end{center}

As a summary, examples of all the affine transformations emulated in this work can be seen in Figure~\ref{fig:distorsiones} and an example of all illuminant changes and different intensities in Figure~\ref{fig:cambio_iluminante}.

\subsection{Datasets}
\label{subsec:datasets}

In terms of dataset selection, we chose four different datasets covering a wide range of features. On the one hand, we have MNIST \cite{MNIST} for the black and white images. A simple set both to understand and to modify. On the other hand, for color images, we have selected CIFAR-10\cite{cifar10} for color images with low resolution, and ImageNet\cite{Imagenet} and TID2013\cite{TID13} for color images with high resolution.

\begin{center}
\begin{figure}[!h]
			\centering
			\includegraphics[width=0.8\linewidth]{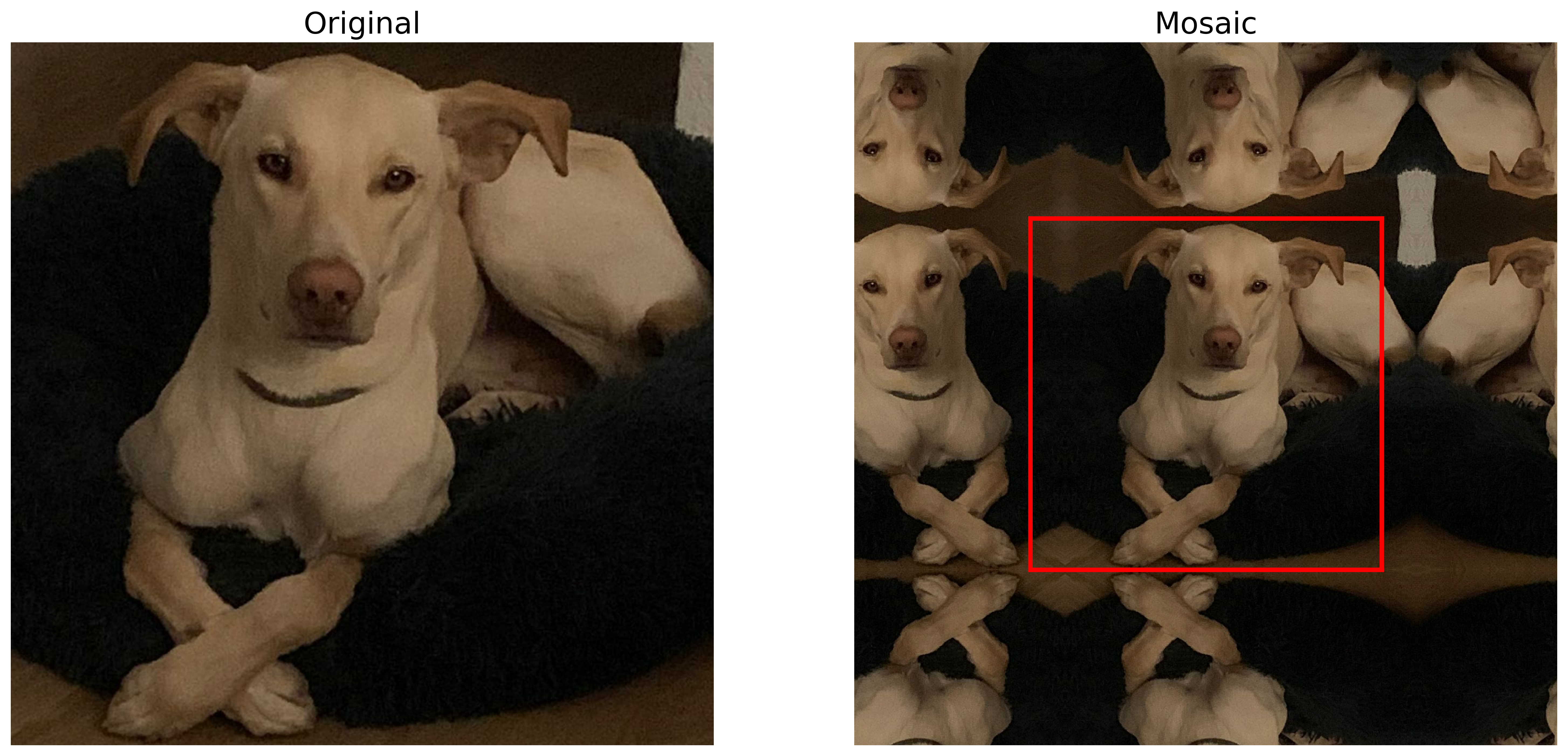}
   \caption{Example of an original image and with the mosaic effect to fill the image when affine transformations modify it out of the original dimensions. Once the mosaic is created, the affine transformation is applied and cropped, as indicated by the red frame, recovering the initial dimensions. This is an example of how the mosaic behaves if we want to move the image to the right. Further examples of the final results can be seen in Figure~\ref{fig:distorsiones}.}
		\label{fig:mosaico}
   \end{figure}

   \end{center}
\begin{figure*}[!h]
			\centering
			\includegraphics[width=1\linewidth]{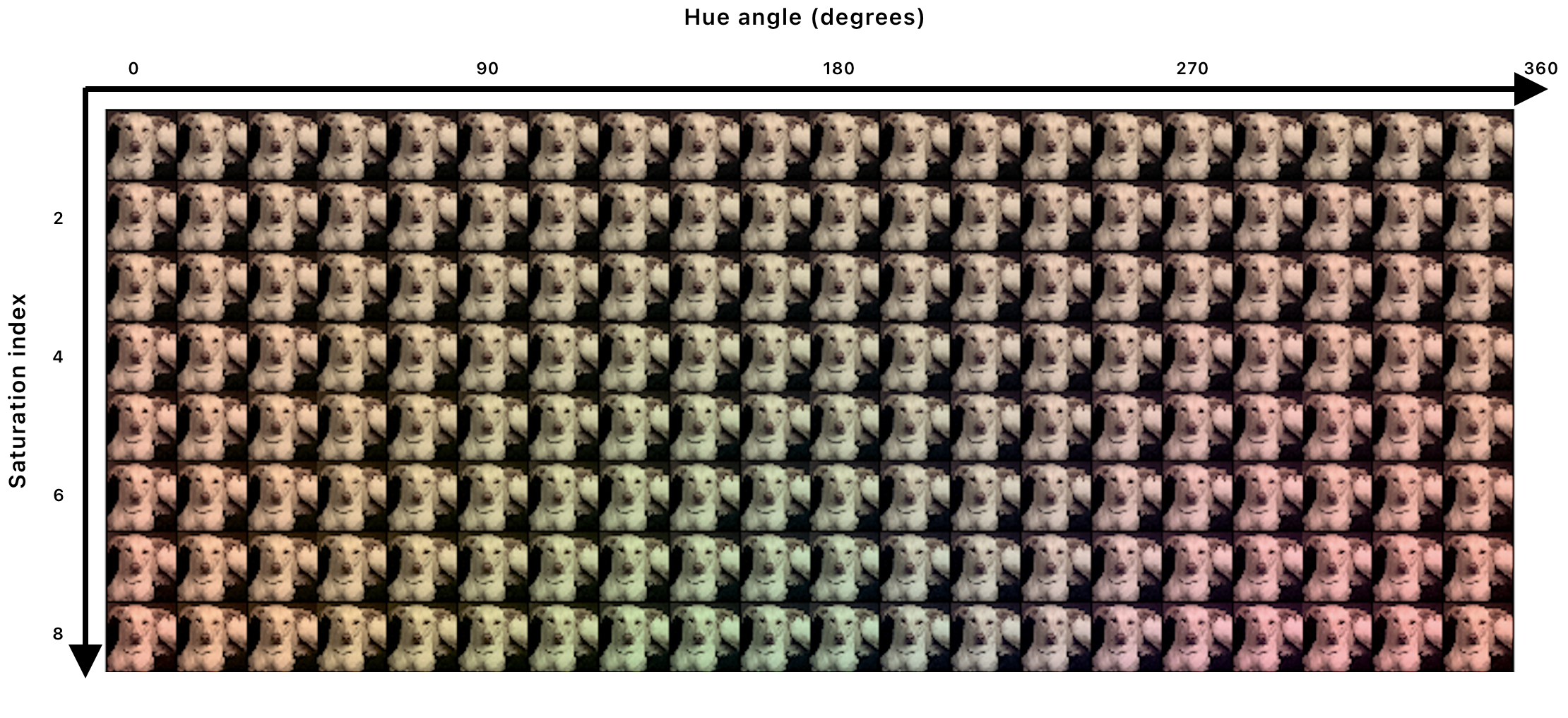}	
   \caption{Examples of illuminant change only applied to the Imagenet dataset in the same distribution as Figure~\ref{fig:hues}. There are changes for 20 hues (columns) in 8 intensities (rows).}
		\label{fig:cambio_iluminante}
   \end{figure*}

Specifically, from each dataset we selected 250 images to reduce the computational burden of applying all affine transformations and comparing all metrics; and they were modified so that, when applying the transformations, the resulting images would not present new artifacts or the central element would disappear. For the MNIST set images -originally with 28x28 images-, we simply enlarged the images (56x56) by adding black pixels around the original images to give us more room to move them. For color images, to avoid the appearance of black borders when applying some transformations, we decided to modify them and generate a mosaic. In addition to generating the mosaic, the images have a mirror effect to make the transitions between the different images smoother. Once the mosaic is created, the transformation is applied and a patch is taken from the original size of the database, thus preserving the size while including the modification. Figure~\ref{fig:mosaico} shows an example of the resulting modified images.

\subsection{Metrics}
\label{subsec:metrics}

We choose different metrics because of their relevance in the context of perceptual metrics, either because of their age and widespread use, or because of the good results obtained in other works and applied to other areas of study. These metrics are only chosen and analyzed to showcase our proposed methodology, we do not intend this work to be an in-depth analysis of state-of-the-art metrics for image quality. To this effect, the following metrics will be used:

\begin{itemize}
    \item Mean Squared Error (MSE): Measures the average squared difference between corresponding pixels in two images. It's a basic metric for image quality assessment, often used in image denoising and restoration tasks. However, it doesn't always align well with human perception.

    \item Structural Similarity Index (SSIM) \cite{SSIM}: For two images - the original and the distorted image - it computes three different comparisons: Luminance, contrast and structure. SSIM provides a more perceptually meaningful measure than MSE.

    \item Learned Perceptual Image Patch Similarity (LPIPS) \cite{LPIPS}: It uses a VGG trained with Imagenet to pass the images and compute distances in different feature spaces. Then, it performs a weighted sum so that the correlation with a perceptual database is maximized.

    \item Deep Image Structural Similarity (DISTS) \cite{DISTS}: As in LPIPS, it uses the VGG network, but in addition, it performs SSIM index at different intermediate layers. This new index combines sensitivity to structural distortions with tolerance to textures sampled elsewhere in the image. 

    \item PerceptNet \cite{Perceptnet}: It proposes an architecture to reflect the structure and stages of the early human visual system by considering a succession of canonical linear and non-linear operations. The network is trained to maximize correlation with a perceptual database.
    
    \item Perceptual Information Metric (PIM) \cite{PIM}: Unlike the other metrics, its training is based on two principles: efficient coding and slowness. On the one hand, it is compressive and, on the other, it captures persistent temporal information. This is achieved by training the network with images extracted from videos over which very short periods of time have passed, which should make it more robust to small variations in the object's movements or subtle changes in lighting. 
\end{itemize}

\section{Results}

Table \ref{tab:tabla_resultados} summarizes the numerical results of the experiments. For each geometric affine transformation, we present the discrimination thresholds per metric, $\theta_\tau^M$. In the case of illuminant changes, we show the errors made with respect to the interpolated MacAdam and the experimentally fitted ellipses. As the ellipses are very similar (Figure~\ref{fig:fit_elipses}), the errors are very similar. In those where they did not coincide completely, the mean of the errors has been calculated and marked with an asterisk (*) in the Table. The Figures from which these numerical results are derived are given in Appendix B and Appendix C to avoid cluttering the main text.

As we are considering both the human thresholds extracted from the literature, $\theta_\tau^{H_L}$,  and the threshold obtained by ourselves with natural images, $\theta_\tau^{H_N}$, a result is marked in bold if $\theta_\tau^{M}$ falls within $\theta_\tau^{H_L}$ and is underlined if it falls within $\theta_\tau^{H_N}$. For the ellipses, we check if the experimental and MacAdam ellipses fall within the uncertainty interval obtained from the psychometric functions. Note that a result can be marked both ways at the same time. A metric with many highlighted boxes will be a metric with a good performance from the point of view of human invariance.

 \begin{table*}[!h]
 \centering
 \caption{Distortion range of the metrics for each geometric affine distortion. For geometric transformations, intervals that match the literature threshold are marked in bold, while those that match the thresholds measured during the work are underlined. For illuminant changes, metrics that return a smaller error with respect to the fitted ellipses are marked in the same way as in geometric ones.}
\vspace{0.2cm}
 \resizebox{18cm}{!} {
\begin{tabular}{cc|c|c|c|c|c|c  | c|}
\cline{3-9}
\multicolumn{2}{c|}{} & RMSE & SSIM & LPIPS & DISTS & PerceptNet & PIM & Human \\ \hline
\multicolumn{1}{|c|}{Translation}&MNIST&0.025$\pm$0.003&0.030$\pm$0.004&0.05$\pm$0.03&0.034$\pm$0.008&0.031$\pm$0.004&0.05$\pm$0.03& \textbf{0.12}\\ \cline{2-8} 
\multicolumn{1}{|c|}{}&CIFAR-10&0.033$\pm$0.004&0.032$\pm$0.004&0.05$\pm$0.03&0.036$\pm$0.019&0.05$\pm$0.02&\textbf{\underline{0.13$\pm$0.12}}&\\\cline{2-8}  
\multicolumn{1}{|c|}{}&TID13&0.027$\pm$0.003&0.028$\pm$0.003&0.034$\pm$0.009&\textbf{0.08$\pm$0.10}&0.0029$\pm$0.003&0.06$\pm$0.04& \underline{0.23}\\\cline{2-8} 
\multicolumn{1}{|c|}{}&ImageNet&0.036$\pm$0.010&0.033$\pm$0.004&0.05$\pm$0.04&\textbf{\underline{0.13$\pm$0.15}}&0.036$\pm$0.010&\textbf{\underline{0.15$\pm$0.13}}& \\ \cline{2-8}  \hline \hline

\multicolumn{1}{|c|}{Rotation}&MNIST&\textbf{2$\pm$1}&\textbf{2.9$\pm$1.5}&3.7$\pm$2&1.2$\pm$0.8&\textbf{\underline{4.0$\pm$1.7}}&10.7$\pm$8.1&  \textbf{3}\\ \cline{2-8} 
\multicolumn{1}{|c|}{}&CIFAR-10&1.1$\pm$0.6&1.5$\pm$0.7&1.4$\pm$1.2&0.7$\pm$0.4&\textbf{\underline{3.6$\pm$1.8}}&1.7$\pm$0.9&\\ \cline{2-8}  
\multicolumn{1}{|c|}{}&TID13&0.10$\pm$0.03&0.11$\pm$0.04&0.10$\pm$0.05&0.1$\pm$0.3&0.11$\pm$0.04&0.09$\pm$0.08&\underline{3.6} \\ \cline{2-8}  
\multicolumn{1}{|c|}{}&ImageNet&0.19$\pm$0.09&0.15$\pm$0.06&0.15$\pm$0.11&0.11$\pm$0.05&0.22$\pm$0.09&0.2$\pm$0.5& \\ \cline{2-8}  \hline \hline


\multicolumn{1}{|c|}{Scale}&MNIST&1.10$\pm$0.04&1.19$\pm$0.12&\textbf{\underline{1.2$\pm$0.2}}&1.089$\pm$0.010&1.16$\pm$0.07&1.3$\pm$0.2&\textbf{1.03}\\ \cline{2-8} 
\multicolumn{1}{|c|}{}&CIFAR-10&1.052$\pm$0.006&1.055$\pm$0.007&1.061$\pm$0.007&1.054$\pm$0.007&1.067$\pm$0.018&\textbf{\underline{1.06$\pm$0.03}}& \\\cline{2-8}  
\multicolumn{1}{|c|}{}&TID13&1.047$\pm$0.006&1.048$\pm$0.006&1.057$\pm$0.007&1.062$\pm$0.008&\textbf{\underline{1.040$\pm$0.005}}&1.059$\pm$0.007&\underline{1.03} \\ \cline{2-8}  
\multicolumn{1}{|c|}{}&ImageNet&1.0080$\pm$0.0010&1.0074$\pm$0.0009&1.0086$\pm$0.0011&1.00813$\pm$0.00010&1.0082$\pm$0.0010&1.009$\pm$0.005& \\ \cline{2-8}  \hline \hline

\multicolumn{1}{|c|}{Illum}&MNIST&0.03$\pm$0.06*&-&-&\textbf{\underline{0.016$\pm$0.018*}}&-&0.05$\pm$0.05*&\textbf{MacAdam}\\ \cline{2-8} 
\multicolumn{1}{|c|}{}&CIFAR-10&0.007*$\pm$0.009*&0.03$\pm$0.05*&0.03$\pm$0.03*&\textbf{\underline{0.005$\pm$0.013}}*&0.05$\pm$0.13*&0.06$\pm$0.06*& \\\cline{2-8}  
\multicolumn{1}{|c|}{}&TID13&0.010$\pm$0.010*&0.02$\pm$0.02*&0.02$\pm$0.02&0.02$\pm$0.02&0.05$\pm$0.09*&\textbf{\underline{0.005$\pm$0.012}}*& \underline{Our ellipse}\\ \cline{2-8}  
\multicolumn{1}{|c|}{}&ImageNet&0.008$\pm$0.010&0.02$\pm$0.02*&0.02$\pm$0.02&0.02$\pm$0.02&0.05$\pm$0.09*& \textbf{\underline{0.005$\pm$0.012}}*& \\ \cline{2-8}  \hline 

\end{tabular}
}
\label{tab:tabla_resultados}
\end{table*}

The results from Table \ref{tab:tabla_resultados} show that following our strong invariance criteria, there is no clear winner. No artificial model shows the required robustness to affine transformations nor type of stimuli.

The comparison of the order between the human thresholds and the sensitivities of the metrics can be seen in Table~\ref{tab:resultados2}, where even if no metric shows complete human behavior, most of them maintain chromatic order in contrast to geometric transformations.

\begin{table*}[]
\centering

 \caption{Results associated with the least demanding test, on the order of the sensitivities and their comparison with the human order. The first row of the table shows the order of the humans, which does not change depending on the database. Displayed in two different colors: In purple, the order of the geometric transformations and, in pink, the order of the YB and RG illuminants.}
\resizebox{16cm}{!} {

\begin{tabular}{l|c|c|c|c|}
\cline{2-5}
 & MNIST & CIFAR-10 & TID13 & ImageNet \\ \hline
\multicolumn{1}{|l|}{\textcolor{magenta}{Human}} & \textcolor{magenta}{S$>$R$>$T$>$}\textcolor{pink}{RG$>$YB} & \textcolor{magenta}{S$>$R$>$T$>$}\textcolor{pink}{RG$>$YB} & \textcolor{magenta}{S$>$T$>$R$>$}\textcolor{pink}{RG$>$YB} & \textcolor{magenta}{S$>$T$>$R$>$}\textcolor{pink}{RG$>$YB} \\ \hline
\multicolumn{1}{|l|}{SSIM} & R$>$S$>$T$>$\textcolor{pink}{RG$>$YB} & YB$>$T$>$S$>$RG$>$R & YB$>$S$>$R$>$T$>$RG & YB$>$T$>$S$>$RG$>$\textcolor{magenta}{R} \\ \hline
\multicolumn{1}{|l|}{LPIPS} & R$>$\textcolor{pink}{RG$>$YB}$>$S$>$T & R$>$YB$>$RG$>$S$>$T & YB$>$RG$>$R$>$S$>$T & YB$>$R$>$RG$>$T$>$S \\ \hline
\multicolumn{1}{|l|}{DISTS} & \textcolor{pink}{RG}$>$S$>$\textcolor{pink}{YB}$>$R$>$T & R$>$\textcolor{pink}{RG$>$YB}$>$S$>$T & R$>$\textcolor{pink}{RG$>$YB}$>$S$>$T & R$>$S$>$\textcolor{pink}{RG$>$YB}$>$T \\ \hline
\multicolumn{1}{|l|}{PerceptNet} & S$>$T$>$R$>$\textcolor{pink}{RG$>$YB} & T$>$S$>$\textcolor{pink}{RG}$>$R$>$\textcolor{pink}{YB} & S$>$R$>$T$>$YB$>$RG & T$>$S$>$R$>$YB$>$RG \\ \hline
\multicolumn{1}{|l|}{PIM} & \textcolor{pink}{RG$>$YB}$>$R$>$S$>$T & R$>$\textcolor{pink}{RG$>$YB}$>$S$>$T &\textcolor{pink}{RG$>$YB}$>$R$>$S$>$T & YB$>$RG$>$R$>$S$>$T \\ \hline
\end{tabular}
}
\label{tab:resultados2}
\end{table*}

\section{Discussion}

\subsection{Analysis of the results}

\begin{figure}[!h]
\centering
			\includegraphics[width=1\linewidth]{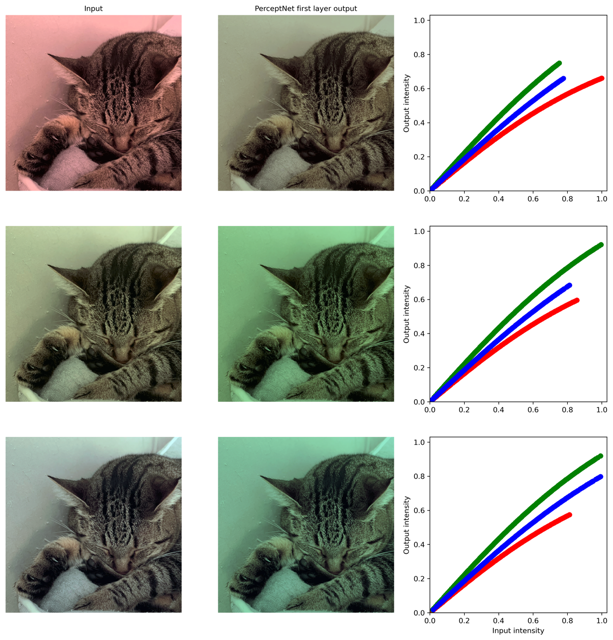}	
		
		\caption{Example of the operation of the first layers of the PerceptNet network. For different inputs with different illuminants, the results of the first layers are all green, which explains their poor performance in the experiments on illuminant changes.}
		\label{fig:PN_green}
	\end{figure}

As we have seen in Table \ref{tab:tabla_resultados}, bearing in mind that no model is highlighted for all related transformations, the conclusion is that there is no metric that behaves, in terms of invariance, like the human being. We can dissect the behavior for the different affine transformation as follows:

\begin{itemize}
    \item Translation: There is no particular model that shows human-like translation invariance but DISTS and PIM are the best in this regard.
    \item Rotation: Perceptnet shows the most human-like behavior but it's very dependent on the database used.
    \item Scale: There are isolated highlighted cells. In general, all the models seem less sensitive than humans except in the ImageNet database, where the threshold is an order of magnitude smaller than the human threshold.
    \item Illuminants: Two models rise above the rest. On one side, DISTS has a better performance with small images, on the other side, PIM performs better with big images. It caught our attention that PerceptNet's discrimination ellipse has a radically different orientation. We attribute this to the fact that the chromatic transformations it applies at its first layers, make all the images greener as shown in Figure~\ref{fig:PN_green}.
\end{itemize}

Getting into the metrics' performance regarding the ordering of the transformations from Table \ref{tab:resultados2}, PerceptNet is the only metric that almost maintains the ordering of being more sensitive to geometric over chromatic transformations. All the metrics are generally more sensitive to RG than YB, which matches human performance in energy definition. All in all, no metric can be said to have human-like behavior even with the least strict test.

In addition, a collateral result that can be obtained from our work is specific invisibility thresholds for each metric. That is, thresholds in units of each metric that indicate variations for which the distance is imperceptible to the human eye. These values are shown in Table \ref{tab:collateral}.

\begin{table*}[!h]
\centering
\caption{Metric-specific invisibility thresholds. Indicates distance variations for which changes between two images are imperceptible to humans. For example, a value equal to or less than 3.09 in PIM implies that these two images are indistinguishable for the observer.}
\begin{tabular}{l|l|l|l|l|l|l|}
\cline{2-7}
 & RMSE & SSIM & LPIPS & DISTS & PerceptNet & PIM \\ \hline
\multicolumn{1}{|l|}{Invisibility Threshold} & \multicolumn{1}{c|}{0.020} & \multicolumn{1}{c|}{0.016} & \multicolumn{1}{c|}{0.017} & \multicolumn{1}{c|}{0.017} & \multicolumn{1}{c|}{0.008} & \multicolumn{1}{c|}{3.09} \\ \hline
\end{tabular}
\label{tab:collateral}
\end{table*}

\subsection{Known limitations and future work}
\label{sec:future}

As this work relies heavily on experimental measurements, multiple experimental error sources influence the invisibility thresholds measured for the artificial models. Going back to Figure \ref{fig:Fig1}, we can identify three different sources of uncertainty. These uncertainties were excluded from the representation to avoid cluttering, but have to be taken into consideration: (1) Human visibility thresholds (black line in Figure~\ref{fig:Fig1}), $\theta^{H}_{\tau}$, are measured with experimental psychometric function and have an associated experimental error. (2) Threshold in the internal distance representation (yellow line in Figure~\ref{fig:Fig1}), $\mathcal{D}_{\tau}$, are obtained via psychometric function and, as such, have an experimental uncertainty. (3) Evaluations of different metrics (continuous curves in blue, red, and green in Figure~\ref{fig:Fig1}) are obtained by applying the transduction function to the responses of the metric. This transduction function is obtained by performing a  least squares fit to a given dataset (TID13), inducing another experimental error.

In order to mitigate the experimental uncertainties aforementioned, there are different avenues of work that could be taken into the future: (1, 2) increasing the number of individuals and image samples used to calculate the psychometric functions would reduce their experimental uncertainty. In particular, (2) would see more improvement because currently, there are few images in lower DMOS ranges. (3) Using a different database with more distortions and images would lead to a more precise least-squares fitting, thus reducing its uncertainty.

One could say that another one of the caveats of the methodology proposed in Section \ref{subsec:human_internal} is using TID13, as it does not contain affine transformations in its range of distortions. A clear path looking forward would be repeating the TID13 experiment but including these affine transformations to conform a bigger and more general dataset which could be used to repeat the methodology and re-measure the threshold in the common internal representation. It has to be said that this was not done in this paper because it would imply a lot of experimentation time that was not the main focus of the project. All in all, we consider this would make the proposed methodology more robust but we do not foresee a drastical change in the results.

\section{Conclusion}
\label{sec:conclussion}

As a complement to the usual reproduction of subjective quality ratings, we argue that the invisibility thresholds of perceptual metrics should also correspond to the invisibility thresholds of humans. This direct comparison (particularly interesting for affine transformations) is a strong test for metrics because human thresholds can be accurately measured with classical psychophysics experiments. This comparison requires (a)~data of human thresholds and (b)~metric thresholds.

Regarding human thresholds, we used both classical and explicitly measured values of the thresholds obtained using the same kind of natural images employed in perceptual metrics. Whereas for metric thresholds, we propose a methodology to assign invisibility regions for any particular metric, which can be expressed in units of the metric themselves or in physical units to facilitate direct comparison with the results of human psychophysics.


We also propose a less restrictive test: instead of reproducing the exact threshold values, we evaluate if the metric just matches the sensitivity ordering. That means, using sensitivity as the inverse of the threshold energy, one can check whether the metrics are more sensitive to one distortion or the other. This ordering can be then compared to the one in humans.

Making the demanding comparison between human and metric thresholds for a range of well-known and stablished deep image quality metrics shows that none of the studied metrics (both deep-learning-based as well as RSMSE and SSIM) succeeds under these criteria: they do not reproduce all the human thresholds for the different affine transformations, and they do not reproduce the order of sensitivities in humans. No metric is capable of reproducing the thresholds of invisibility and human sensibilities.

This means that tuning the models exclusively to predict quality ratings may disregard other properties of human vision as invariances or invisibility thresholds.

\section*{Acknowledgment}
This work was supported in part by MICIIN/FEDER/UE under Grant PID2020-118071GB-I00  and PDC2021-121522-C21, in part by Generalitat Valenciana under Projects GV/2021/ 074, CIPROM/2021/056, CIAPOT/2021/9, CIACIF/2023/223 and in part by valgrAI - GVA. Some computer resources were provided by Artemisa, funded by the European Union ERDF and Comunitat Valenciana as well as the technical support provided by the Instituto de Física Corpuscular, IFIC (CSIC-UV).




%

\bibliographystyle{cas-model2-names}

\bibliography{cas-refs}

\begin{thebibliography}{48}
\expandafter\ifx\csname natexlab\endcsname\relax\def\natexlab#1{#1}\fi
\providecommand{\url}[1]{\texttt{#1}}
\providecommand{\href}[2]{#2}
\providecommand{\path}[1]{#1}
\providecommand{\DOIprefix}{doi:}
\providecommand{\ArXivprefix}{arXiv:}
\providecommand{\URLprefix}{URL: }
\providecommand{\Pubmedprefix}{pmid:}
\providecommand{\doi}[1]{\href{http://dx.doi.org/#1}{\path{#1}}}
\providecommand{\Pubmed}[1]{\href{pmid:#1}{\path{#1}}}
\providecommand{\bibinfo}[2]{#2}
\ifx\xfnm\relax \def\xfnm[#1]{\unskip,\space#1}\fi
\bibitem[{Baldwin et~al.(2017)Baldwin, Fu, Farivar and Hess}]{Baldwin16}
\bibinfo{author}{Baldwin, A.}, \bibinfo{author}{Fu, M.}, \bibinfo{author}{Farivar, R.}, \bibinfo{author}{Hess, R.}, \bibinfo{year}{2017}.
\newblock \bibinfo{title}{The equivalent internal orientation and position noise for contour integration}.
\newblock \bibinfo{journal}{Scientific Reports} \bibinfo{volume}{7}.
\newblock \DOIprefix\doi{10.1038/s41598-017-13244-z}.
\bibitem[{Bhardwaj et~al.(2021)Bhardwaj, Fischer, Ballé and Chinen}]{PIM}
\bibinfo{author}{Bhardwaj, S.}, \bibinfo{author}{Fischer, I.}, \bibinfo{author}{Ballé, J.}, \bibinfo{author}{Chinen, T.}, \bibinfo{year}{2021}.
\newblock \bibinfo{title}{An unsupervised information-theoretic perceptual quality metric}.
\newblock \href{http://arxiv.org/abs/2006.06752}{\tt arXiv:2006.06752}.
\bibitem[{Bruna and Mallat(2013)}]{Bruna13_scatter}
\bibinfo{author}{Bruna, J.}, \bibinfo{author}{Mallat, S.}, \bibinfo{year}{2013}.
\newblock \bibinfo{title}{Invariant scattering convolution networks}.
\newblock \bibinfo{journal}{IEEE Transactions on Pattern Analysis and Machine Intelligence} \bibinfo{volume}{35}, \bibinfo{pages}{1872--1886}.
\newblock \DOIprefix\doi{10.1109/TPAMI.2012.230}.
\bibitem[{Campbell and Robson(1968)}]{Campbell68}
\bibinfo{author}{Campbell, F.}, \bibinfo{author}{Robson, J.}, \bibinfo{year}{1968}.
\newblock \bibinfo{title}{Application of fourier analysis to the visibility of gratings}.
\newblock \bibinfo{journal}{The Journal of Physiology} \DOIprefix\doi{10.1113/jphysiol.1968.sp008574}.
\bibitem[{Chang and Yeh(2022)}]{ICV}
\bibinfo{author}{Chang, H.H.}, \bibinfo{author}{Yeh, C.H.}, \bibinfo{year}{2022}.
\newblock \bibinfo{title}{Face anti-spoofing detection based on multi-scale image quality assessment}.
\newblock \bibinfo{journal}{Image and Vision Computing} \bibinfo{volume}{121}.
\newblock \DOIprefix\doi{https://doi.org/10.1016/j.imavis.2022.104428}.
\bibitem[{Deng et~al.(2009)Deng, Dong, Socher, Li, Li and Fei-Fei}]{Imagenet}
\bibinfo{author}{Deng, J.}, \bibinfo{author}{Dong, W.}, \bibinfo{author}{Socher, R.}, \bibinfo{author}{Li, L.J.}, \bibinfo{author}{Li, K.}, \bibinfo{author}{Fei-Fei, L.}, \bibinfo{year}{2009}.
\newblock \bibinfo{title}{Imagenet: A large-scale hierarchical image database}, in: \bibinfo{booktitle}{2009 IEEE Conference on Computer Vision and Pattern Recognition}, pp. \bibinfo{pages}{248--255}.
\newblock \DOIprefix\doi{10.1109/CVPR.2009.5206848}.
\bibitem[{Ding et~al.(2020)Ding, Ma, Wang and Simoncelli}]{DISTS}
\bibinfo{author}{Ding, K.}, \bibinfo{author}{Ma, K.}, \bibinfo{author}{Wang, S.}, \bibinfo{author}{Simoncelli, E.P.}, \bibinfo{year}{2020}.
\newblock \bibinfo{title}{Image quality assessment: Unifying structure and texture similarity}.
\newblock \bibinfo{journal}{{IEEE} Transactions on Pattern Analysis and Machine Intelligence} , \bibinfo{pages}{1--1}\URLprefix \url{https://doi.org/10.1109\%2Ftpami.2020.3045810}, \DOIprefix\doi{10.1109/tpami.2020.3045810}.
\bibitem[{Eckert and Bradley(1998)}]{IQA_compression}
\bibinfo{author}{Eckert, M.}, \bibinfo{author}{Bradley, A.}, \bibinfo{year}{1998}.
\newblock \bibinfo{title}{Perceptual quality metrics applied to still image compression}.
\newblock \bibinfo{journal}{Signal Processing} \bibinfo{volume}{70}, \bibinfo{pages}{177--200}.
\newblock \URLprefix \url{https://doi.org/10.1016/S0165-1684(98)00124-8}.
\bibitem[{Egiazarian et~al.(2018)Egiazarian, Ponomarenko, Lukin and Ieremeiev}]{IQA_denoising}
\bibinfo{author}{Egiazarian, K.}, \bibinfo{author}{Ponomarenko, M.}, \bibinfo{author}{Lukin, V.}, \bibinfo{author}{Ieremeiev, O.}, \bibinfo{year}{2018}.
\newblock \bibinfo{title}{Statistical evaluation of visual quality metrics for image denoising}, in: \bibinfo{booktitle}{2018 IEEE International Conference on Acoustics, Speech and Signal Processing (ICASSP)}, pp. \bibinfo{pages}{6752--6756}.
\newblock \DOIprefix\doi{10.1109/ICASSP.2018.8462294}.
\bibitem[{Fairchild(2013)}]{chrom_adap}
\bibinfo{author}{Fairchild, M.D.}, \bibinfo{year}{2013}.
\newblock \bibinfo{title}{Front Matter}. \bibinfo{publisher}{John Wiley and Sons, Ltd}.
\newblock pp. \bibinfo{pages}{i--xxii}.
\newblock \DOIprefix\doi{https://doi.org/10.1002/9781118653128.fmatter}.
\bibitem[{Freeman and Simoncelli(2011)}]{FreemanSimoncelli11_NN}
\bibinfo{author}{Freeman, J.}, \bibinfo{author}{Simoncelli, E.}, \bibinfo{year}{2011}.
\newblock \bibinfo{title}{Metamers of the visual stream}.
\newblock \bibinfo{journal}{Nature neuroscience} \bibinfo{volume}{14}, \bibinfo{pages}{1195--201}.
\newblock \DOIprefix\doi{10.1038/nn.2889}.
\bibitem[{Heeger(1992)}]{contrast_const}
\bibinfo{author}{Heeger, D.J.}, \bibinfo{year}{1992}.
\newblock \bibinfo{title}{Normalization of cell responses in cat striate cortex}.
\newblock \bibinfo{journal}{Visual Neuroscience} \bibinfo{volume}{9}, \bibinfo{pages}{181–197}.
\newblock \DOIprefix\doi{10.1017/S0952523800009640}.
\bibitem[{Hepburn et~al.(2020)Hepburn, Laparra, Malo, McConville and Santos-Rodriguez}]{Perceptnet}
\bibinfo{author}{Hepburn, A.}, \bibinfo{author}{Laparra, V.}, \bibinfo{author}{Malo, J.}, \bibinfo{author}{McConville, R.}, \bibinfo{author}{Santos-Rodriguez, R.}, \bibinfo{year}{2020}.
\newblock \bibinfo{title}{Perceptnet: A human visual system inspired neural network for estimating perceptual distance}, in: \bibinfo{booktitle}{2020 {IEEE} International Conference on Image Processing ({ICIP})}, \bibinfo{publisher}{{IEEE}}.
\newblock \URLprefix \url{https://doi.org/10.1109\%2Ficip40778.2020.9190691}, \DOIprefix\doi{10.1109/icip40778.2020.9190691}.
\bibitem[{Hepburn et~al.(2022)Hepburn, Laparra, Santos-Rodriguez, Ball{\'e} and Malo}]{Hepburn22}
\bibinfo{author}{Hepburn, A.}, \bibinfo{author}{Laparra, V.}, \bibinfo{author}{Santos-Rodriguez, R.}, \bibinfo{author}{Ball{\'e}, J.}, \bibinfo{author}{Malo, J.}, \bibinfo{year}{2022}.
\newblock \bibinfo{title}{On the relation between statistical learning and perceptual distances}, in: \bibinfo{booktitle}{International Conference on Learning Representations}.
\newblock \URLprefix \url{https://openreview.net/forum?id=zXM0b4hi5\_B}.
\bibitem[{Jennings and Barbur(2010)}]{Barbur10}
\bibinfo{author}{Jennings, B.}, \bibinfo{author}{Barbur, J.}, \bibinfo{year}{2010}.
\newblock \bibinfo{title}{Colour detection thresholds as a function of chromatic adaptation and light level}.
\newblock \bibinfo{journal}{Ophthalmic \& physiological optics : the journal of the British College of Ophthalmic Opticians (Optometrists)} \bibinfo{volume}{30}, \bibinfo{pages}{560--7}.
\newblock \DOIprefix\doi{10.1111/j.1475-1313.2010.00773.x}.
\bibitem[{Johnson et~al.(2011)Johnson, Krupinski, Yan, Roehrig, Graham and Weinstein}]{pf_artifacts}
\bibinfo{author}{Johnson, J.}, \bibinfo{author}{Krupinski, E.}, \bibinfo{author}{Yan, M.}, \bibinfo{author}{Roehrig, H.}, \bibinfo{author}{Graham, A.}, \bibinfo{author}{Weinstein, R.}, \bibinfo{year}{2011}.
\newblock \bibinfo{title}{Using a visual discrimination model for the detection of compression artifacts in virtual pathology images}.
\newblock \bibinfo{journal}{IEEE Trans. Med. Imaging} \bibinfo{volume}{30}, \bibinfo{pages}{306--314}.
\bibitem[{Kelly(1979)}]{Kelly79}
\bibinfo{author}{Kelly, D.H.}, \bibinfo{year}{1979}.
\newblock \bibinfo{title}{Motion and vision. ii. stabilized spatio-temporal threshold surface}.
\newblock \bibinfo{journal}{J. Opt. Soc. Am.} \bibinfo{volume}{69}, \bibinfo{pages}{1340--1349}.
\newblock \URLprefix \url{http://www.osapublishing.org/abstract.cfm?URI=josa-69-10-1340}, \DOIprefix\doi{10.1364/JOSA.69.001340}.
\bibitem[{Kingdom and Prins(2009)}]{Kingdom13}
\bibinfo{author}{Kingdom, F.A.}, \bibinfo{author}{Prins, N.}, \bibinfo{year}{2009}.
\newblock \bibinfo{title}{Psychophysics}.
\newblock \bibinfo{publisher}{Academic Press}.
\bibitem[{Krizhevsky(2009)}]{cifar10}
\bibinfo{author}{Krizhevsky, A.}, \bibinfo{year}{2009}.
\newblock \bibinfo{title}{Learning multiple layers of features from tiny images}.
\newblock \URLprefix \url{https://api.semanticscholar.org/CorpusID:18268744}.
\bibitem[{Laparra et~al.(2016)Laparra, Ballé, Berardino and Simoncelli}]{Laparra16_NLPD}
\bibinfo{author}{Laparra, V.}, \bibinfo{author}{Ballé, J.}, \bibinfo{author}{Berardino, A.}, \bibinfo{author}{Simoncelli, E.}, \bibinfo{year}{2016}.
\newblock \bibinfo{title}{Perceptual image quality assessment using a normalized laplacian pyramid}.
\newblock \bibinfo{journal}{Electronic Imaging} \bibinfo{volume}{2016}, \bibinfo{pages}{1--6}.
\newblock \DOIprefix\doi{10.2352/ISSN.2470-1173.2016.16.HVEI-103}.
\bibitem[{Laparra et~al.(2017)Laparra, Berardino, Ballé and Simoncelli}]{LaparraJOSA17}
\bibinfo{author}{Laparra, V.}, \bibinfo{author}{Berardino, A.}, \bibinfo{author}{Ballé, J.}, \bibinfo{author}{Simoncelli, E.P.}, \bibinfo{year}{2017}.
\newblock \bibinfo{title}{Perceptually optimized image rendering}.
\newblock \bibinfo{journal}{Journal of the Optical Society of America A} \bibinfo{volume}{34}, \bibinfo{pages}{1511}.
\newblock \URLprefix \url{http://dx.doi.org/10.1364/JOSAA.34.001511}, \DOIprefix\doi{10.1364/josaa.34.001511}.
\bibitem[{Laparra et~al.(2010)Laparra, Muñoz and Malo}]{Laparra10}
\bibinfo{author}{Laparra, V.}, \bibinfo{author}{Muñoz, J.}, \bibinfo{author}{Malo, J.}, \bibinfo{year}{2010}.
\newblock \bibinfo{title}{Divisive normalization image quality metric revisited}.
\newblock \bibinfo{journal}{Journal of the Optical Society of America. A, Optics, image science, and vision} \bibinfo{volume}{27}, \bibinfo{pages}{852--64}.
\newblock \DOIprefix\doi{10.1364/JOSAA.27.000852}.
\bibitem[{Lecun et~al.(1998)Lecun, Bottou, Bengio and Haffner}]{MNIST}
\bibinfo{author}{Lecun, Y.}, \bibinfo{author}{Bottou, L.}, \bibinfo{author}{Bengio, Y.}, \bibinfo{author}{Haffner, P.}, \bibinfo{year}{1998}.
\newblock \bibinfo{title}{Gradient-based learning applied to document recognition}.
\newblock \bibinfo{journal}{Proceedings of the IEEE} \bibinfo{volume}{86}, \bibinfo{pages}{2278--2324}.
\newblock \DOIprefix\doi{10.1109/5.726791}.
\bibitem[{Legge and Campbell(1981)}]{umbral_traslacion}
\bibinfo{author}{Legge, G.E.}, \bibinfo{author}{Campbell, F.}, \bibinfo{year}{1981}.
\newblock \bibinfo{title}{Displacement detection in human vision}.
\newblock \bibinfo{journal}{Vision Research} \bibinfo{volume}{21}, \bibinfo{pages}{205--213}.
\newblock \URLprefix \url{https://www.sciencedirect.com/science/article/pii/0042698981901140}, \DOIprefix\doi{https://doi.org/10.1016/0042-6989(81)90114-0}.
\bibitem[{Ma et~al.(2018)Ma, Duanmu and Wang}]{parecido1}
\bibinfo{author}{Ma, K.}, \bibinfo{author}{Duanmu, Z.}, \bibinfo{author}{Wang, Z.}, \bibinfo{year}{2018}.
\newblock \bibinfo{title}{Geometric transformation invariant image quality assessment using convolutional neural networks}, in: \bibinfo{booktitle}{2018 IEEE International Conference on Acoustics, Speech and Signal Processing (ICASSP)}, pp. \bibinfo{pages}{6732--6736}.
\newblock \DOIprefix\doi{10.1109/ICASSP.2018.8462176}.
\bibitem[{MacAdam(1942)}]{MacAdam}
\bibinfo{author}{MacAdam, D.L.}, \bibinfo{year}{1942}.
\newblock \bibinfo{title}{Visual sensitivities to color differences in daylight$\ast$}.
\newblock \bibinfo{journal}{J. Opt. Soc. Am.} \bibinfo{volume}{32}, \bibinfo{pages}{247--274}.
\newblock \URLprefix \url{https://opg.optica.org/abstract.cfm?URI=josa-32-5-247}, \DOIprefix\doi{10.1364/JOSA.32.000247}.
\bibitem[{Malo et~al.(1997)Malo, Pons and Artigas}]{Malo97}
\bibinfo{author}{Malo, J.}, \bibinfo{author}{Pons, A.}, \bibinfo{author}{Artigas, J.}, \bibinfo{year}{1997}.
\newblock \bibinfo{title}{Subjective image fidelity metric based on bit allocation of the human visual system in the dct domain}.
\newblock \bibinfo{journal}{Image and Vision Computing} \bibinfo{volume}{15}, \bibinfo{pages}{535--548}.
\newblock \URLprefix \url{https://doi.org/10.1016/S0262-8856(96)00004-2}.
\bibitem[{Maloney and Wandell(1987)}]{color_const}
\bibinfo{author}{Maloney, L.T.}, \bibinfo{author}{Wandell, B.A.}, \bibinfo{year}{1987}.
\newblock \bibinfo{title}{Color constancy: a method for recovering surface spectral reflectance}, in: \bibinfo{editor}{Fischler, M.A.}, \bibinfo{editor}{Firschein, O.} (Eds.), \bibinfo{booktitle}{Readings in Computer Vision}. \bibinfo{publisher}{Morgan Kaufmann}, \bibinfo{address}{San Francisco (CA)}, pp. \bibinfo{pages}{293--297}.
\newblock \URLprefix \url{https://www.sciencedirect.com/science/article/pii/B9780080515816500349}, \DOIprefix\doi{https://doi.org/10.1016/B978-0-08-051581-6.50034-9}.
\bibitem[{Mantiuk et~al.(2011)Mantiuk, Kim, Rempel and Heidrich}]{Mantiuk11}
\bibinfo{author}{Mantiuk, R.K.}, \bibinfo{author}{Kim, K.J.}, \bibinfo{author}{Rempel, A.G.}, \bibinfo{author}{Heidrich, W.}, \bibinfo{year}{2011}.
\newblock \bibinfo{title}{Hdr-vdp-2: a calibrated visual metric for visibility and quality predictions in all luminance conditions}.
\newblock \bibinfo{journal}{ACM SIGGRAPH 2011 papers} \URLprefix \url{https://api.semanticscholar.org/CorpusID:756729}.
\bibitem[{Martinez-Garcia et~al.(2019)Martinez-Garcia, Bertalmío and Malo}]{Martinez19}
\bibinfo{author}{Martinez-Garcia, M.}, \bibinfo{author}{Bertalmío, M.}, \bibinfo{author}{Malo, J.}, \bibinfo{year}{2019}.
\newblock \bibinfo{title}{In praise of artifice reloaded: Caution with subjective image quality databases}.
\newblock \bibinfo{journal}{Frontiers in Neuroscience} \href{http://arxiv.org/abs/1801.09632}{\tt arXiv:1801.09632}.
\bibitem[{Martinez-Garcia et~al.(2018)Martinez-Garcia, Cyriac, Batard, Bertalmío and Malo}]{Martinez18}
\bibinfo{author}{Martinez-Garcia, M.}, \bibinfo{author}{Cyriac, P.}, \bibinfo{author}{Batard, T.}, \bibinfo{author}{Bertalmío, M.}, \bibinfo{author}{Malo, J.}, \bibinfo{year}{2018}.
\newblock \bibinfo{title}{Derivatives and inverse of cascaded linear+nonlinear neural models}.
\newblock \bibinfo{journal}{PLOS ONE} \bibinfo{volume}{13}, \bibinfo{pages}{e0201326}.
\newblock \URLprefix \url{http://dx.doi.org/10.1371/journal.pone.0201326}, \DOIprefix\doi{10.1371/journal.pone.0201326}.
\bibitem[{Mather et~al.(1998)Mather, Verstraten and Anstis}]{motion_compens}
\bibinfo{author}{Mather, G.}, \bibinfo{author}{Verstraten, F.}, \bibinfo{author}{Anstis, S.}, \bibinfo{year}{1998}.
\newblock \bibinfo{publisher}{MIT Press}.
\newblock \URLprefix \url{https://direct.mit.edu/books/edited-volume/4697/The-Motion-AftereffectA-Modern-Perspective}, \DOIprefix\doi{https://doi.org/10.7551/mitpress/4779.001.0001}.
\bibitem[{Ninassi et~al.(2006)Ninassi, Autrusseau and Le~Callet}]{pf_NR}
\bibinfo{author}{Ninassi, A.}, \bibinfo{author}{Autrusseau, F.}, \bibinfo{author}{Le~Callet, P.}, \bibinfo{year}{2006}.
\newblock \bibinfo{title}{Pseudo no reference image quality metric using perceptual data hiding}.
\newblock \bibinfo{journal}{Human Vision and Electronic Imaging} \DOIprefix\doi{10.1117/12.650780}.
\bibitem[{oy and Flusser(1998)}]{ajuste_elipses}
\bibinfo{author}{oy, R.H.}, \bibinfo{author}{Flusser, J.}, \bibinfo{year}{1998}.
\newblock \bibinfo{title}{Numerically stable direct least squares fitting of ellipses}.
\newblock \URLprefix \url{https://api.semanticscholar.org/CorpusID:15772208}.
\bibitem[{Ponomarenko et~al.(2015)Ponomarenko, Jin, Ieremeiev, Lukin, Egiazarian, Astola, Vozel, Chehdi, Carli, Battisti and {Jay Kuo}}]{TID13}
\bibinfo{author}{Ponomarenko, N.}, \bibinfo{author}{Jin, L.}, \bibinfo{author}{Ieremeiev, O.}, \bibinfo{author}{Lukin, V.}, \bibinfo{author}{Egiazarian, K.}, \bibinfo{author}{Astola, J.}, \bibinfo{author}{Vozel, B.}, \bibinfo{author}{Chehdi, K.}, \bibinfo{author}{Carli, M.}, \bibinfo{author}{Battisti, F.}, \bibinfo{author}{{Jay Kuo}, C.C.}, \bibinfo{year}{2015}.
\newblock \bibinfo{title}{Image database tid2013: Peculiarities, results and perspectives}.
\newblock \bibinfo{journal}{Signal Processing: Image Communication} \bibinfo{volume}{30}, \bibinfo{pages}{57--77}.
\newblock \URLprefix \url{https://www.sciencedirect.com/science/article/pii/S0923596514001490}, \DOIprefix\doi{https://doi.org/10.1016/j.image.2014.10.009}.
\bibitem[{Regan et~al.(1996)Regan, Gray and Hamstra}]{umbral_rotacion}
\bibinfo{author}{Regan, D.}, \bibinfo{author}{Gray, R.}, \bibinfo{author}{Hamstra, S.}, \bibinfo{year}{1996}.
\newblock \bibinfo{title}{Evidence for a neural mechanism that encodes angles}.
\newblock \bibinfo{journal}{Vision Research} \bibinfo{volume}{36}, \bibinfo{pages}{323--IN3}.
\newblock \URLprefix \url{https://www.sciencedirect.com/science/article/pii/004269899500113E}, \DOIprefix\doi{https://doi.org/10.1016/0042-6989(95)00113-E}.
\bibitem[{Scheirer et~al.(2014)Scheirer, Anthony, Nakayama and Cox}]{pf_computervision}
\bibinfo{author}{Scheirer, W.J.}, \bibinfo{author}{Anthony, S.E.}, \bibinfo{author}{Nakayama, K.}, \bibinfo{author}{Cox, D.D.}, \bibinfo{year}{2014}.
\newblock \bibinfo{title}{Perceptual annotation: Measuring human vision to improve computer vision}.
\newblock \bibinfo{journal}{IEEE Transactions on Pattern Analysis and Machine Intelligence} \bibinfo{volume}{36}, \bibinfo{pages}{1679--1686}.
\newblock \DOIprefix\doi{10.1109/TPAMI.2013.2297711}.
\bibitem[{Shi et~al.(2015)Shi, Ngan, Li, Paramesran and Li}]{pf_segmentation}
\bibinfo{author}{Shi, R.}, \bibinfo{author}{Ngan, K.}, \bibinfo{author}{Li, S.}, \bibinfo{author}{Paramesran, R.}, \bibinfo{author}{Li, H.}, \bibinfo{year}{2015}.
\newblock \bibinfo{title}{Visual quality evaluation of image object segmentation: Subjective assessment and objective metric}.
\newblock \bibinfo{journal}{IEEE Transactions on Image Processing} \bibinfo{volume}{24}, \bibinfo{pages}{1--1}.
\newblock \DOIprefix\doi{10.1109/TIP.2015.2473099}.
\bibitem[{Strasburger et~al.(2018)Strasburger, Huber and Rose}]{Strasburger18}
\bibinfo{author}{Strasburger, H.}, \bibinfo{author}{Huber, J.}, \bibinfo{author}{Rose, D.}, \bibinfo{year}{2018}.
\newblock \bibinfo{title}{Ewald hering’s (1899) on the limits of visual acuity: A translation and commentary: With a supplement on alfred volkmann’s (1863) physiological investigations in the field of optics}.
\newblock \bibinfo{journal}{i-Perception} \bibinfo{volume}{9}, \bibinfo{pages}{204166951876367}.
\newblock \DOIprefix\doi{10.1177/2041669518763675}.
\bibitem[{Teghtsoonian(1971)}]{umbral_escala}
\bibinfo{author}{Teghtsoonian, R.}, \bibinfo{year}{1971}.
\newblock \bibinfo{title}{On the exponents in stevens' law and the constant in ekman's law}.
\newblock \bibinfo{journal}{Psychological review} \DOIprefix\doi{10.1037/h0030300}.
\bibitem[{Wallis and Bex(2011)}]{pf_naturalscenes}
\bibinfo{author}{Wallis, T.S.A.}, \bibinfo{author}{Bex, P.J.}, \bibinfo{year}{2011}.
\newblock \bibinfo{title}{Image correlates of crowding in natural scenes}.
\newblock \bibinfo{journal}{Journal of vision} \bibinfo{volume}{12 7}.
\newblock \URLprefix \url{https://api.semanticscholar.org/CorpusID:20325077}.
\bibitem[{Wang et~al.(2024)Wang, Zhang, Wei, Chen and Zhao}]{ICV2}
\bibinfo{author}{Wang, M.}, \bibinfo{author}{Zhang, K.}, \bibinfo{author}{Wei, H.}, \bibinfo{author}{Chen, W.}, \bibinfo{author}{Zhao, T.}, \bibinfo{year}{2024}.
\newblock \bibinfo{title}{Underwater image quality optimization: Researches, challenges, and future trends}.
\newblock \bibinfo{journal}{Image and Vision Computing} \bibinfo{volume}{146}.
\newblock \DOIprefix\doi{https://doi.org/10.1016/j.imavis.2024.104995.}
\bibitem[{Wang et~al.(2004)Wang, Bovik, Sheikh and Simoncelli}]{SSIM}
\bibinfo{author}{Wang, Z.}, \bibinfo{author}{Bovik, A.}, \bibinfo{author}{Sheikh, H.}, \bibinfo{author}{Simoncelli, E.}, \bibinfo{year}{2004}.
\newblock \bibinfo{title}{Image quality assessment: from error visibility to structural similarity}.
\newblock \bibinfo{journal}{IEEE Transactions on Image Processing} \bibinfo{volume}{13}, \bibinfo{pages}{600--612}.
\newblock \DOIprefix\doi{10.1109/TIP.2003.819861}.
\bibitem[{Wang and Bovik(2009)}]{MSE_Loveit}
\bibinfo{author}{Wang, Z.}, \bibinfo{author}{Bovik, A.C.}, \bibinfo{year}{2009}.
\newblock \bibinfo{title}{Mean squared error: Love it or leave it? a new look at signal fidelity measures}.
\newblock \bibinfo{journal}{IEEE Signal Processing Magazine} \bibinfo{volume}{26}, \bibinfo{pages}{98--117}.
\newblock \DOIprefix\doi{10.1109/MSP.2008.930649}.
\bibitem[{Wang and Simoncelli(2005)}]{Wang&Sim}
\bibinfo{author}{Wang, Z.}, \bibinfo{author}{Simoncelli, E.}, \bibinfo{year}{2005}.
\newblock \bibinfo{title}{An adaptive linear system framework for image distortion analysis}, in: \bibinfo{booktitle}{IEEE International Conference on Image Processing 2005}, pp. \bibinfo{pages}{III--1160}.
\newblock \DOIprefix\doi{10.1109/ICIP.2005.1530603}.
\bibitem[{Webster(2011)}]{scale_const}
\bibinfo{author}{Webster, M.}, \bibinfo{year}{2011}.
\newblock \bibinfo{title}{Adaptation and visual coding}.
\newblock \bibinfo{journal}{Journal of vision} \bibinfo{volume}{11}.
\newblock \DOIprefix\doi{10.1167/11.5.3}.
\bibitem[{Wyszecki and Stiles(2000)}]{Stiles00}
\bibinfo{author}{Wyszecki, G.}, \bibinfo{author}{Stiles, W.}, \bibinfo{year}{2000}.
\newblock \bibinfo{title}{Color science: Concepts and methods, quantitative data and formulae, 2nd edition}.
\newblock \bibinfo{journal}{Color Science: Concepts and Methods, Quantitative Data and Formulae, 2nd Edition, by Gunther Wyszecki, W. S. Stiles, pp. 968. ISBN 0-471-39918-3. Wiley-VCH , July 2000.} .
\bibitem[{Zhang et~al.(2018)Zhang, Isola, Efros, Shechtman and Wang}]{LPIPS}
\bibinfo{author}{Zhang, R.}, \bibinfo{author}{Isola, P.}, \bibinfo{author}{Efros, A.A.}, \bibinfo{author}{Shechtman, E.}, \bibinfo{author}{Wang, O.}, \bibinfo{year}{2018}.
\newblock \bibinfo{title}{The unreasonable effectiveness of deep features as a perceptual metric}.
\newblock \bibinfo{journal}{2018 IEEE/CVF Conference on Computer Vision and Pattern Recognition} , \bibinfo{pages}{586--595}\URLprefix \url{https://api.semanticscholar.org/CorpusID:4766599}.

\end{thebibliography}

\appendix
\section{Appendix A.}
\label{app:apndA}

 The following figures show examples from the TID13 database sorted by DMOS for which the reader can make a first estimate of the threshold value of human invisibility for the DMOS, i.e., a first estimation of $\mathcal{D}_\tau$. For each reference image (at the top) some distorted images are selected, sorted in increasing order by the DMOS value. With these images, it can be tested, in a rather coarse way, which DMOS value tells us what is invisible and what is not to the observer. Roughly, the value is around 0.45, since the second image is usually indistinguishable from the original but the third one always differs. 

    \begin{figure*}[!h]
			\begin{center}
                \vspace{-1cm}
			\includegraphics[width=1\linewidth]{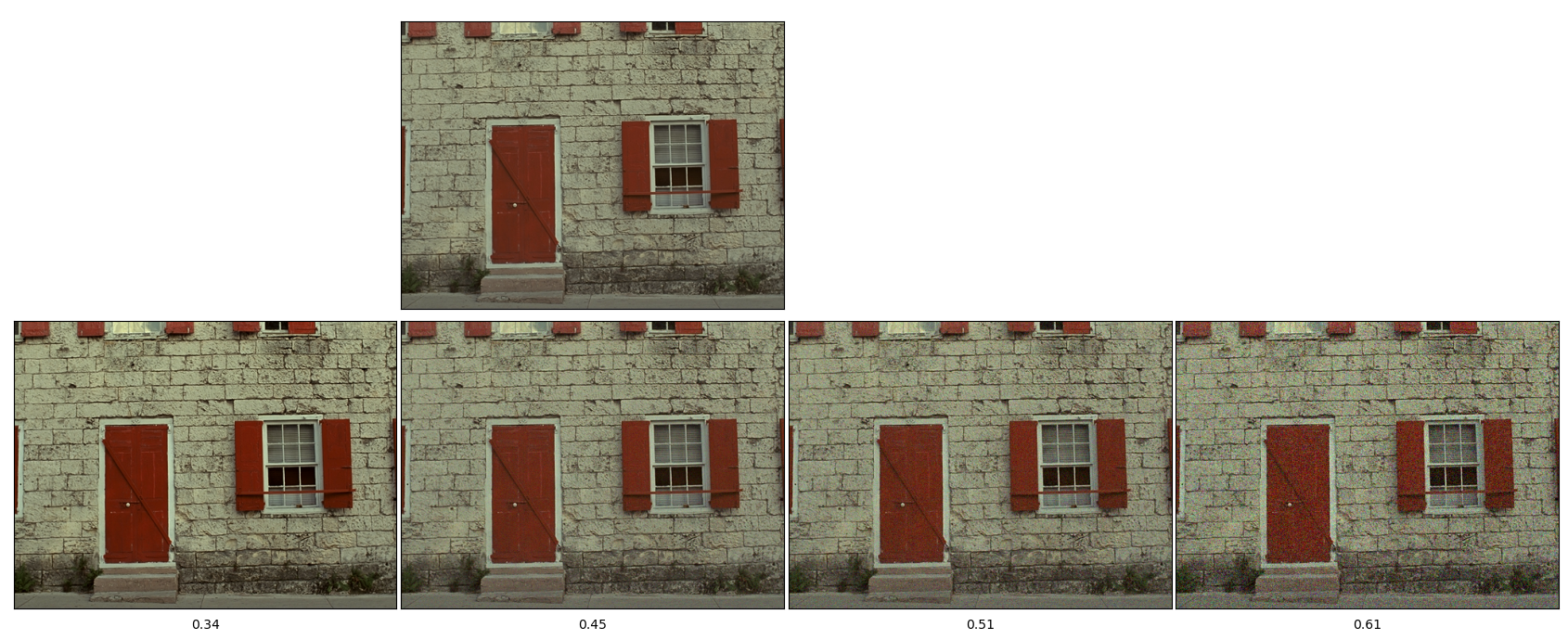}\\
   \includegraphics[width=1\linewidth]{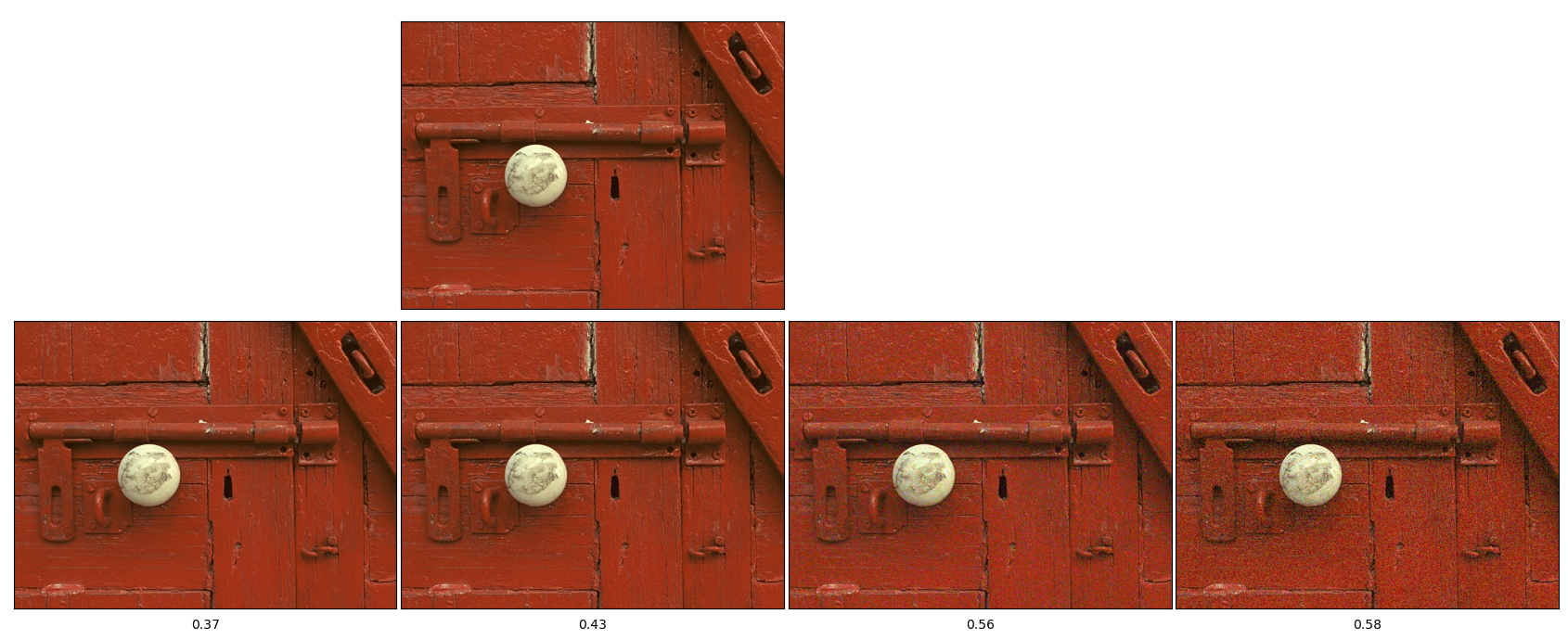}\\
   \includegraphics[width=1\linewidth]{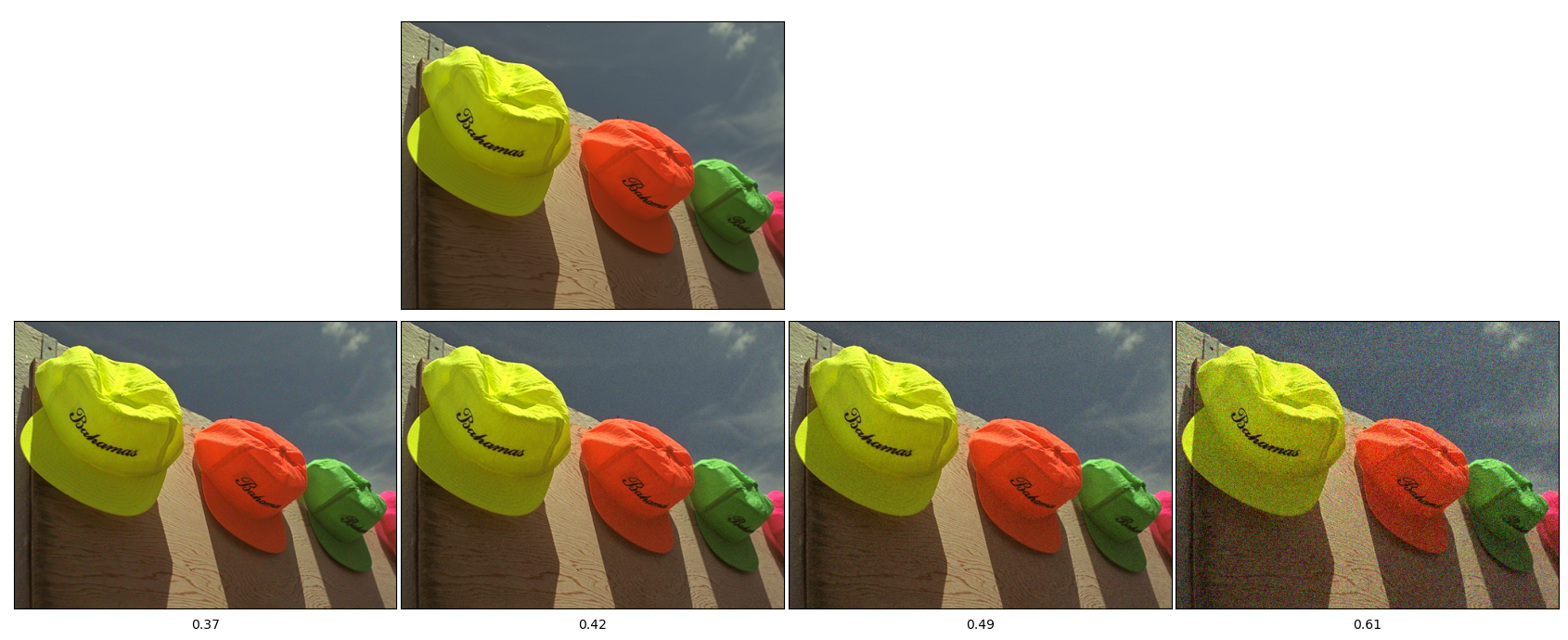}
   \end{center}
    \end{figure*}

    \begin{figure*}[!h]
			\begin{center}
                \vspace{-1cm}
			\includegraphics[width=1\linewidth]{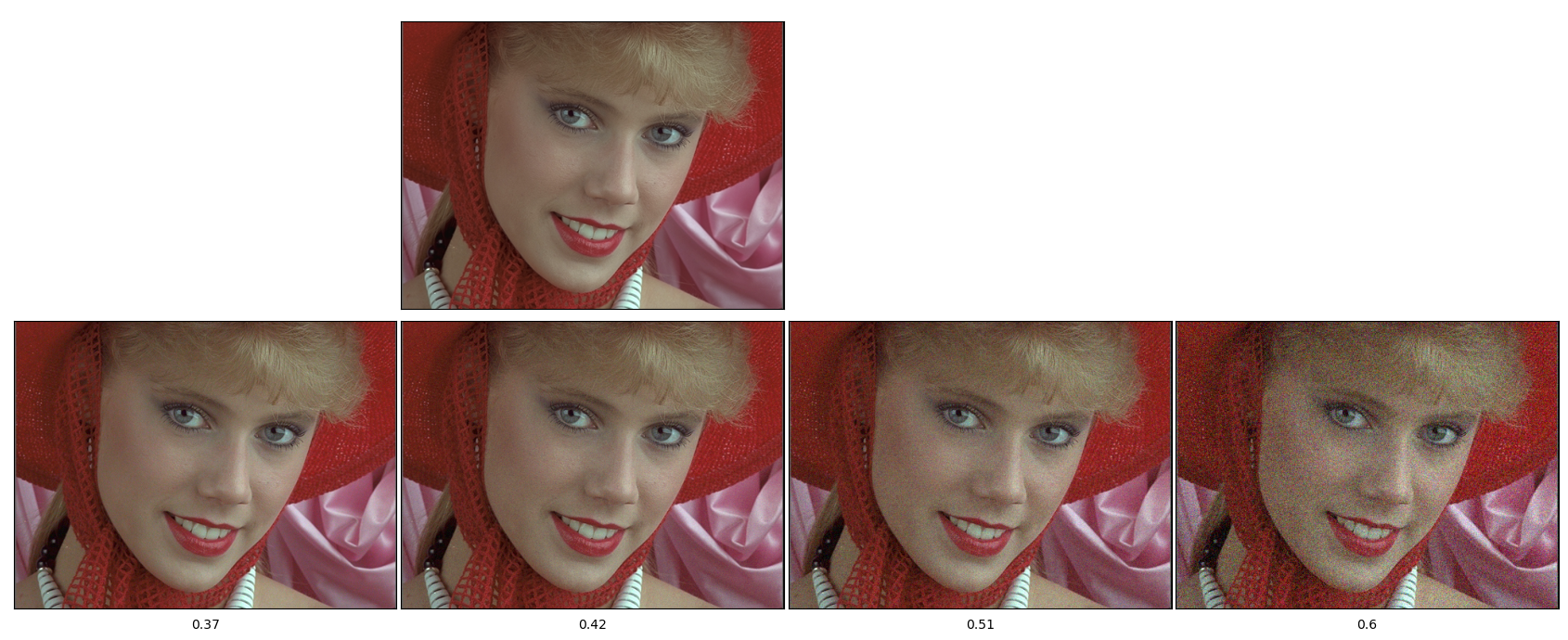}\\
   \includegraphics[width=1\linewidth]{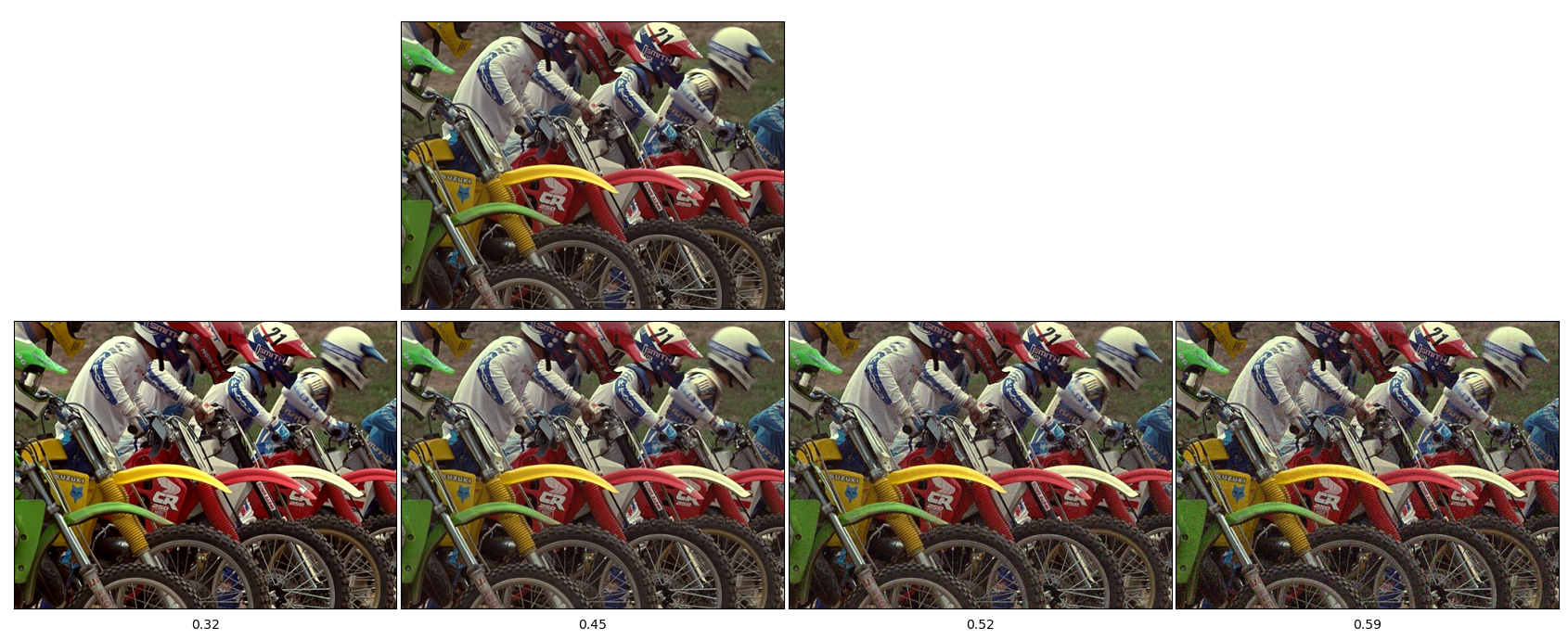}\\
   \includegraphics[width=1\linewidth]{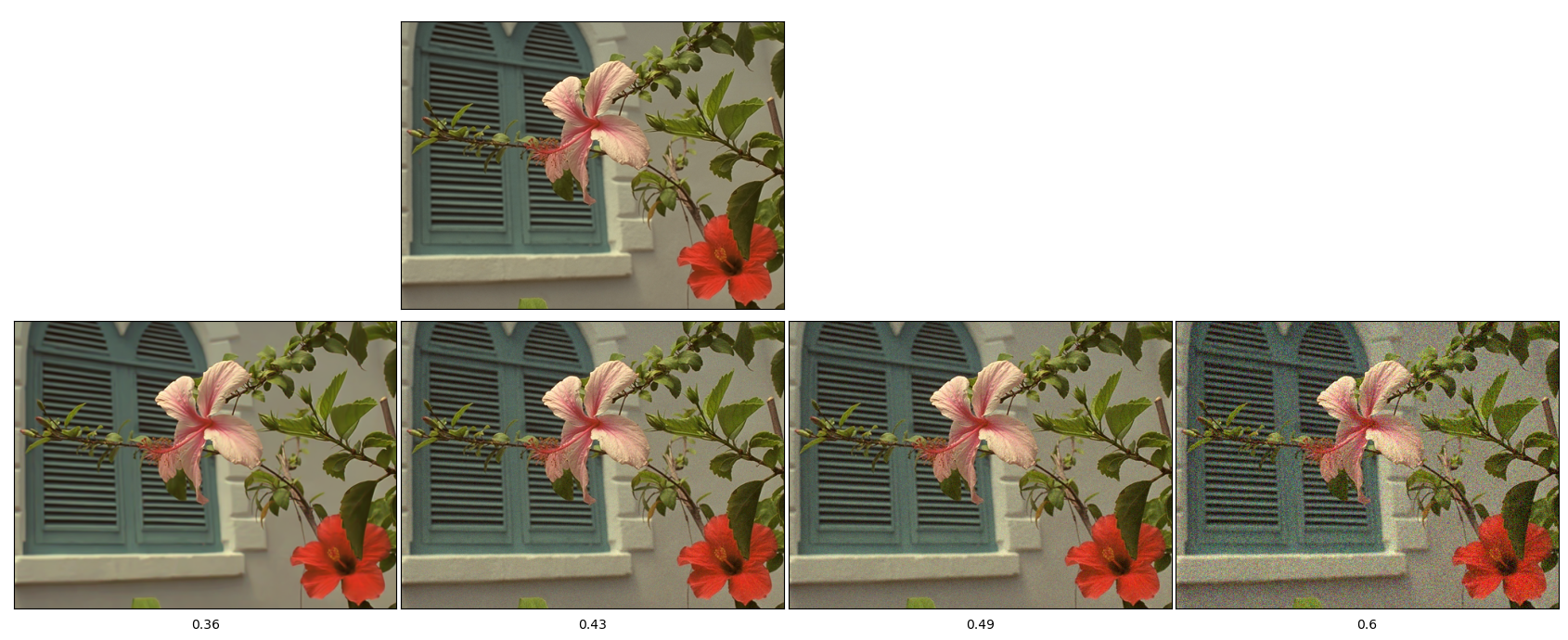}
   \end{center}
    \end{figure*}

\begin{figure*}[!h]
			\begin{center}
                \vspace{-1cm}
			\includegraphics[width=1\linewidth]{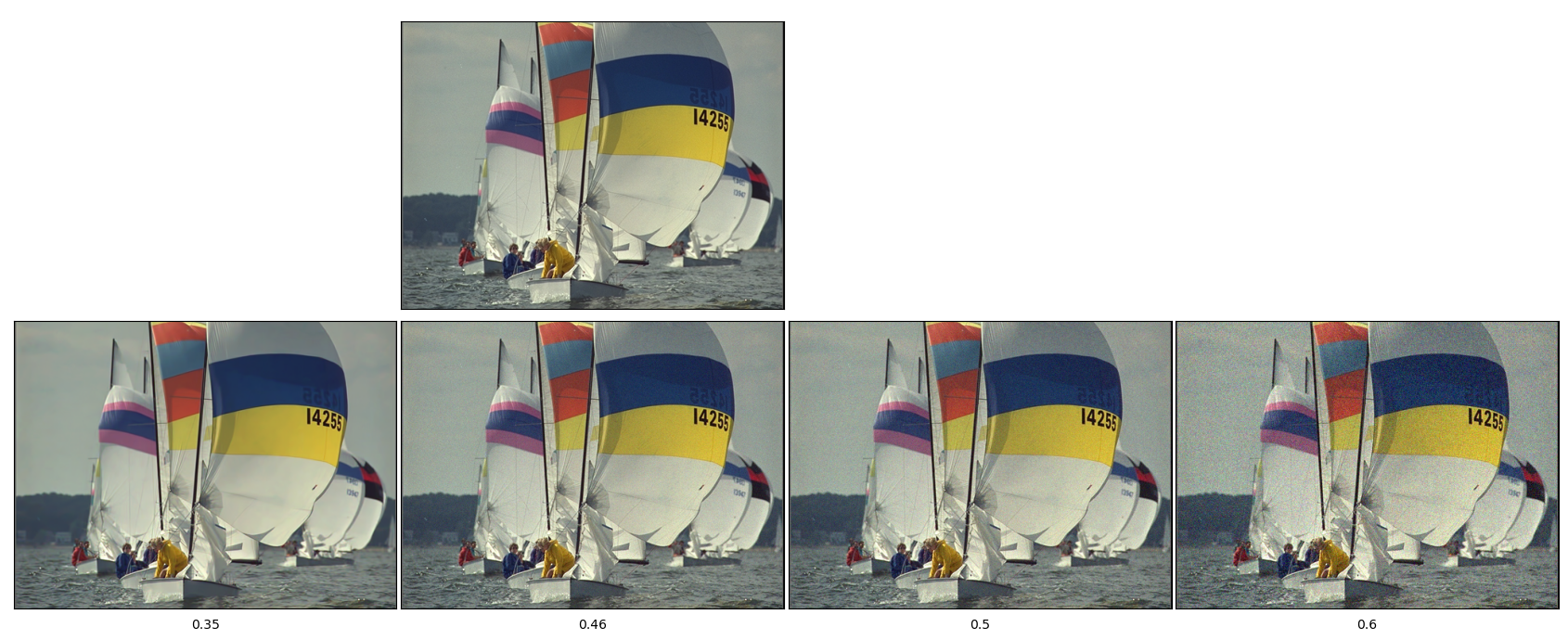}\\
   \includegraphics[width=1\linewidth]{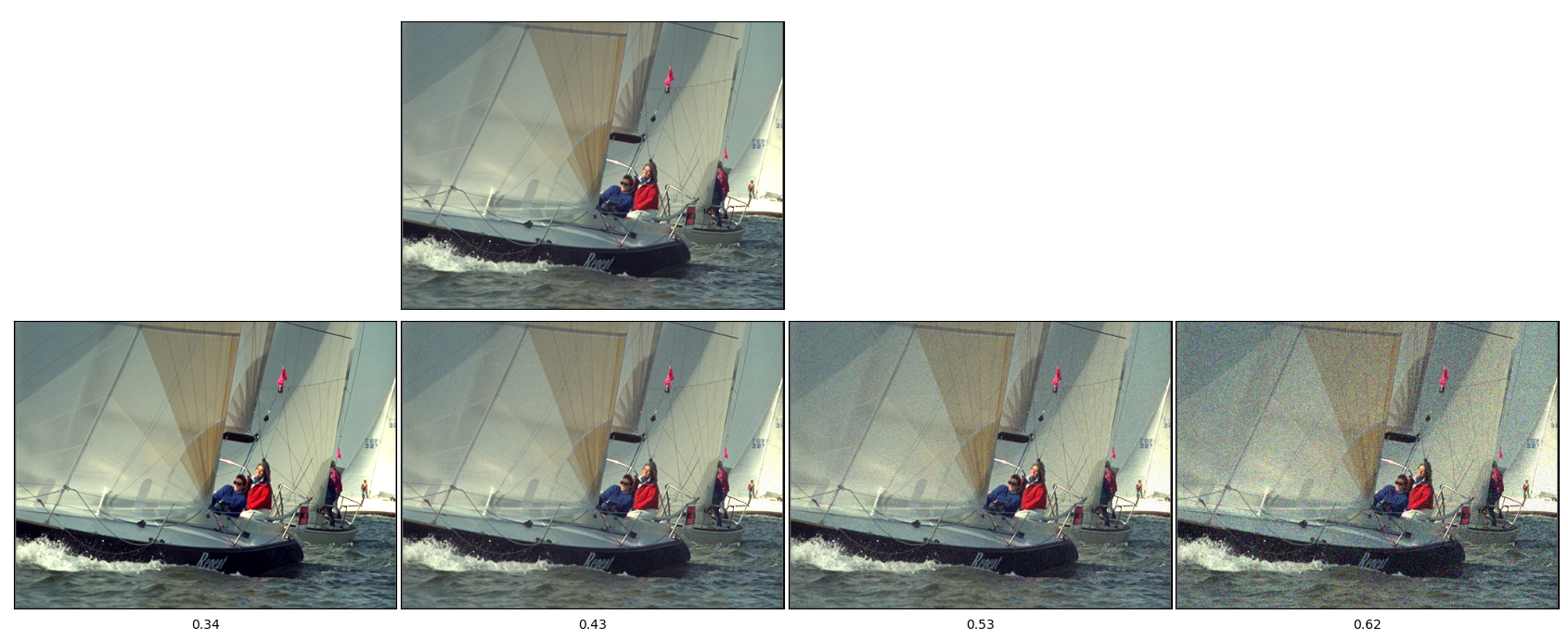}\\
   \includegraphics[width=1\linewidth]{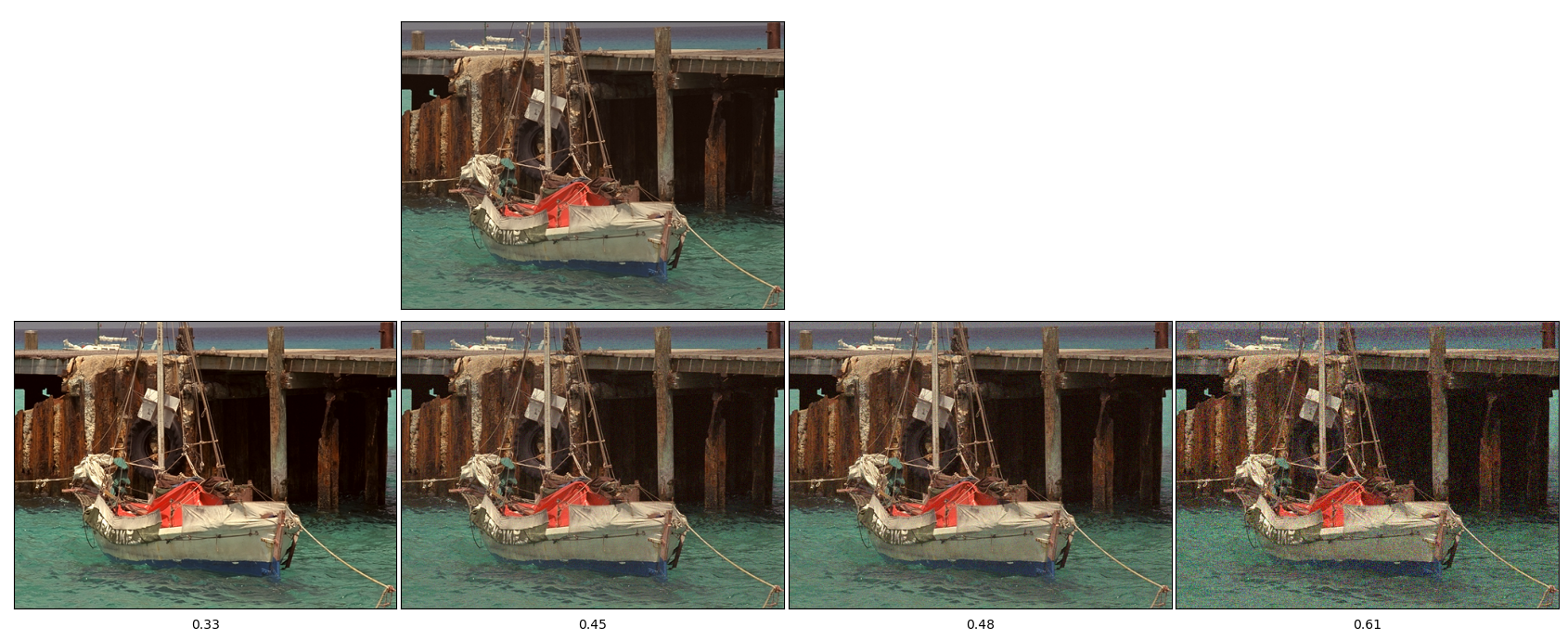}
   \end{center}
    \end{figure*}

    \begin{figure*}[!h]
			\begin{center}
                \vspace{-1cm}
			\includegraphics[width=1\linewidth]{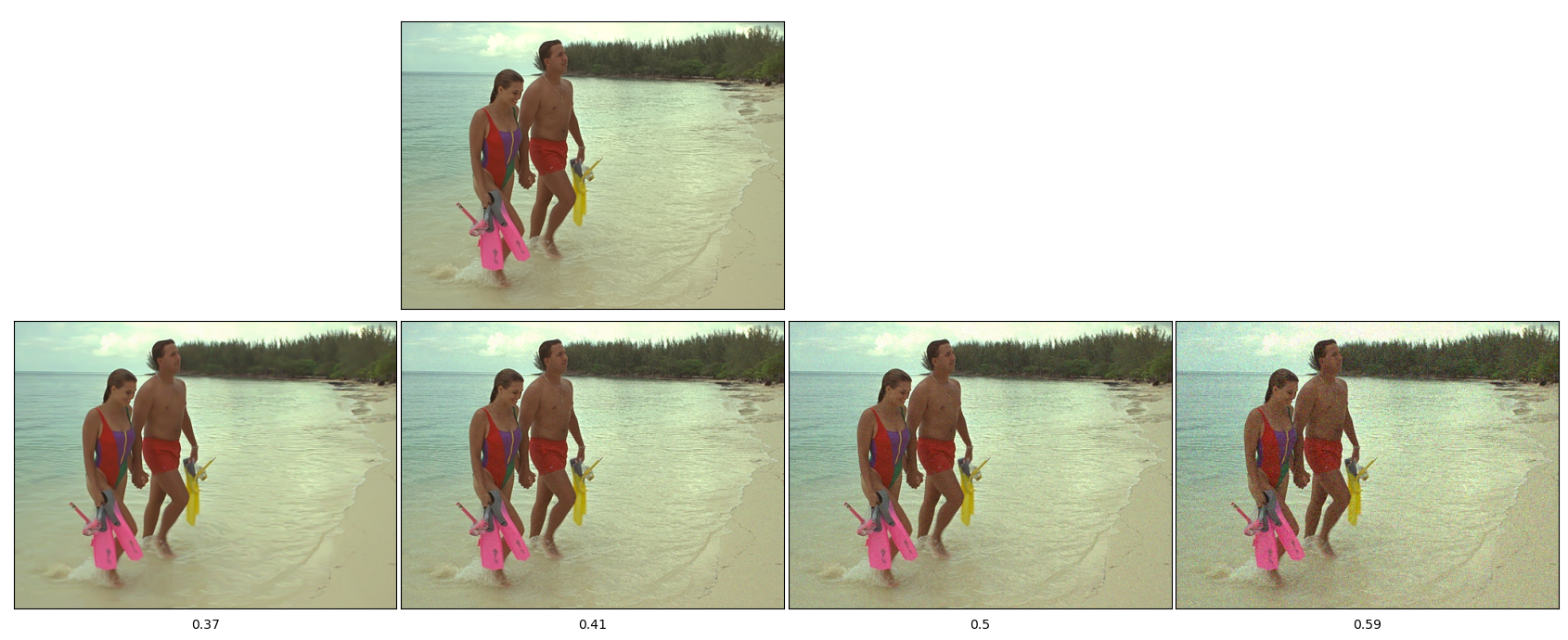}\\
   \includegraphics[width=1\linewidth]{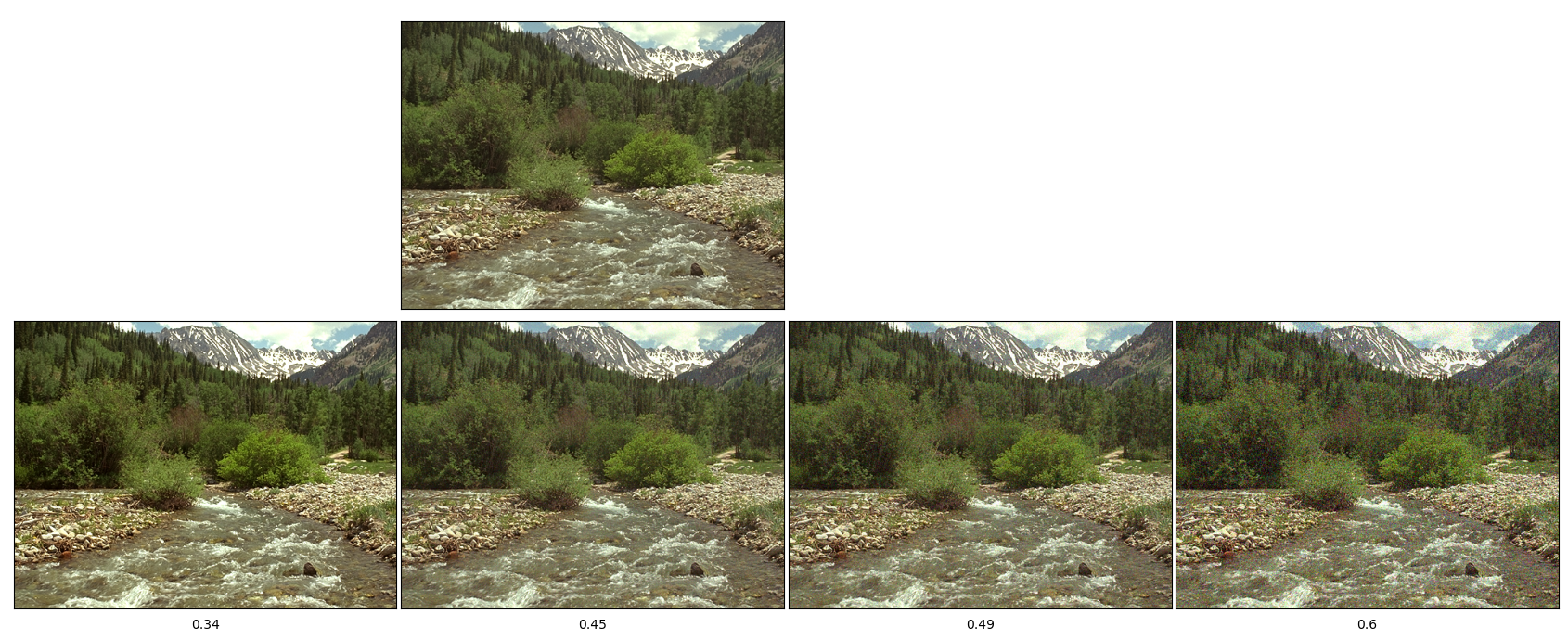}\\
   \includegraphics[width=1\linewidth]{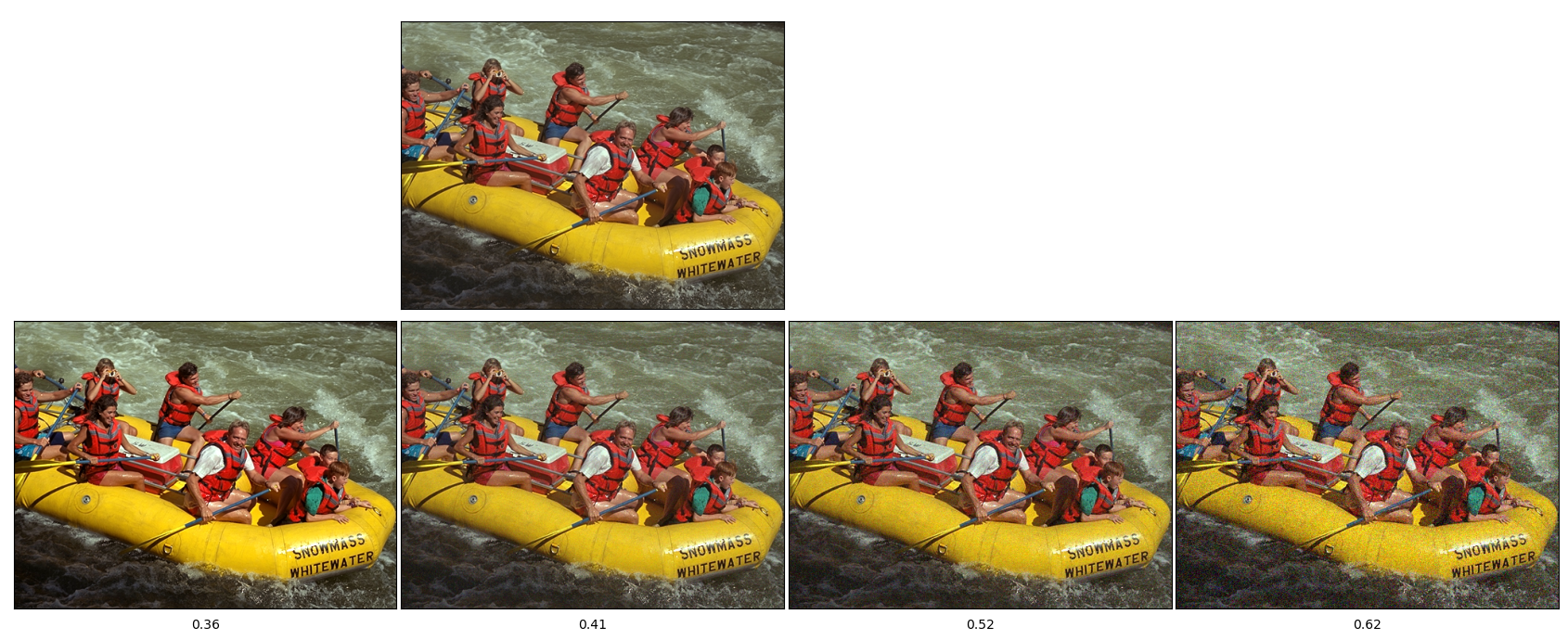}
   \end{center}
    \end{figure*}

        \begin{figure*}[!h]
			\begin{center}
                \vspace{-1cm}
			\includegraphics[width=1\linewidth]{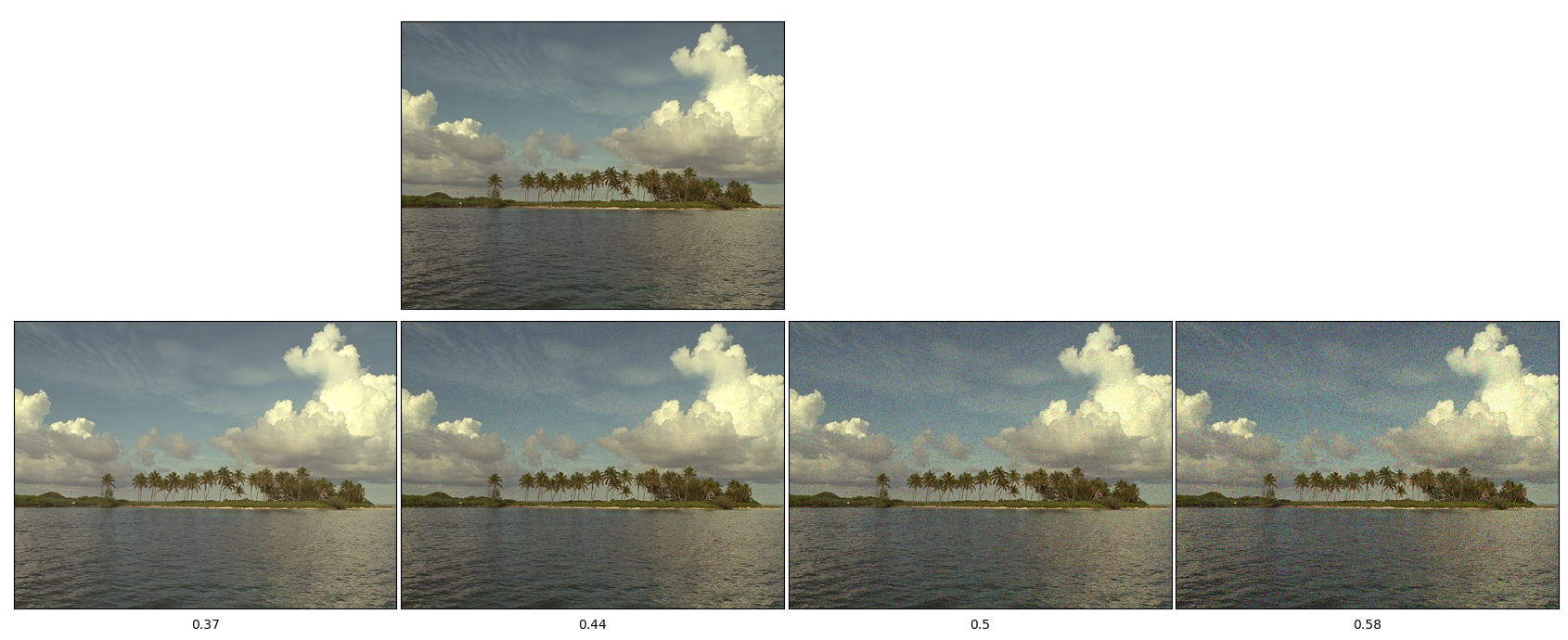}\\
   \includegraphics[width=1\linewidth]{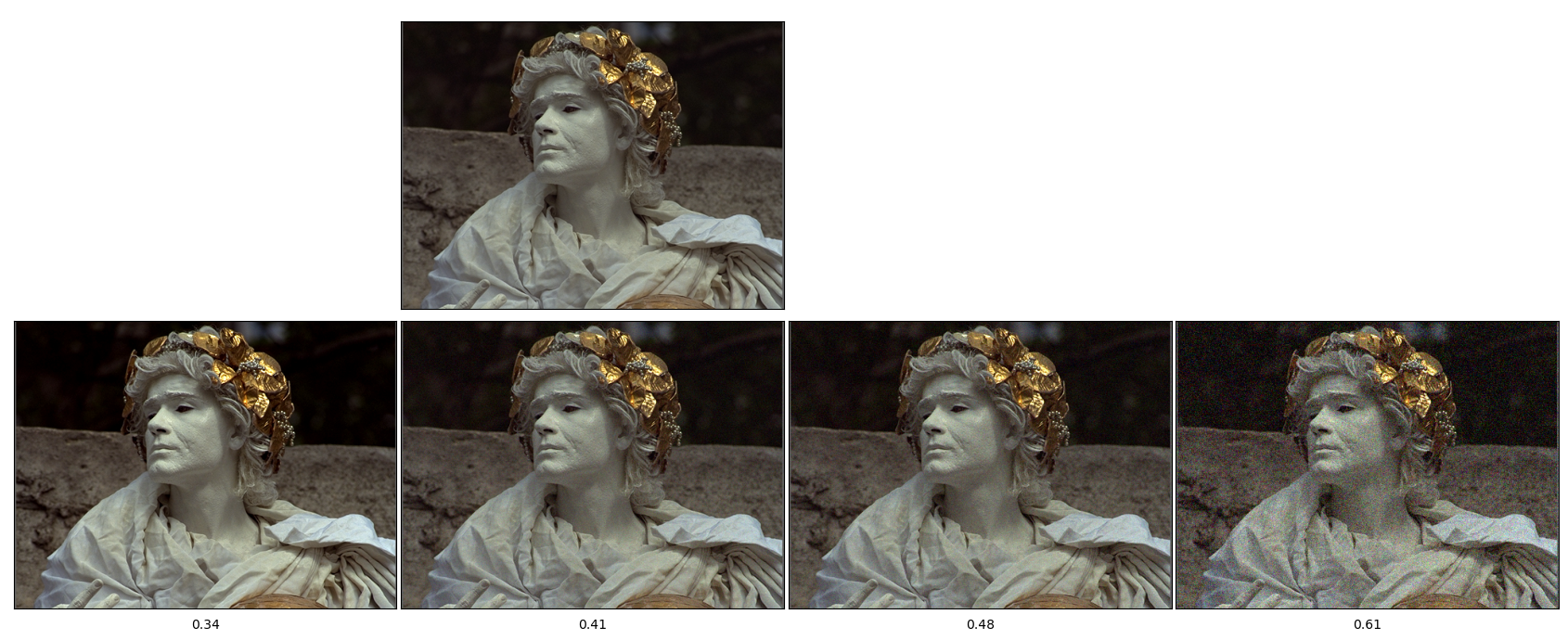}\\
   \includegraphics[width=1\linewidth]{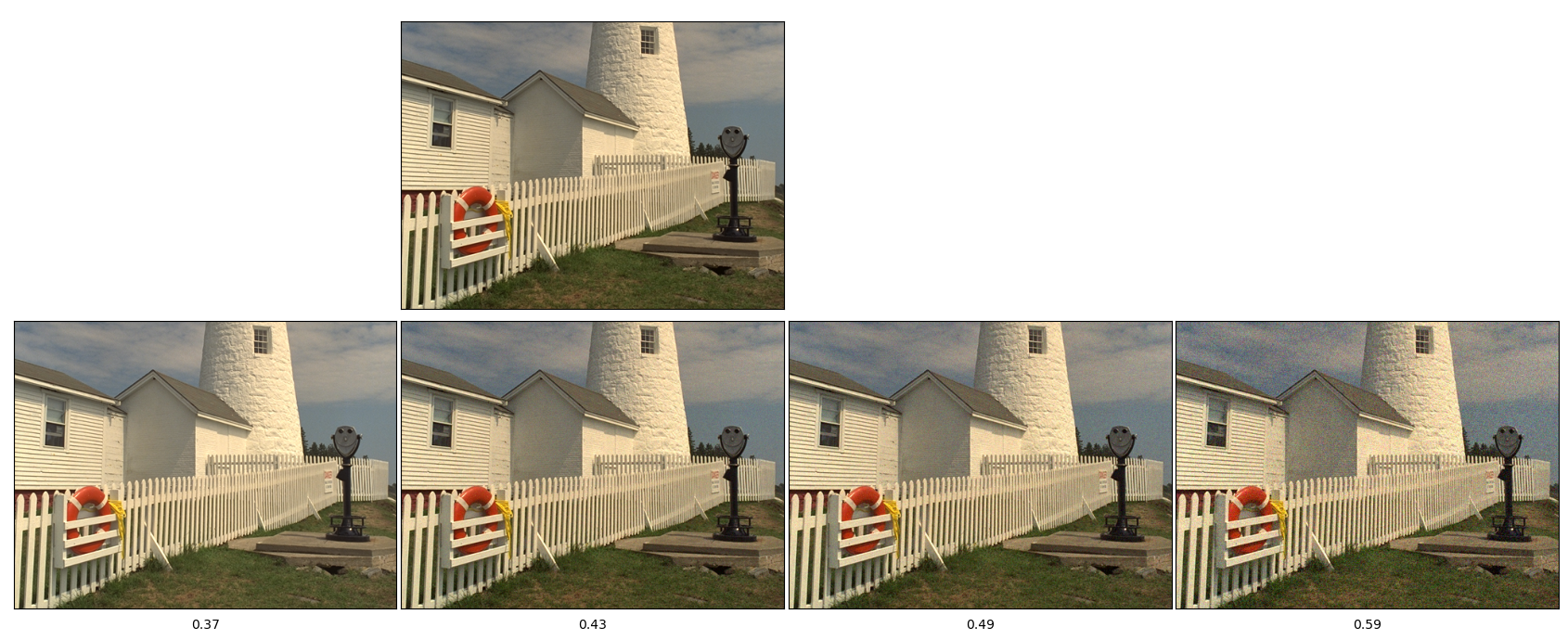}
   \end{center}
    \end{figure*}

    \begin{figure*}[!h]
			\begin{center}
                \vspace{-1cm}
			\includegraphics[width=1\linewidth]{Images/im20.png}\\
   \includegraphics[width=1\linewidth]{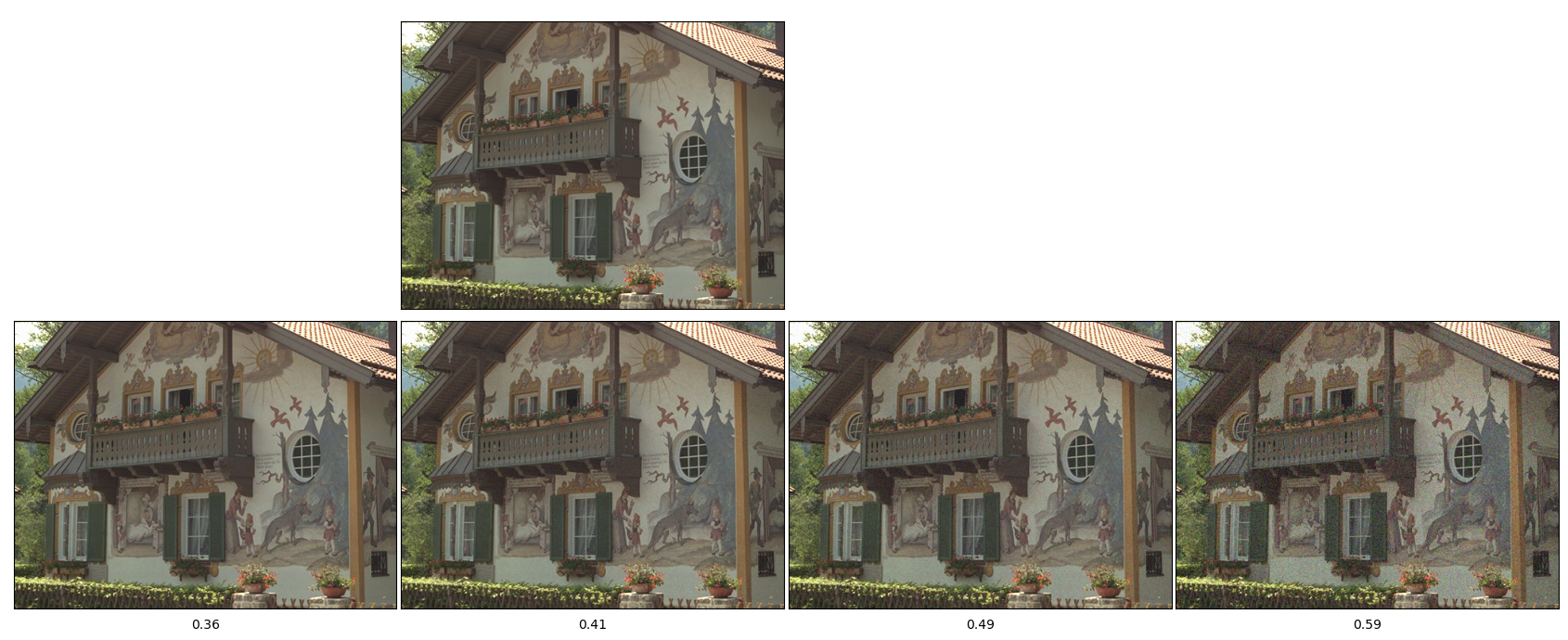}\\
   \includegraphics[width=1\linewidth]{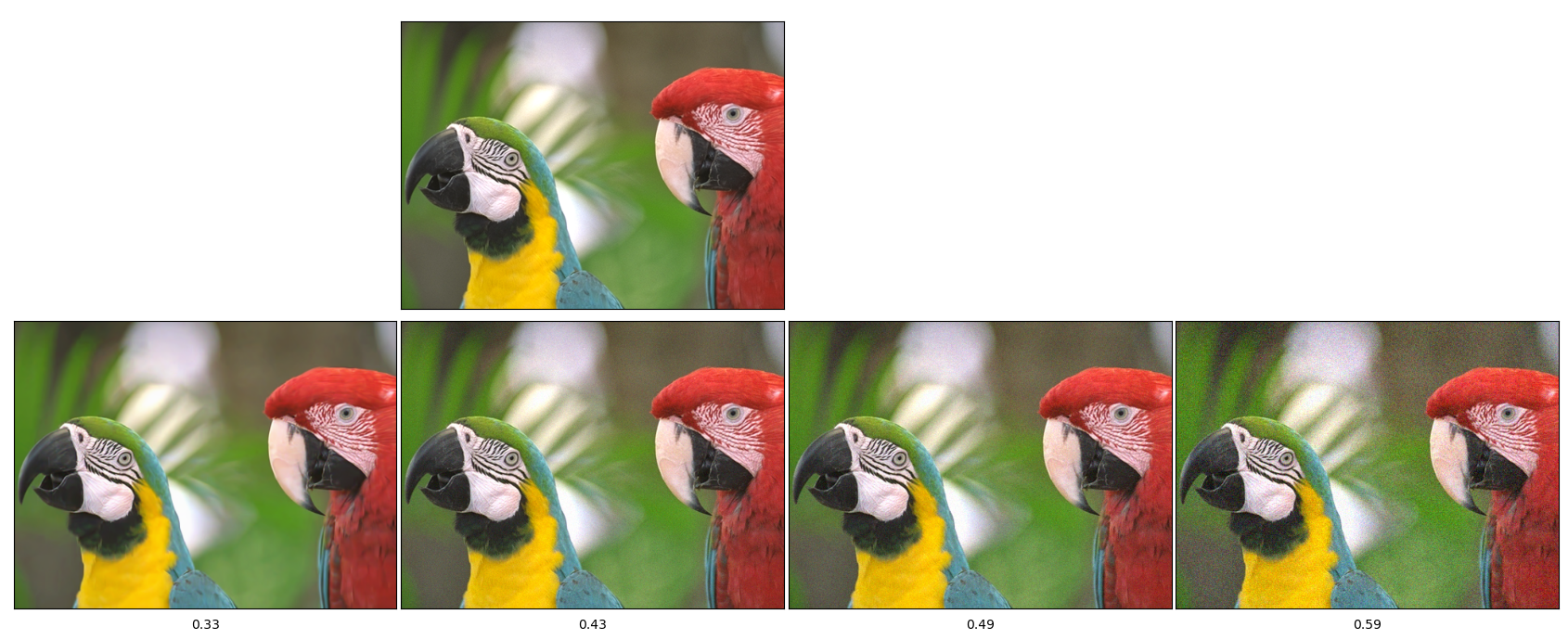}
   \end{center}
    \end{figure*}

\appendix
\section{Appendix B.}
\label{app:apndB}

\begin{figure*}[!h]
\centering
			\includegraphics[width=1\linewidth]{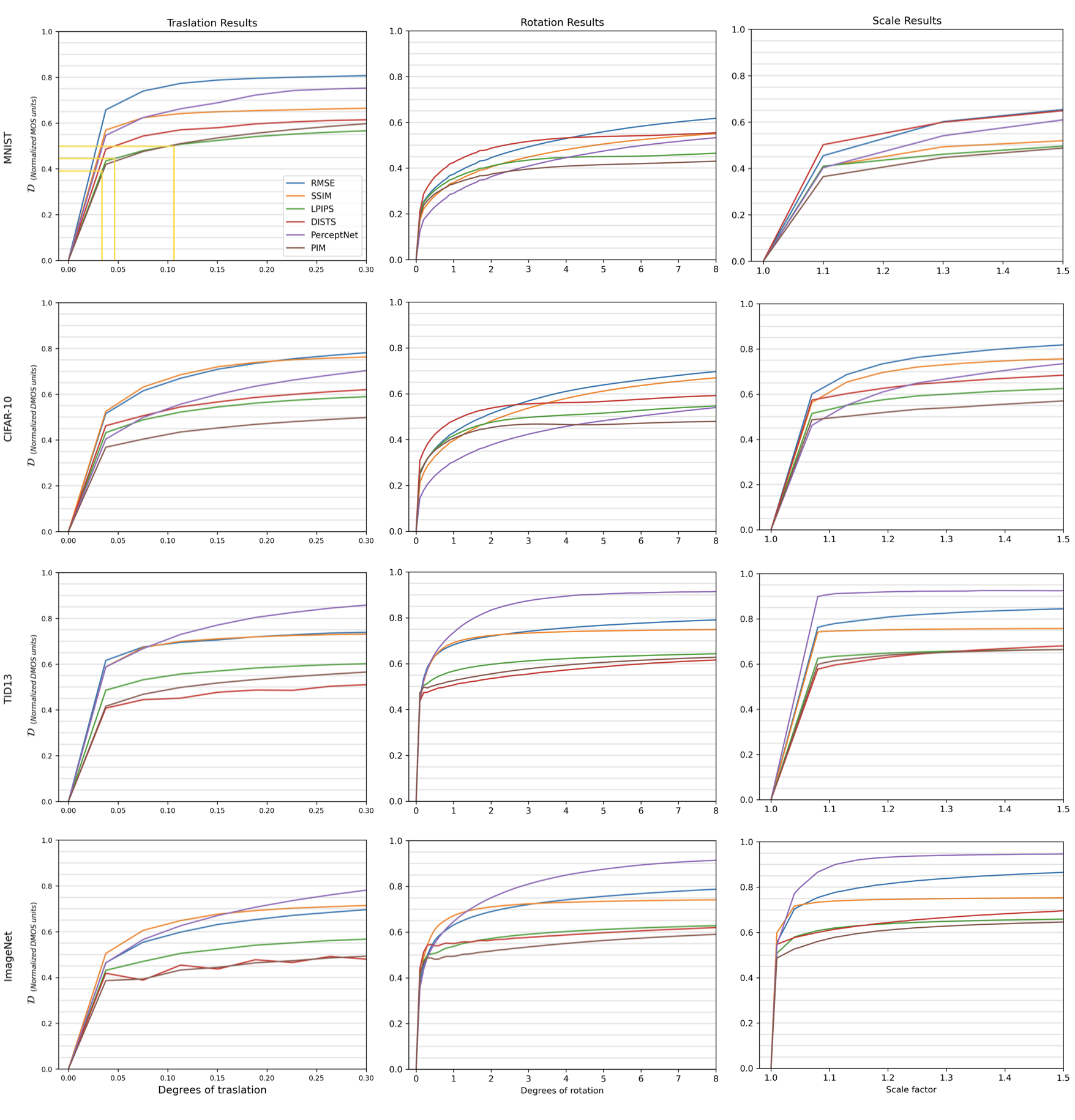}	
		
		\caption{Results of the different metrics in the experiments for the geometric affine transformations. On these graphs we will apply what we have seen in Figure~\ref{fig:Fig1} to create intervals to compare with the human thresholds.  In the different columns, we have the results differentiated by database (and in the rows, by affine transformation (translation, rotation and scale).}
		\label{fig:results_geo}
\end{figure*}

Figure~\ref{fig:results_geo} shows the results for the all databases for the geometric affine transformations (translation, rotation and scale). In addition, the first subgraph shows again the proposed methodology with the cuts on the y-axis, marked in gold color; it is the only one that also shows the legend and it is shared by all the others.

\appendix
\section{Appendix C.}
\label{app:apndC}

\begin{figure*}[!h]
		
		\begin{subfigure}{0.33\textwidth}
			\centering
			\includegraphics[width=1\linewidth]{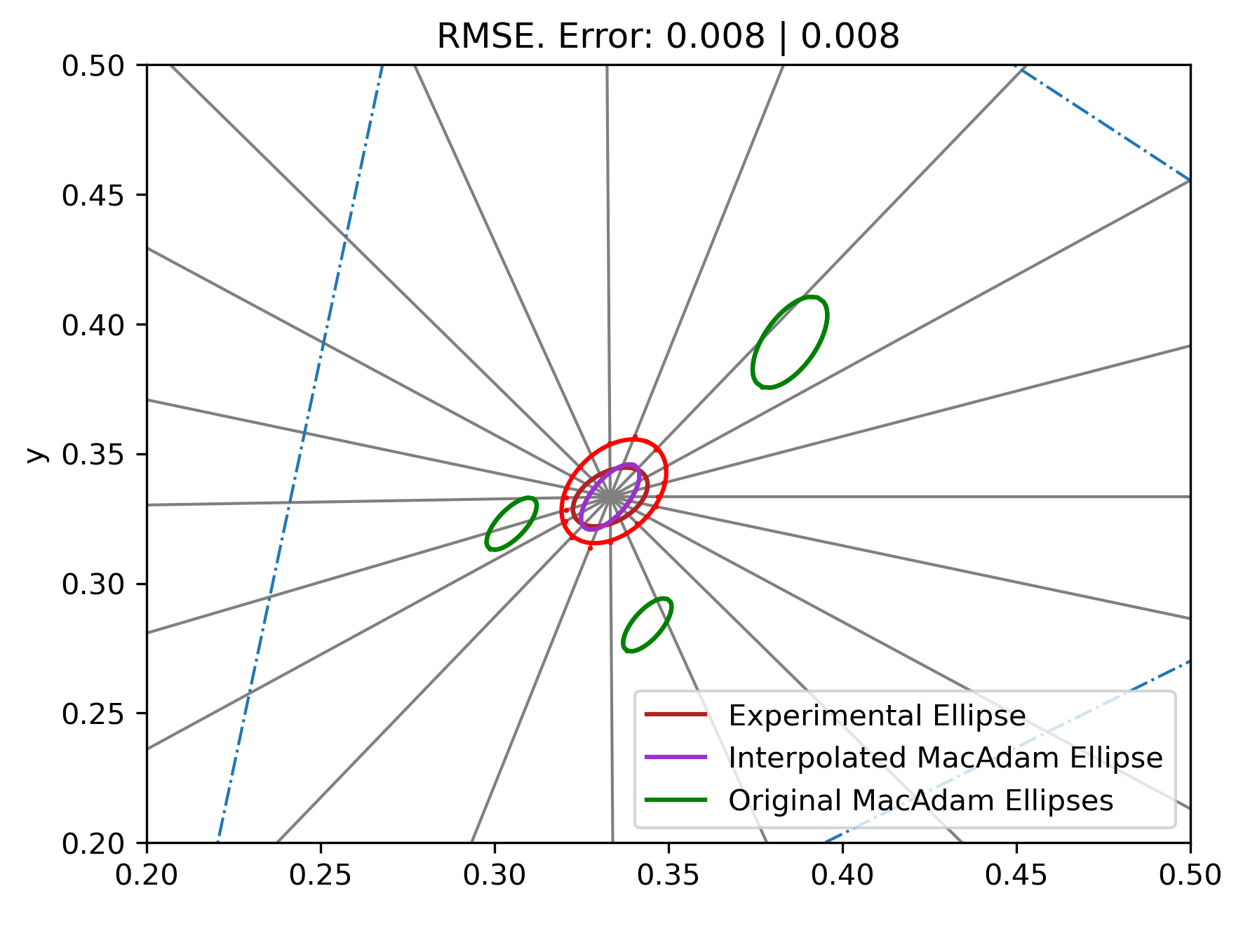}
		\end{subfigure}%
		\begin{subfigure}{0.33\textwidth}
			\centering
			\includegraphics[width=1\linewidth]{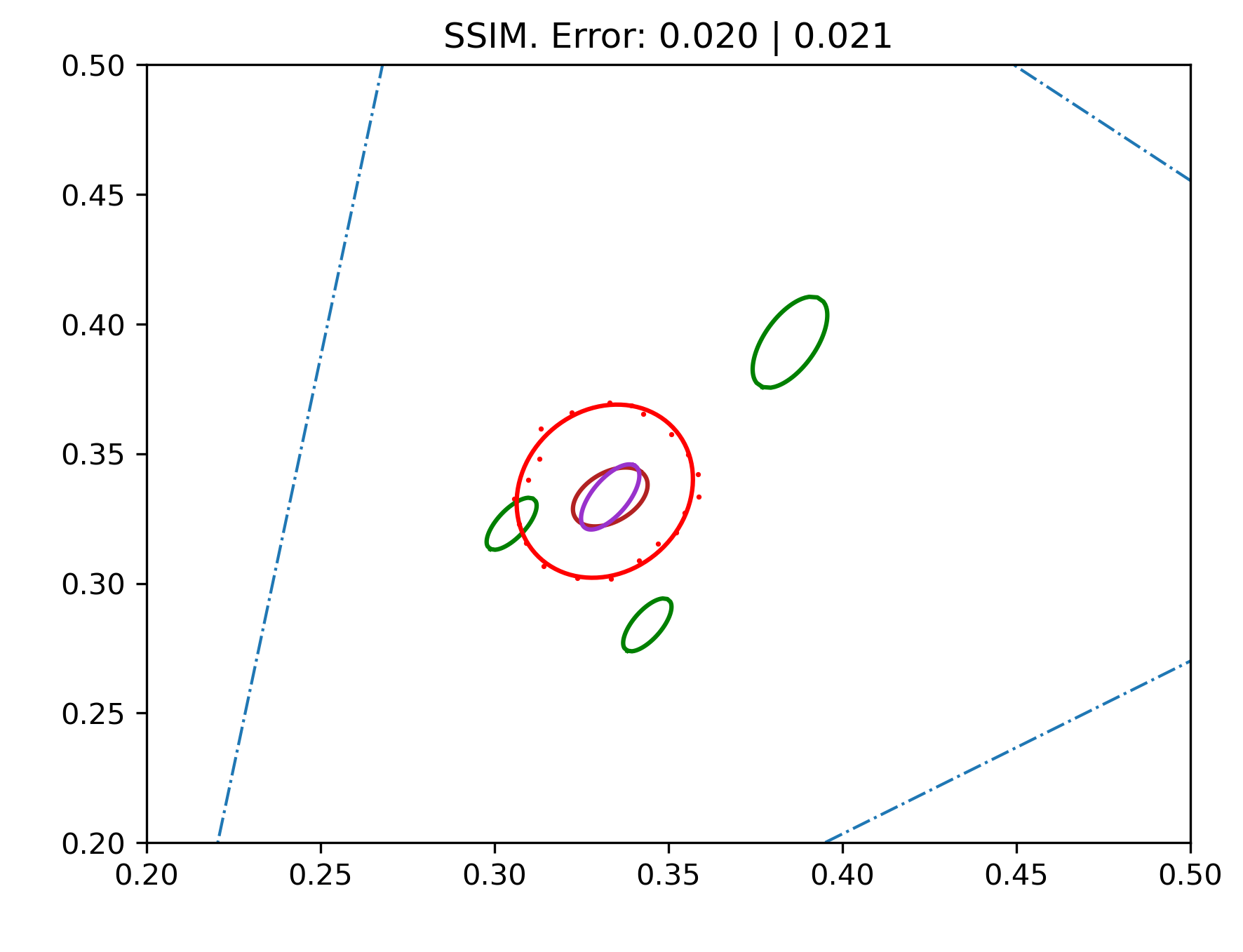}			
		\end{subfigure}
            \begin{subfigure}{0.33\textwidth}
			\centering
			\includegraphics[width=1\linewidth]{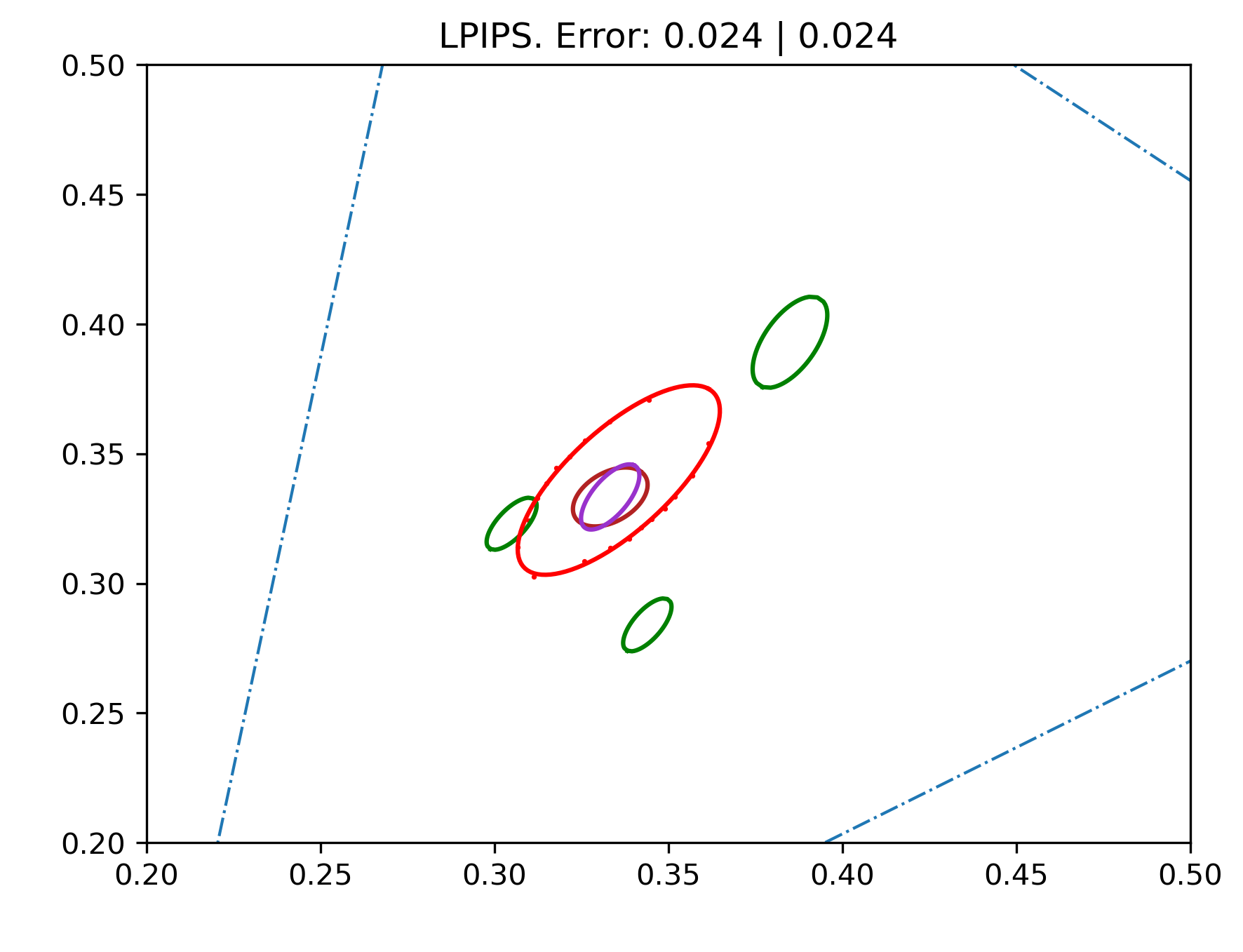}			
		\end{subfigure}

            \begin{subfigure}{0.33\textwidth}
			\centering
			\includegraphics[width=1\linewidth]{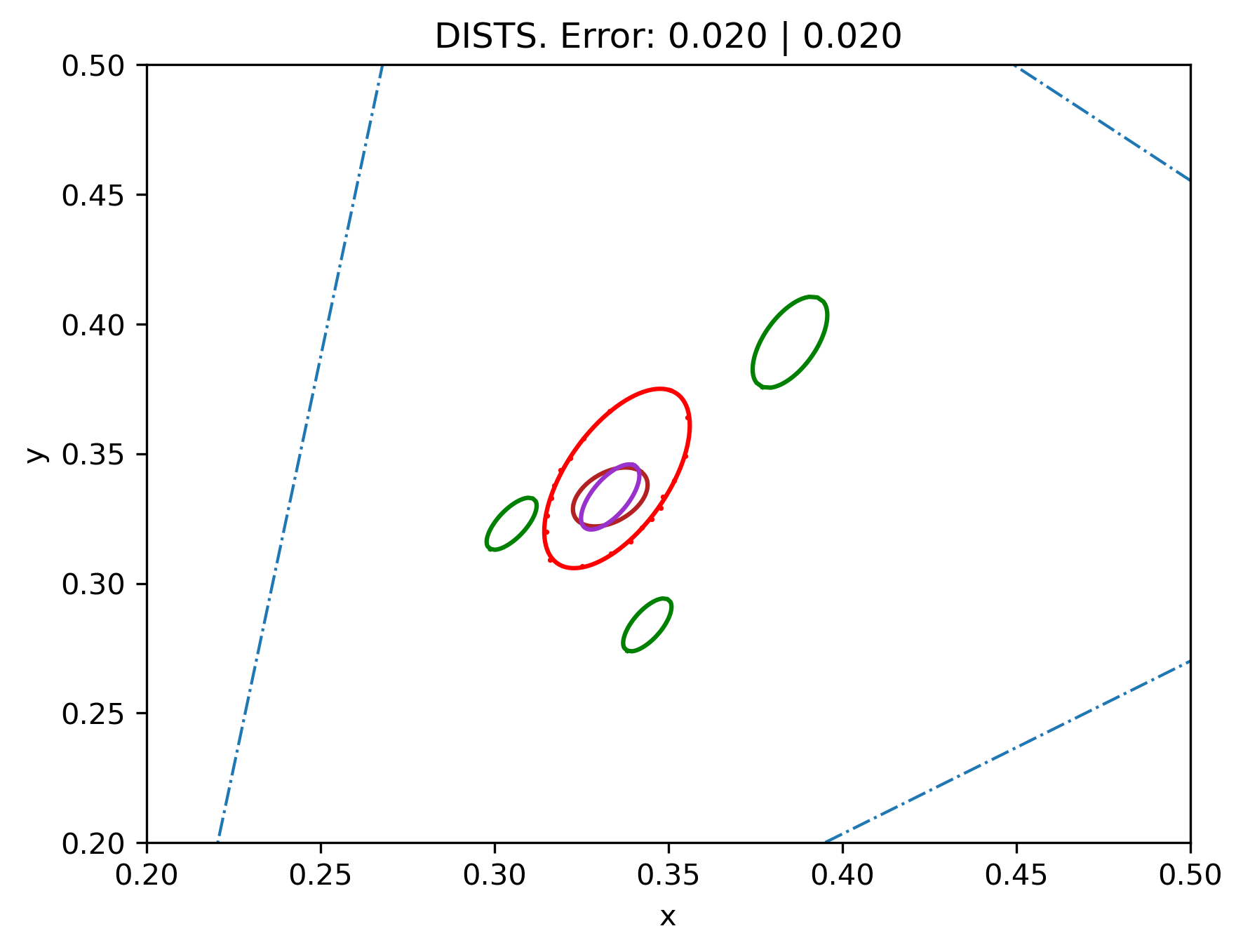}
		\end{subfigure}%
		\begin{subfigure}{0.33\textwidth}
			\centering
			\includegraphics[width=1\linewidth]{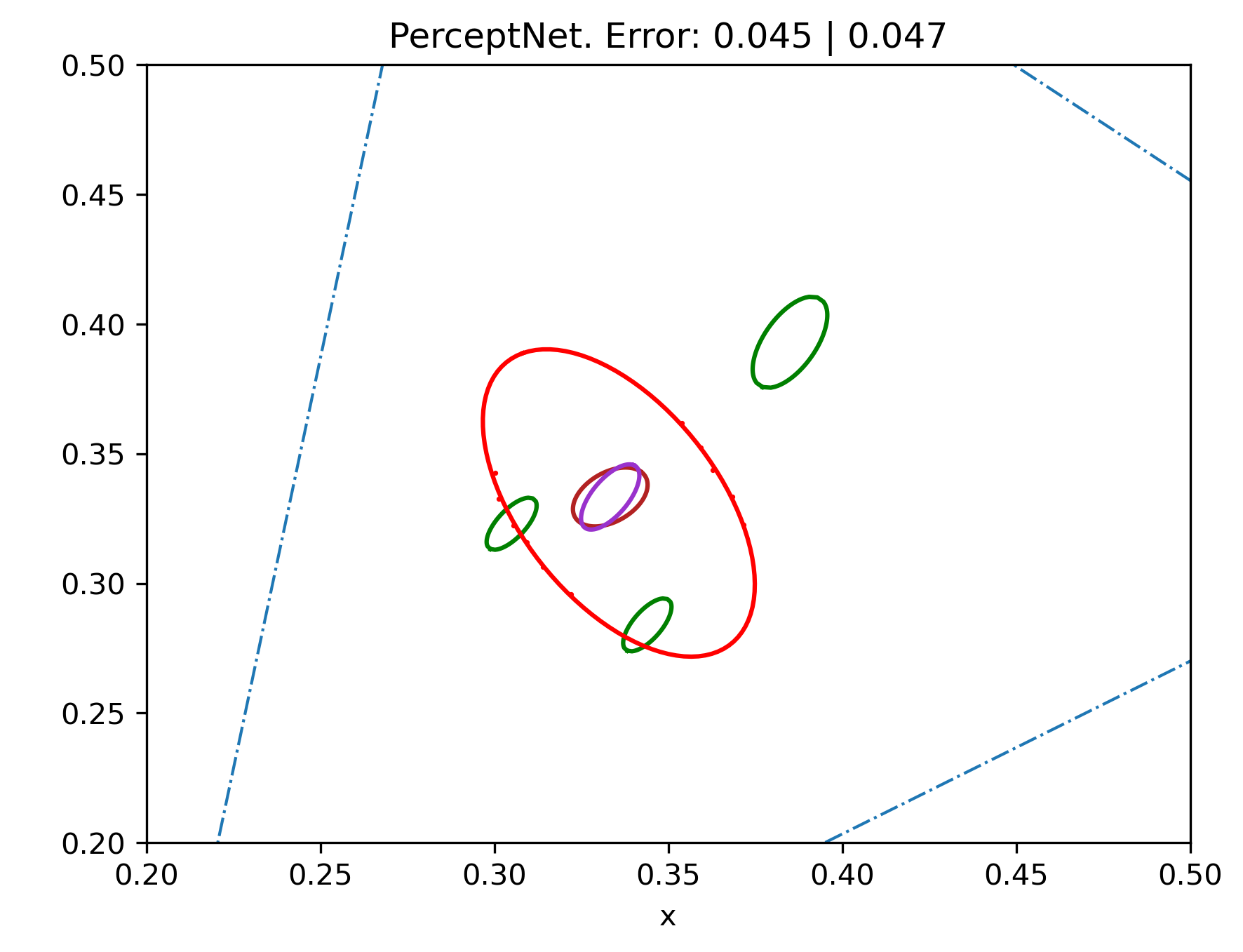}
		\end{subfigure}
            \begin{subfigure}{0.33\textwidth}
			\centering
			\includegraphics[width=1\linewidth]{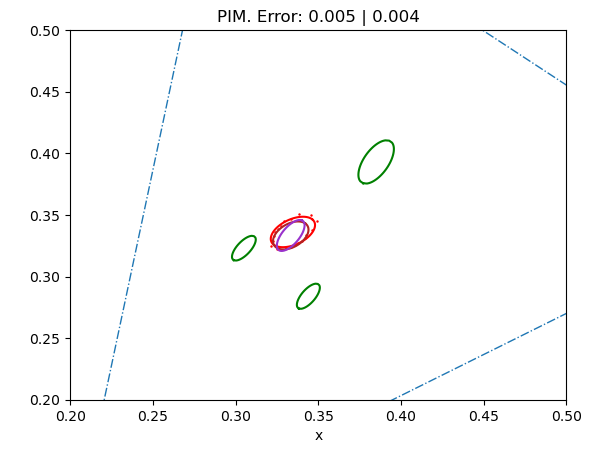}
		\end{subfigure}
            \label{fig:leaky}
		        \caption{Results of the different metrics in the color discrimination experiments, together with their error for ImageNet data. The error with respect to the fitted MacAdam ellipse is shown on the left whereas the error with respect to the experimentally obtained ellipse is shown on the right. The first subfigure also shows the direction of the 20 hues. The metric that makes the least error with respect to the MacAdam ellipses is PIM in both cases.}
		\label{fig:results_ilu_ima}
	\end{figure*}

\begin{figure*}[!h]
		
		\begin{subfigure}{0.33\textwidth}
			\centering
			\includegraphics[width=1\linewidth]{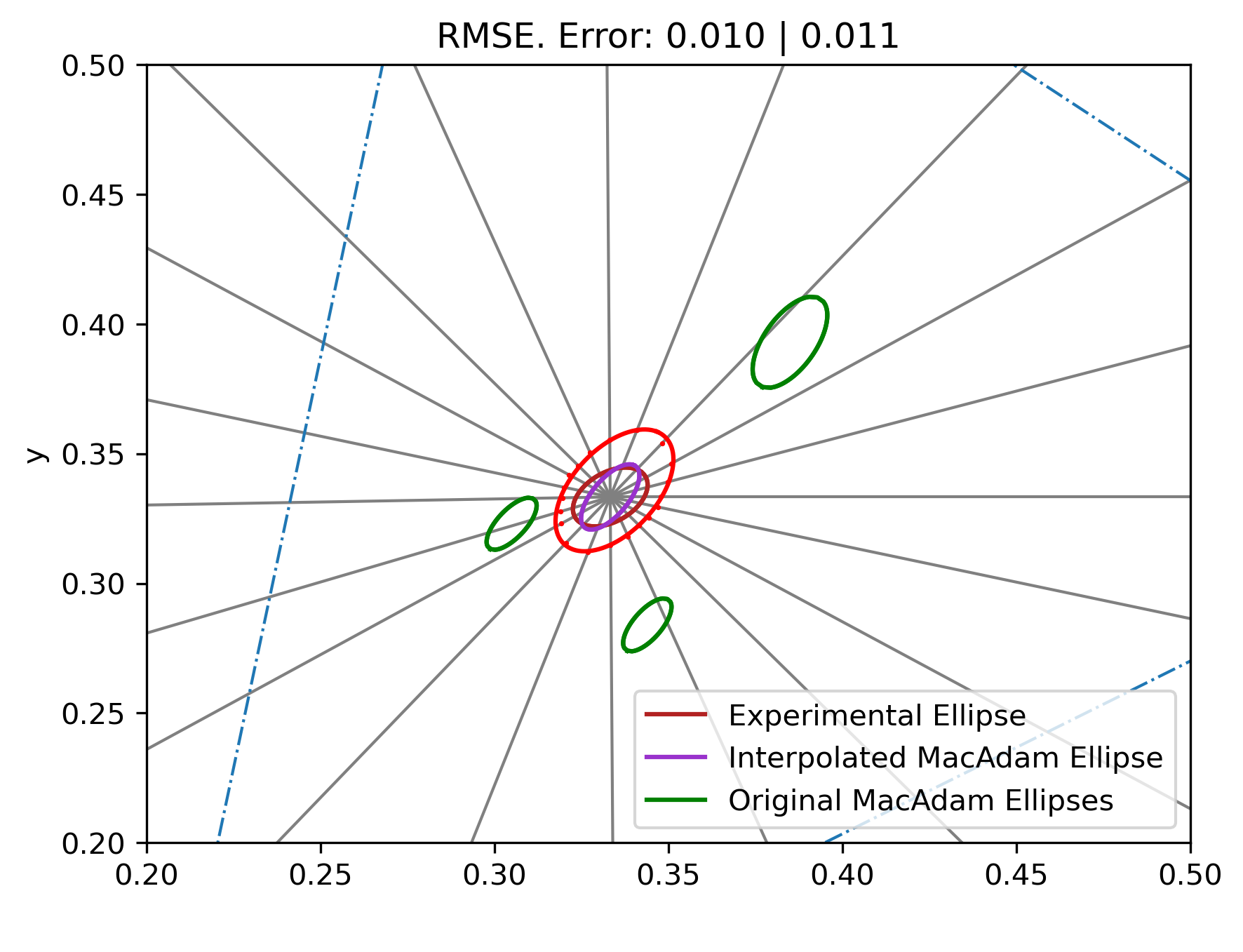}
		\end{subfigure}%
		\begin{subfigure}{0.33\textwidth}
			\centering
			\includegraphics[width=1\linewidth]{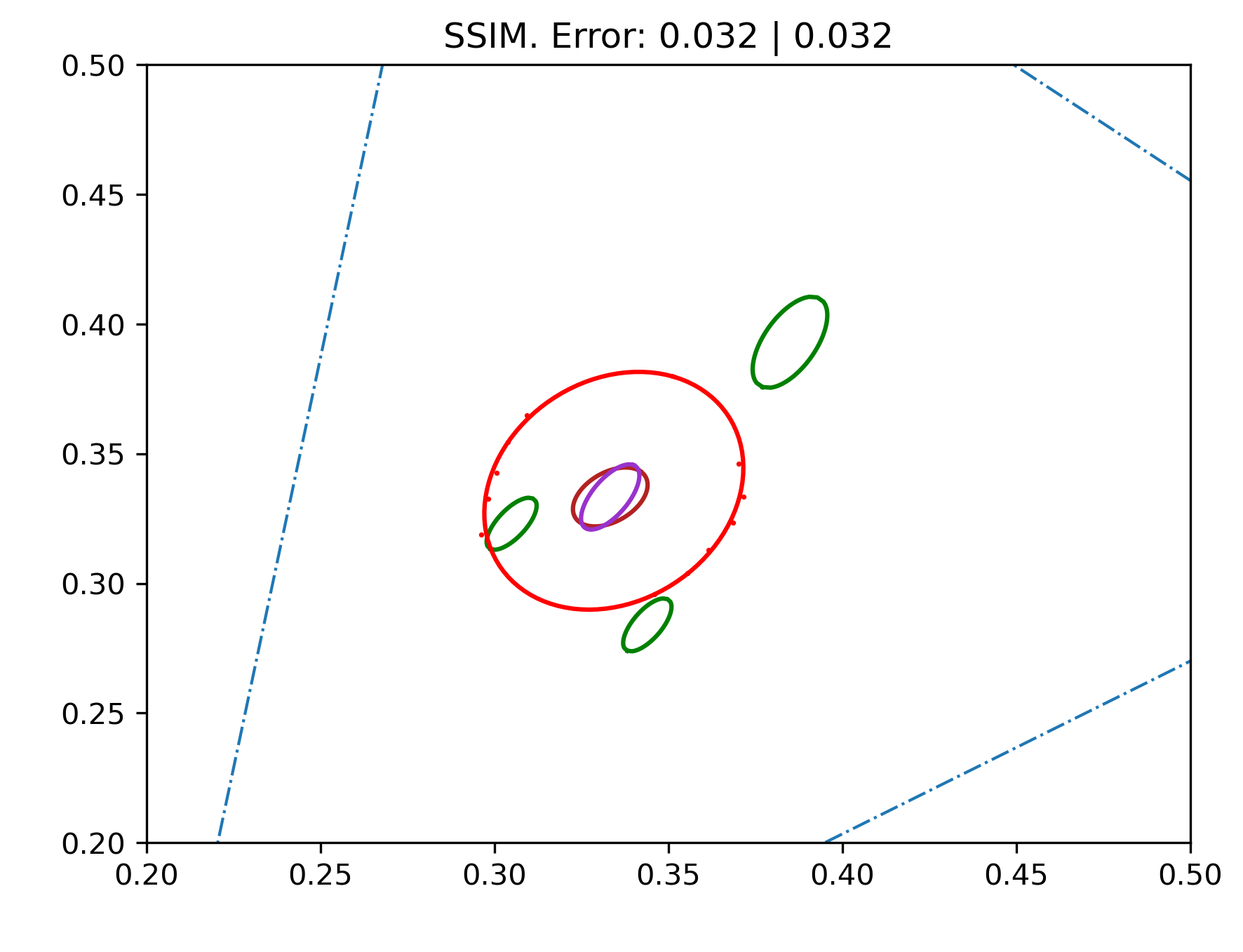}			
		\end{subfigure}
            \begin{subfigure}{0.33\textwidth}
			\centering
			\includegraphics[width=1\linewidth]{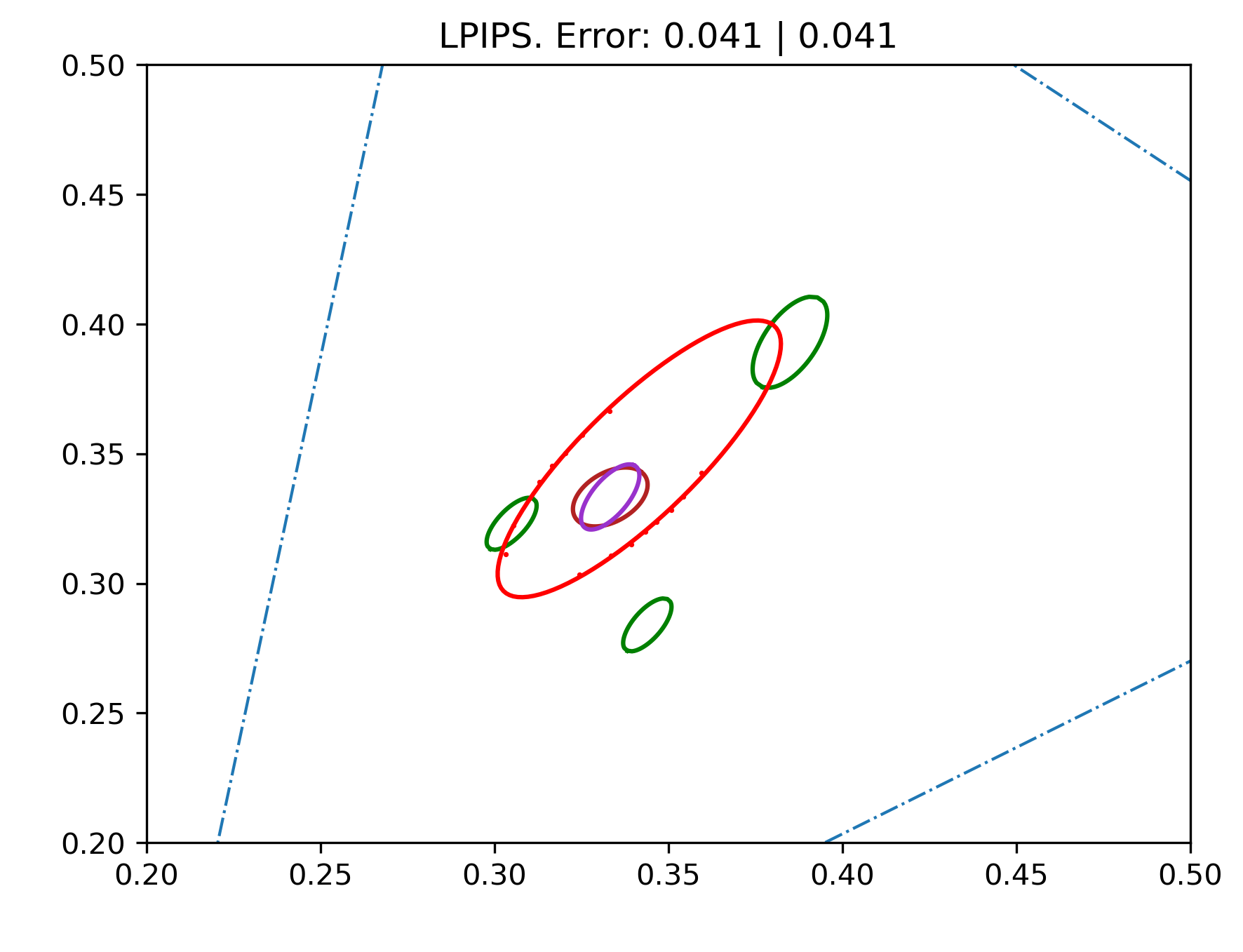}			
		\end{subfigure}

            \begin{subfigure}{0.33\textwidth}
			\centering
			\includegraphics[width=1\linewidth]{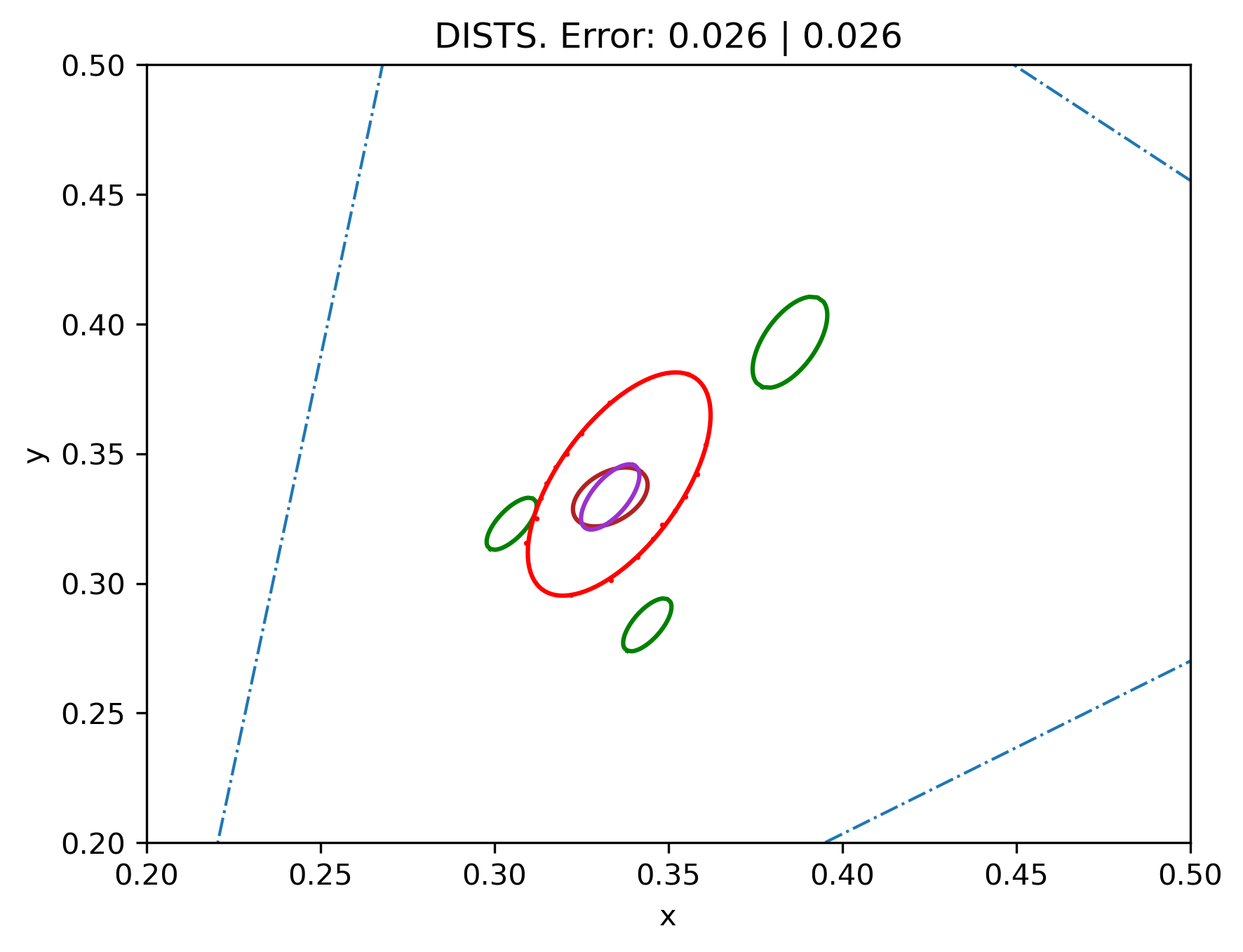}
		\end{subfigure}%
		\begin{subfigure}{0.33\textwidth}
			\centering
			\includegraphics[width=1\linewidth]{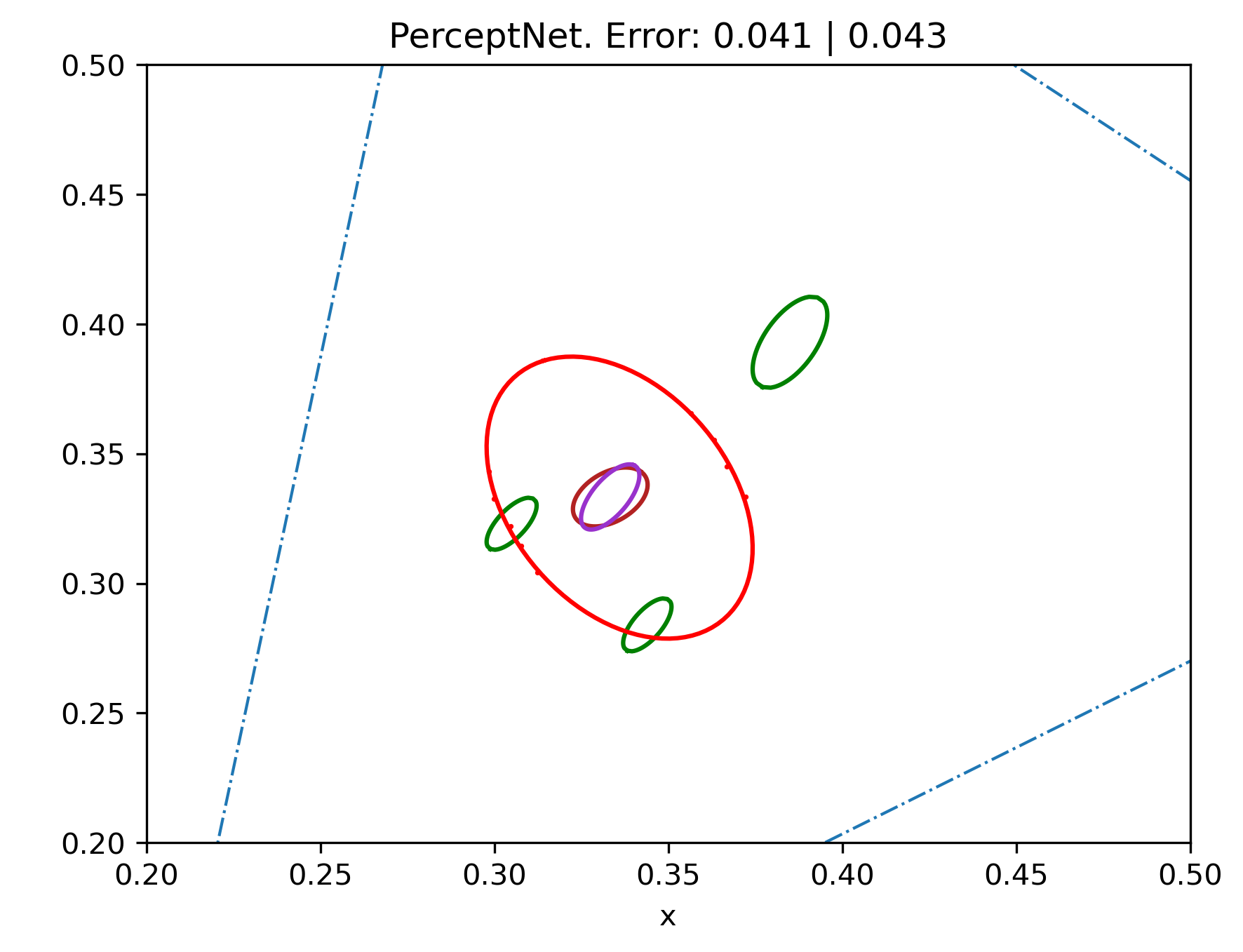}
		\end{subfigure}
            \begin{subfigure}{0.33\textwidth}
			\centering
			\includegraphics[width=1\linewidth]{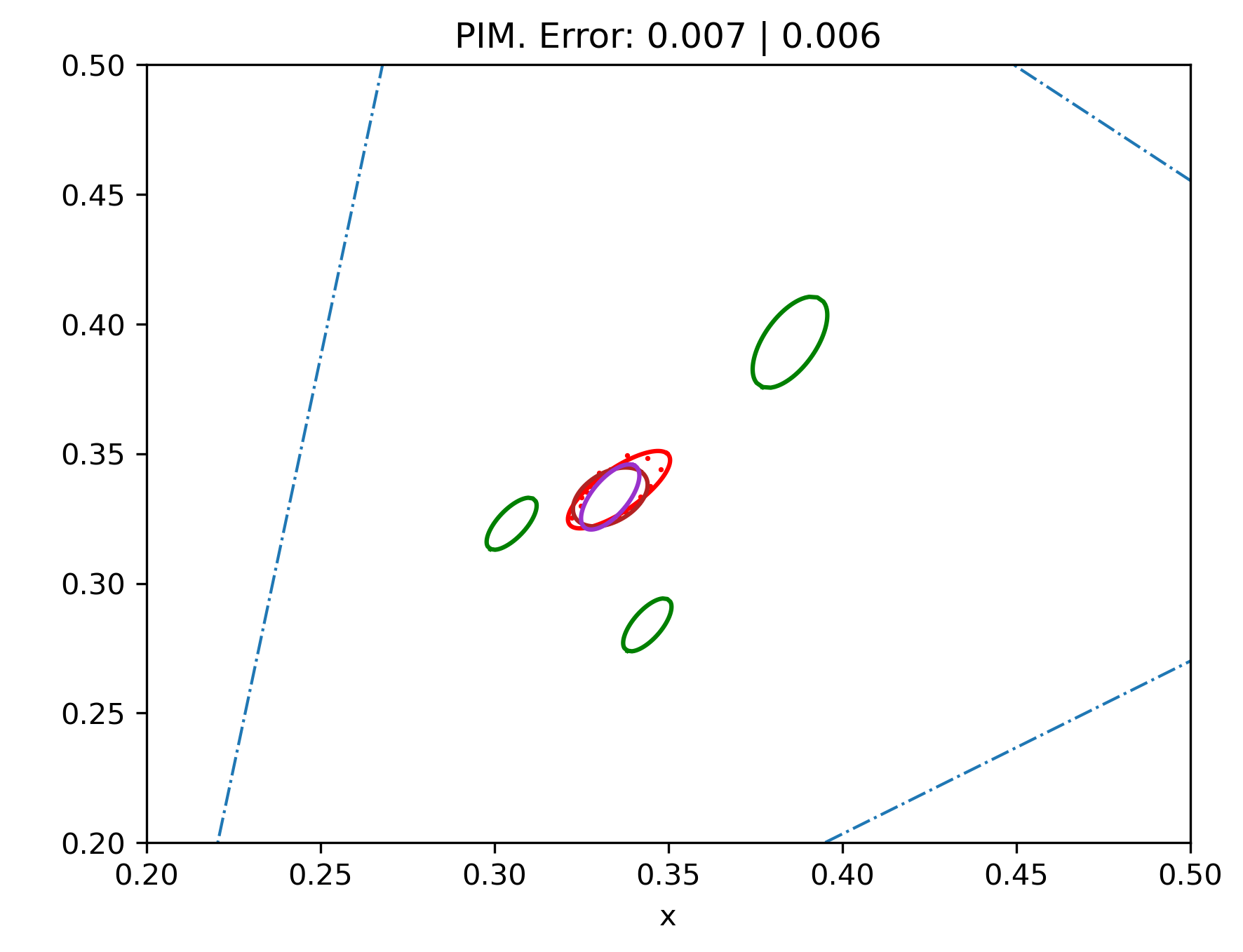}
		\end{subfigure}
            \label{fig:leaky}
		        \caption{Results of the different metrics in the color discrimination experiments, together with their error for TID13 data. The error with respect to the fitted MacAdam ellipse is shown on the left whereas the error with respect to the experimentally obtained ellipse is shown on the right. The first subfigure also shows the direction of the 20 hues. The metric that makes the least error with respect to the MacAdam ellipses is PIM in both cases.}
		\label{fig:results_ilu_TID}
	\end{figure*}

\begin{figure*}[!h]
		
		\begin{subfigure}{0.33\textwidth}
			\centering
			\includegraphics[width=1\linewidth]{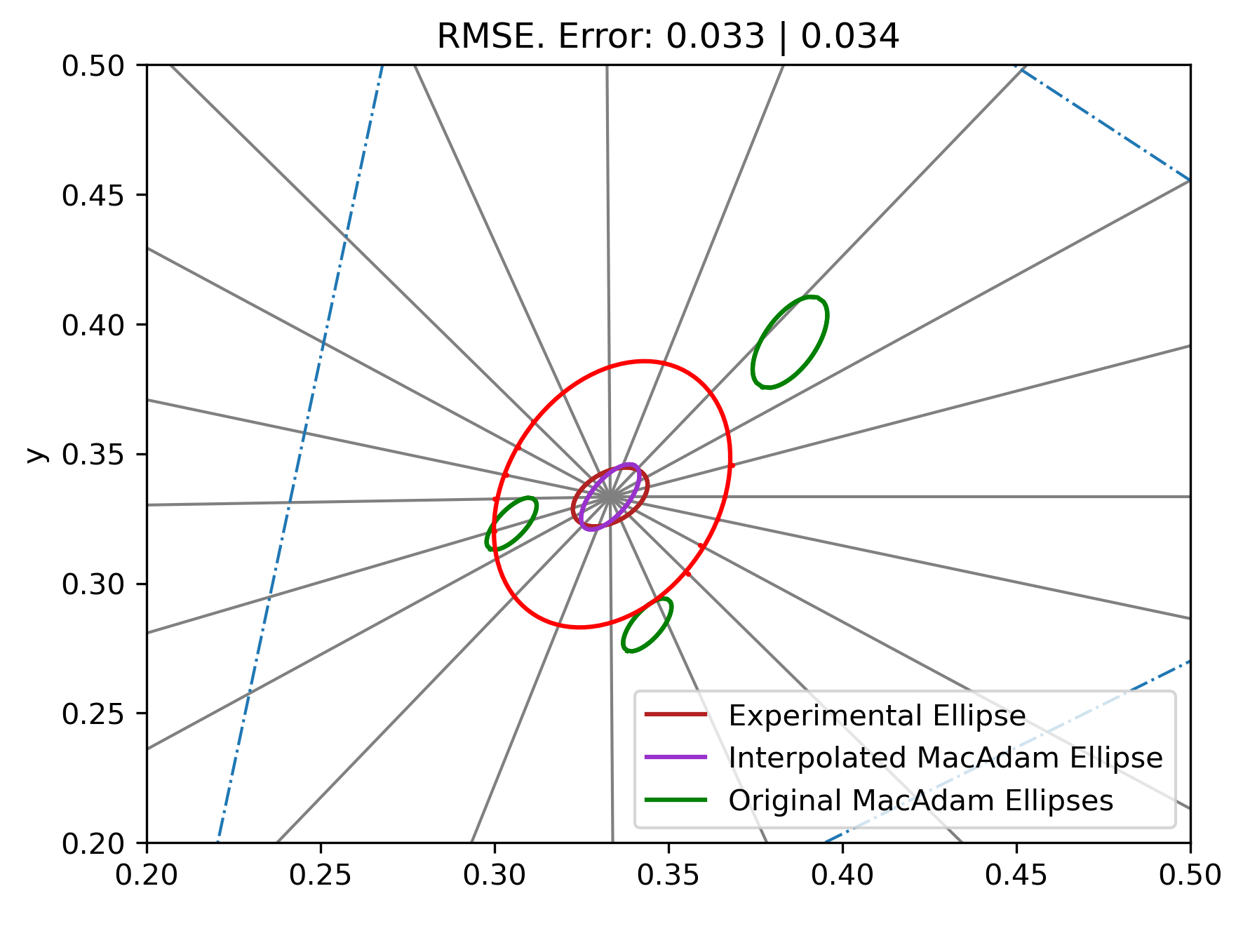}
		\end{subfigure}%
		\begin{subfigure}{0.33\textwidth}
			\centering
			\includegraphics[width=1\linewidth]{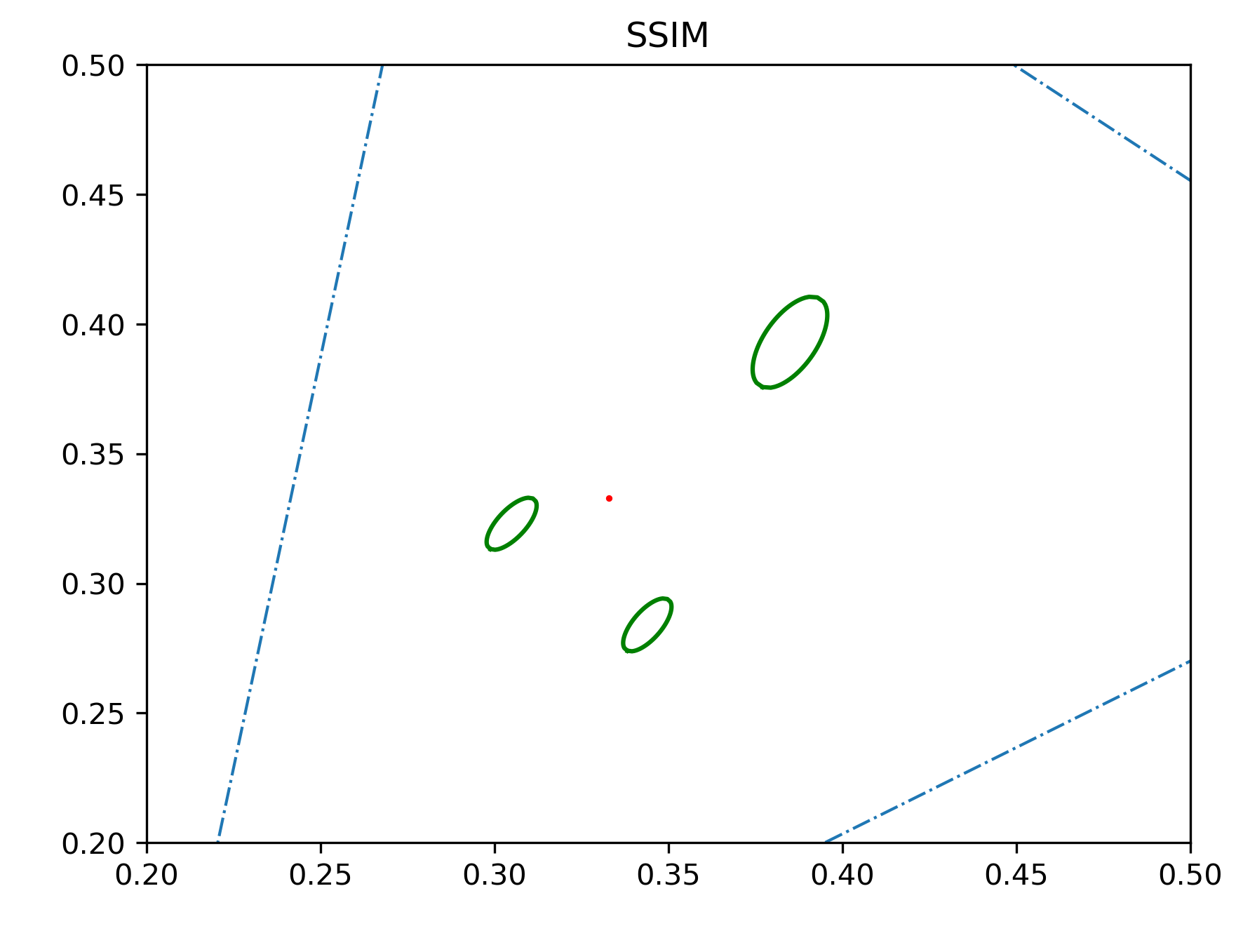}			
		\end{subfigure}
            \begin{subfigure}{0.33\textwidth}
			\centering
			\includegraphics[width=1\linewidth]{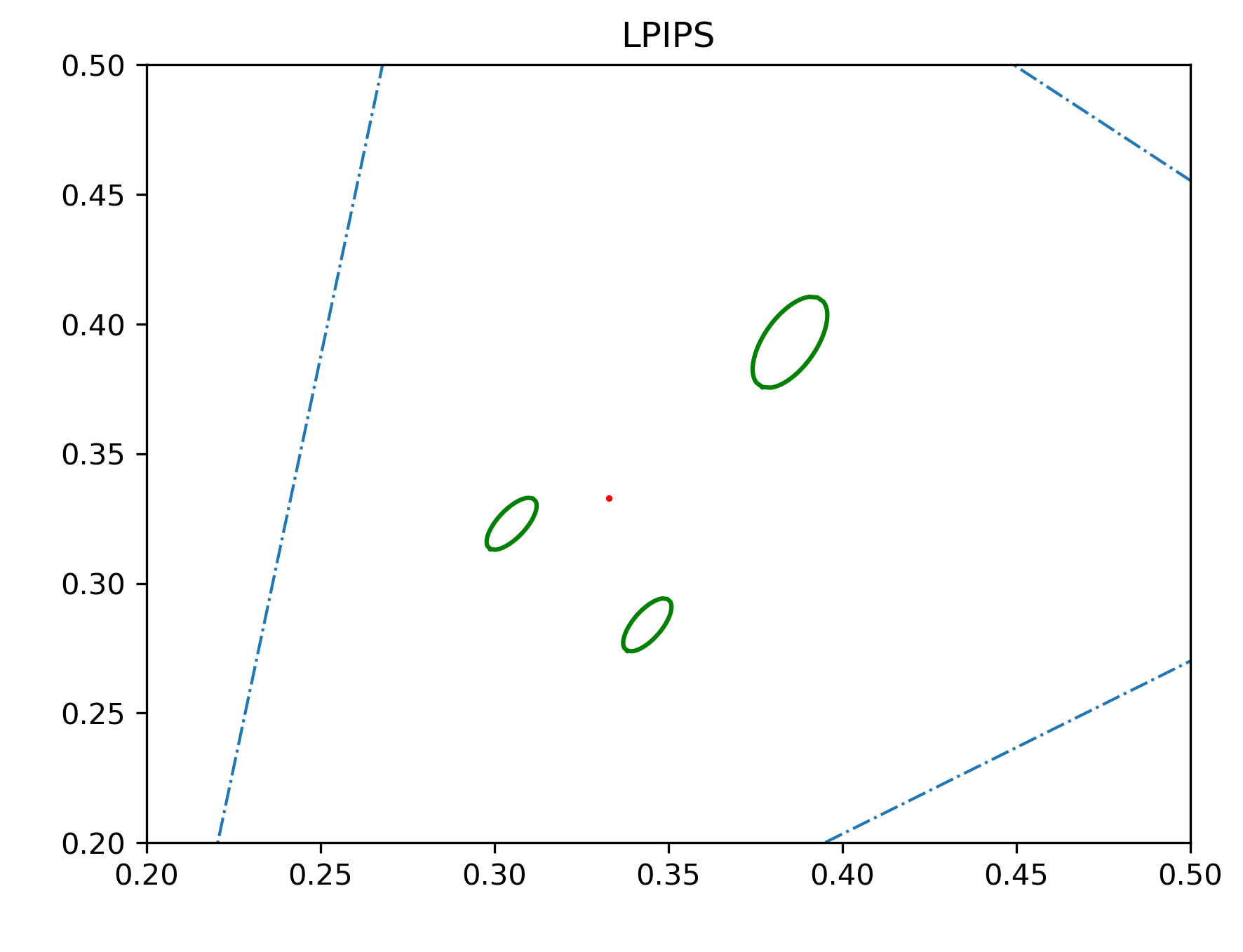}			
		\end{subfigure}

            \begin{subfigure}{0.33\textwidth}
			\centering
			\includegraphics[width=1\linewidth]{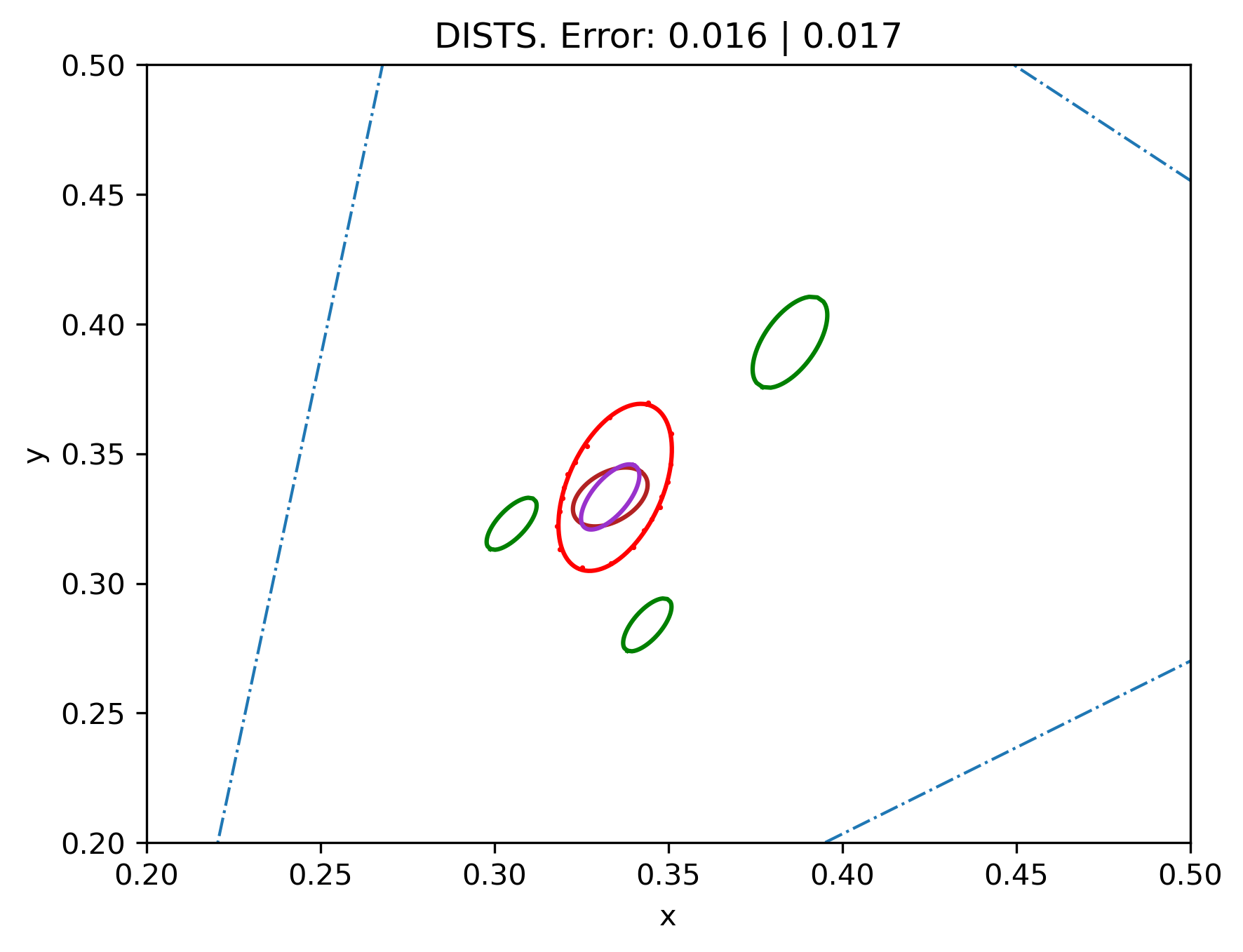}
		\end{subfigure}%
		\begin{subfigure}{0.33\textwidth}
			\centering
			\includegraphics[width=1\linewidth]{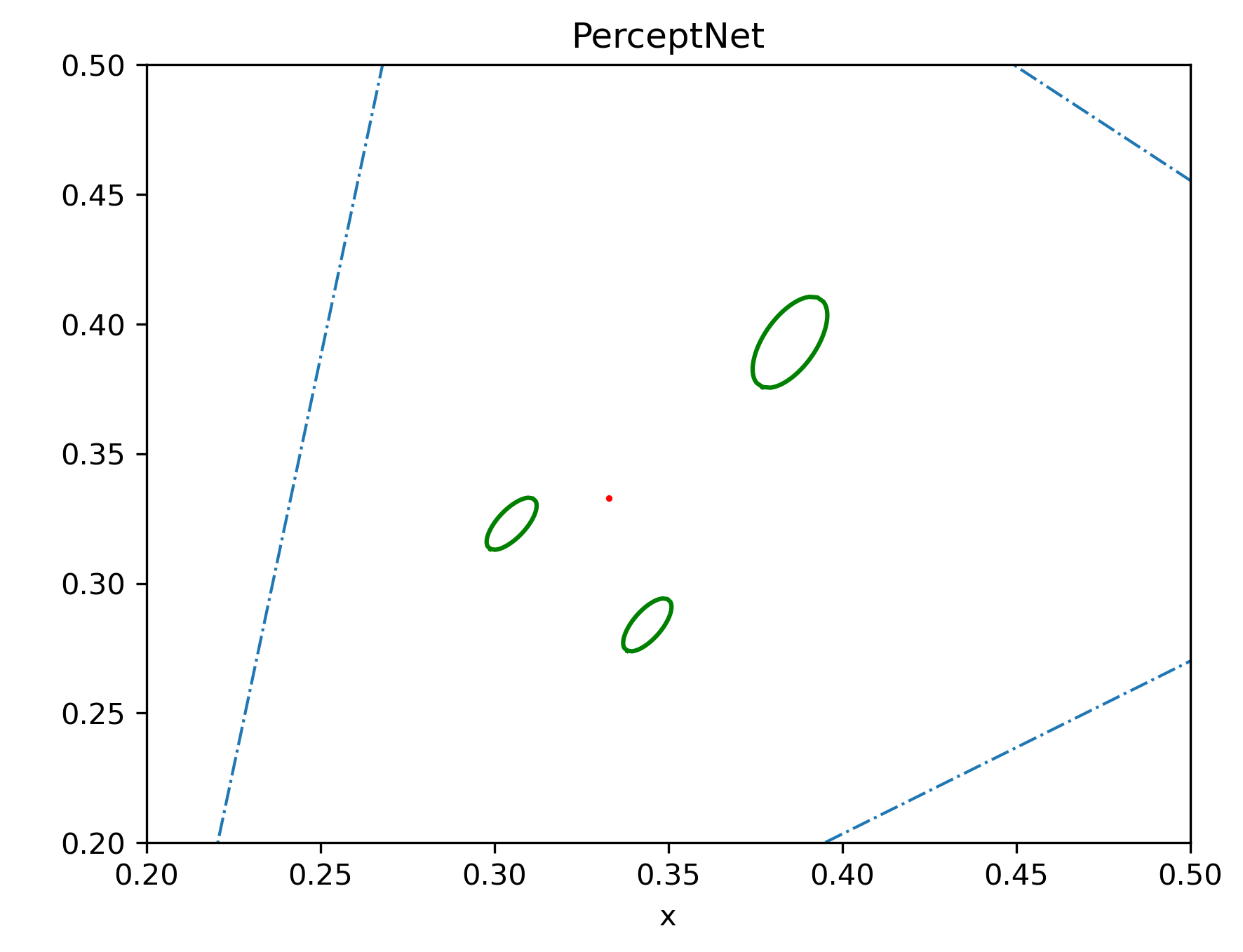}
		\end{subfigure}
            \begin{subfigure}{0.33\textwidth}
			\centering
			\includegraphics[width=1\linewidth]{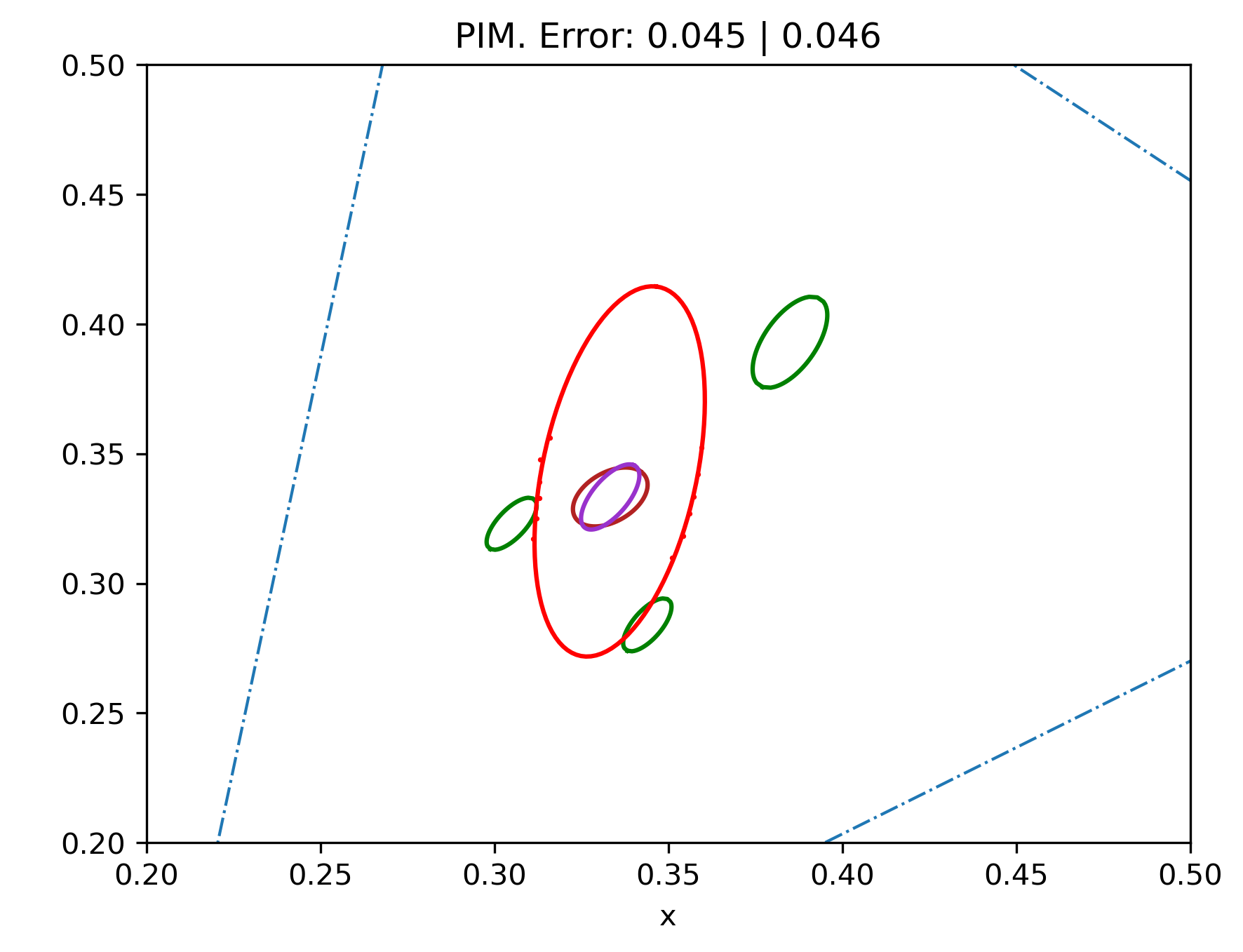}
		\end{subfigure}
            \label{fig:leaky}
		        \caption{Results of the different metrics in the color discrimination experiments, together with their error for MNIST data. The error concerning the fitted MacAdam ellipse is shown on the left whereas the error with respect to the experimentally obtained ellipse is shown on the right. The first subfigure also shows the direction of the 20 hues. The metric that makes the least error with respect to the MacAdam ellipses is DISTS in both cases.}
		\label{fig:results_ilu_MNIST}
	\end{figure*}

\begin{figure*}[!h]
		
		\begin{subfigure}{0.33\textwidth}
			\centering
			\includegraphics[width=1\linewidth]{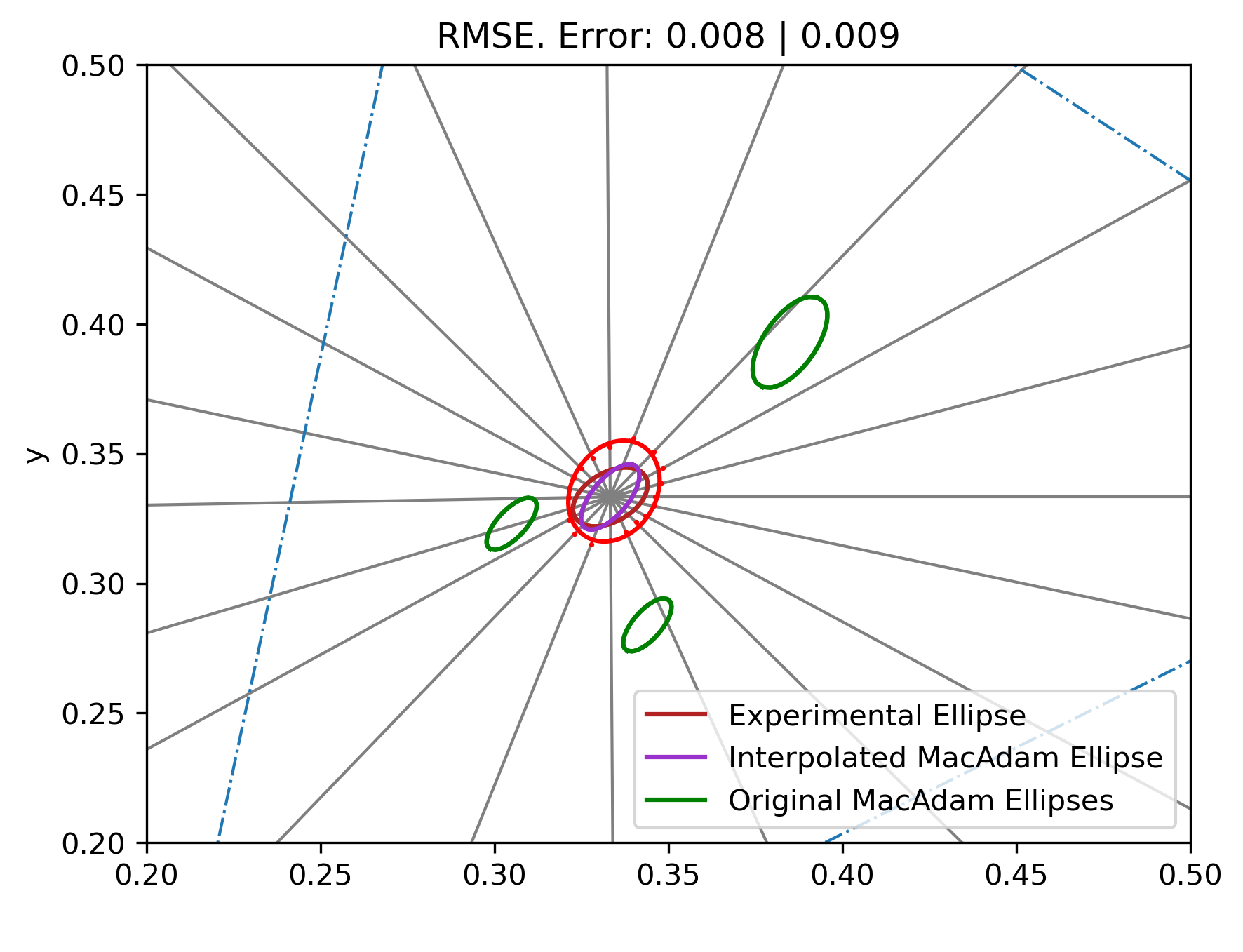}
		\end{subfigure}%
		\begin{subfigure}{0.33\textwidth}
			\centering
			\includegraphics[width=1\linewidth]{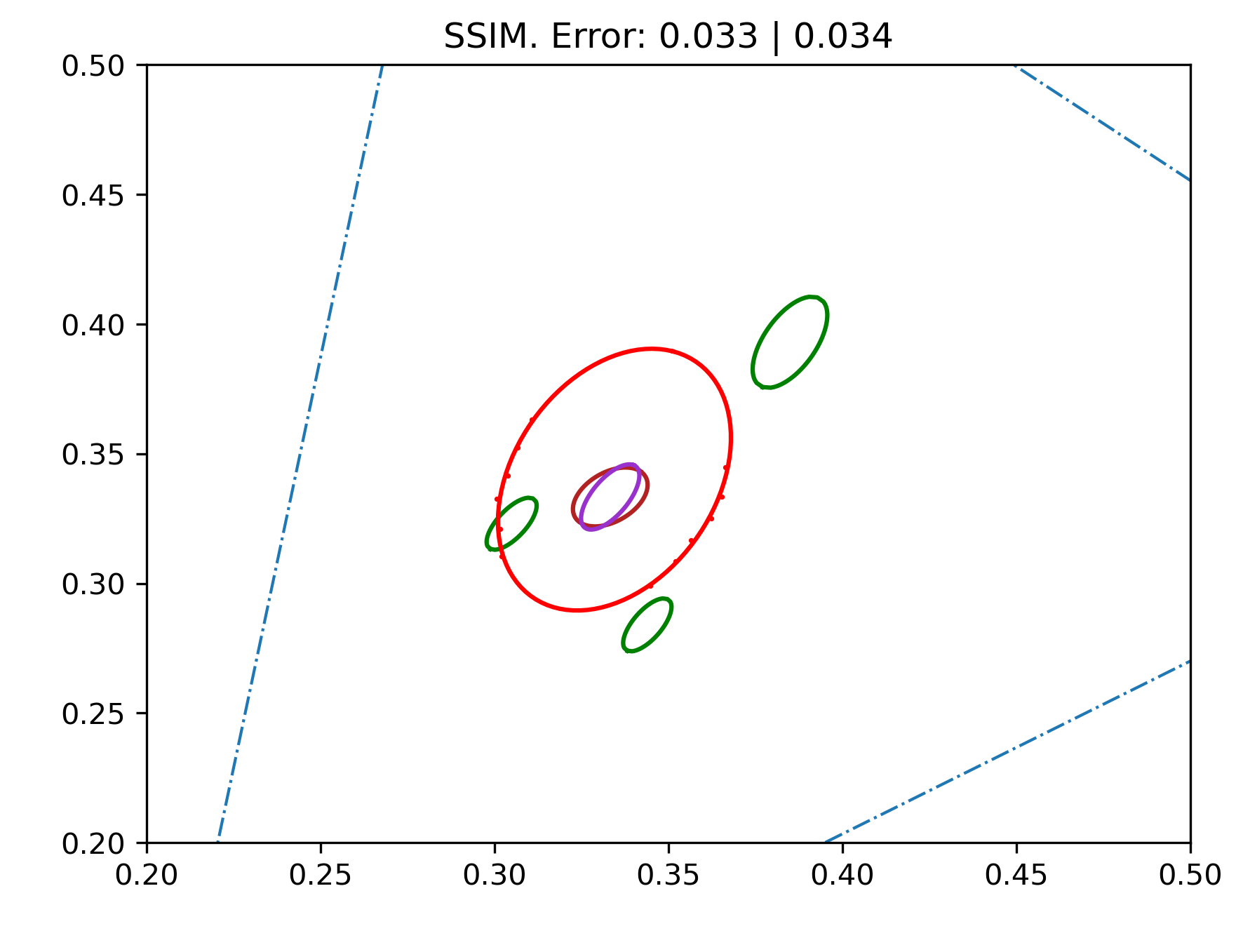}			
		\end{subfigure}
            \begin{subfigure}{0.33\textwidth}
			\centering
			\includegraphics[width=1\linewidth]{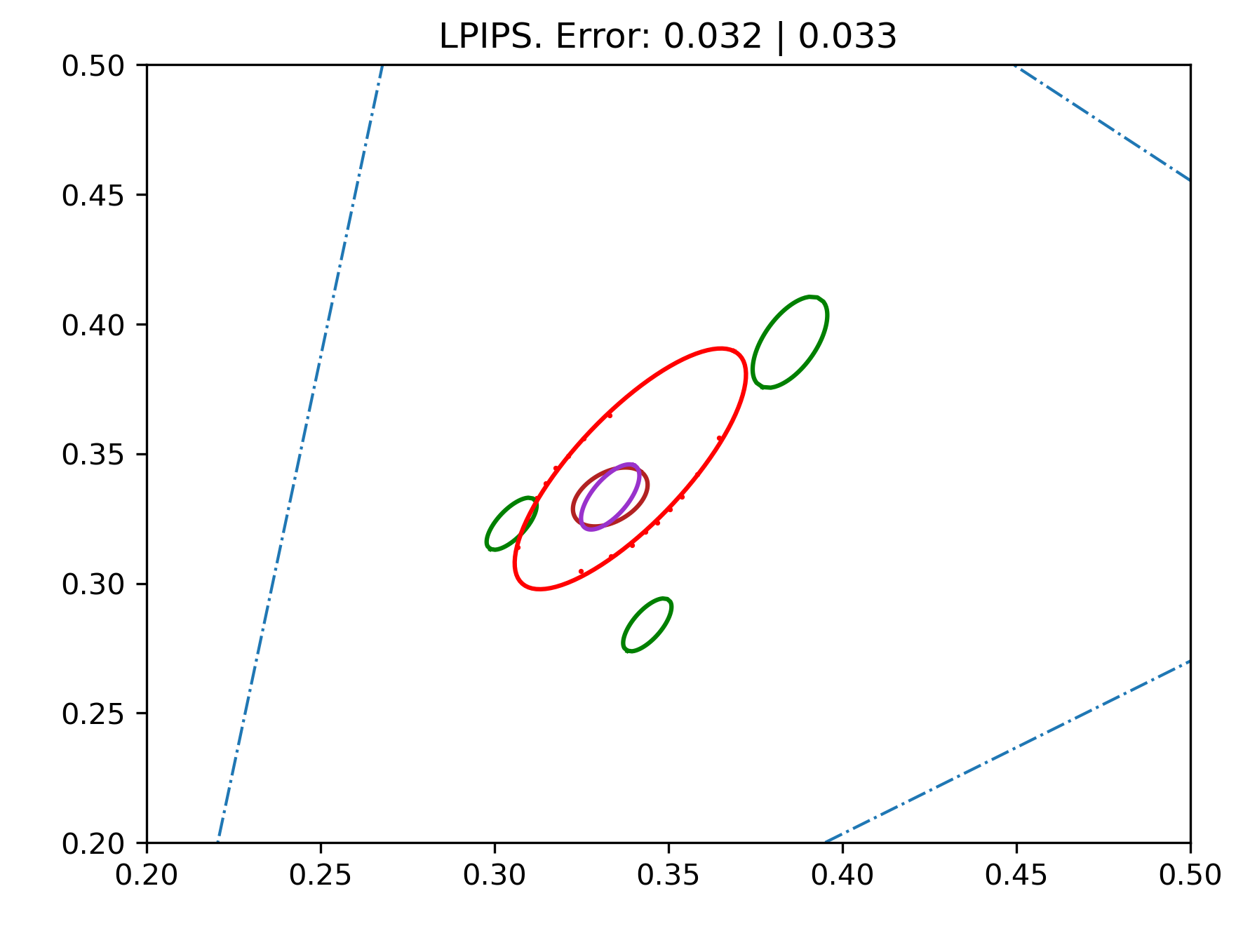}			
		\end{subfigure}

            \begin{subfigure}{0.33\textwidth}
			\centering
			\includegraphics[width=1\linewidth]{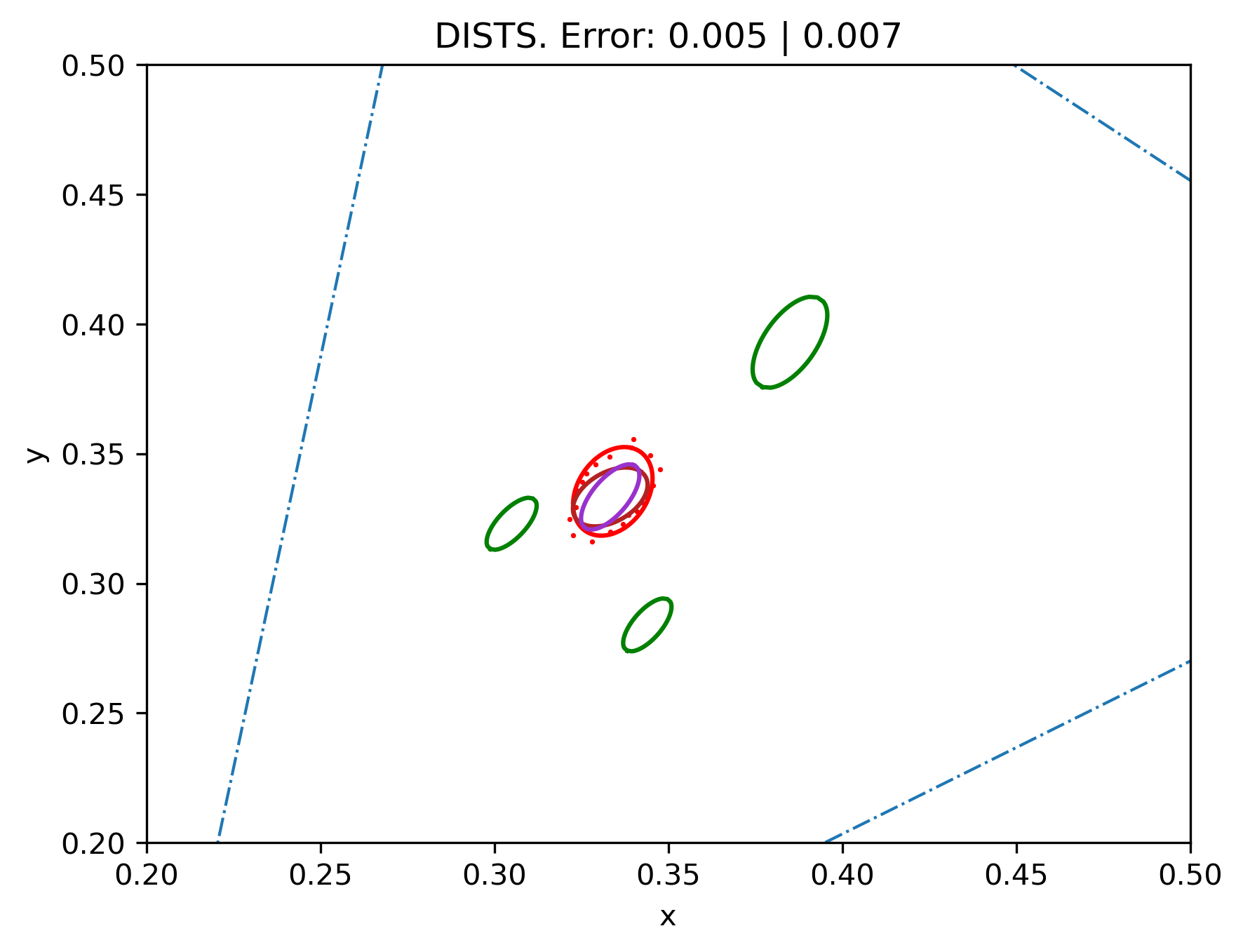}
		\end{subfigure}%
		\begin{subfigure}{0.33\textwidth}
			\centering
			\includegraphics[width=1\linewidth]{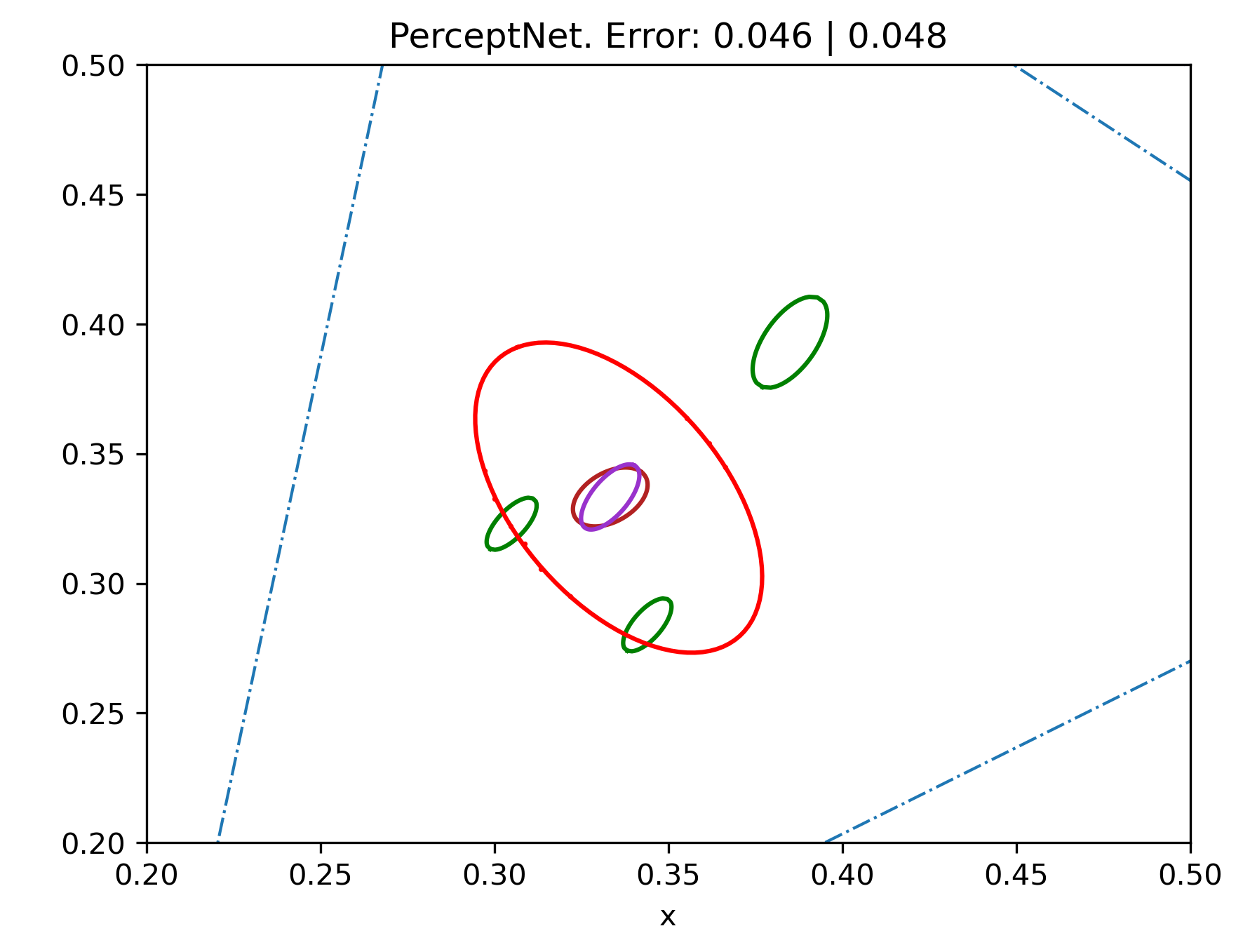}
		\end{subfigure}
            \begin{subfigure}{0.33\textwidth}
			\centering
			\includegraphics[width=1\linewidth]{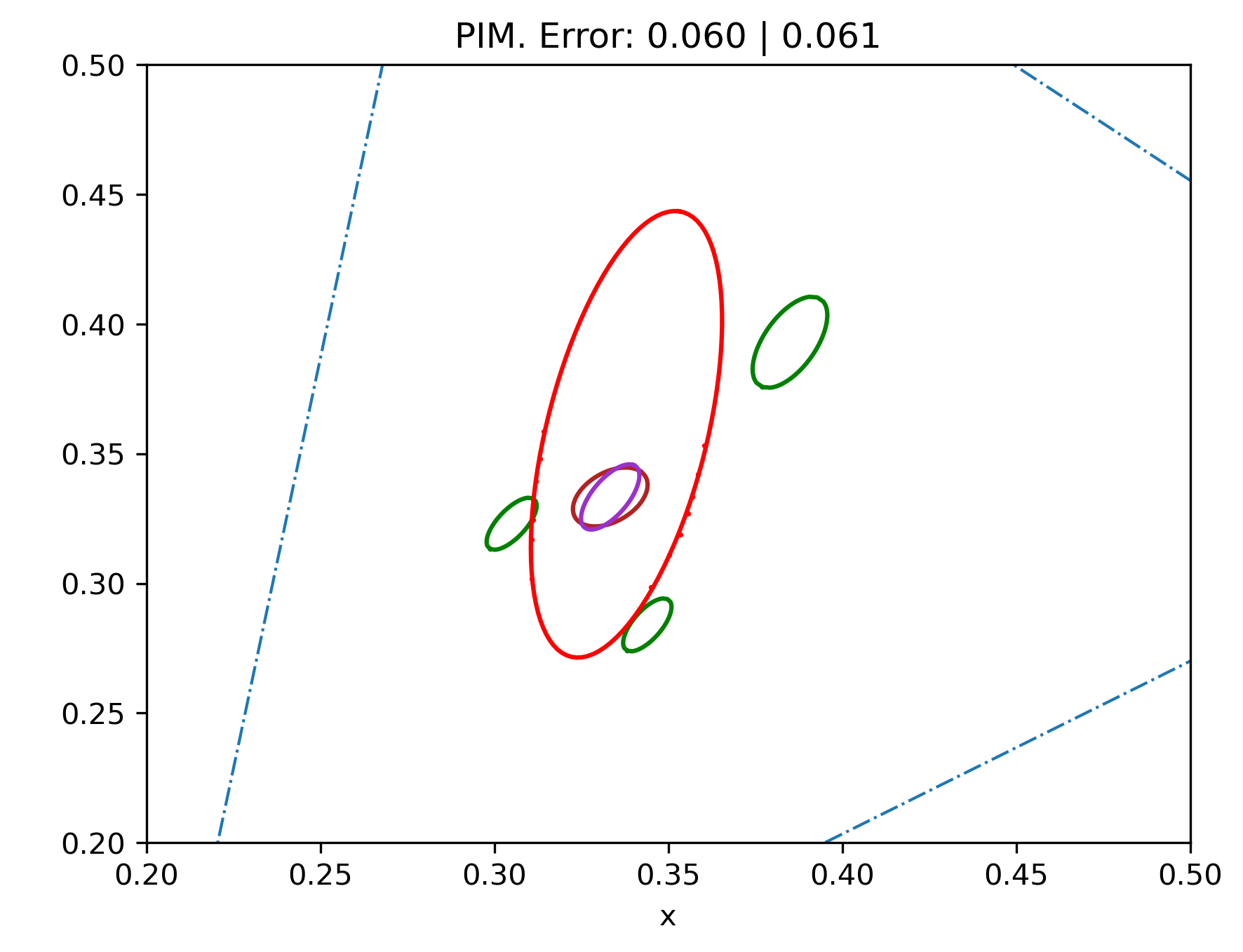}
		\end{subfigure}
            \label{fig:leaky}
		        \caption{Results of the different metrics in the color discrimination experiments, together with their error for ImageNet data. The error with respect to the fitted MacAdam ellipse is shown on the left whereas the error with respect to the experimentally obtained ellipse is shown on the right. The first subfigure also shows the direction of the 20 hues. The metric that makes the least error with respect to the MacAdam ellipses is DISTS in both cases.}
		\label{fig:results_ilu_CIFAR}
	\end{figure*}

Figures~\ref{fig:results_ilu_MNIST}, \ref{fig:results_ilu_CIFAR}, \ref{fig:results_ilu_TID}, \ref{fig:results_ilu_ima} show the results for the all databases for the illuminant transformations showing the final ellipse in the chromatic diagram. The title of each subfigure specifies the errors made with respect to the MacAdam ellipse at that point in space (left) and with respect to the experimentally obtained ellipse (right).

\appendix
\section{Appendix D.}
\label{app:apndD}

\begin{figure*}[!h]
		
		\begin{subfigure}{0.33\textwidth}
			\centering
			\includegraphics[width=1\linewidth]{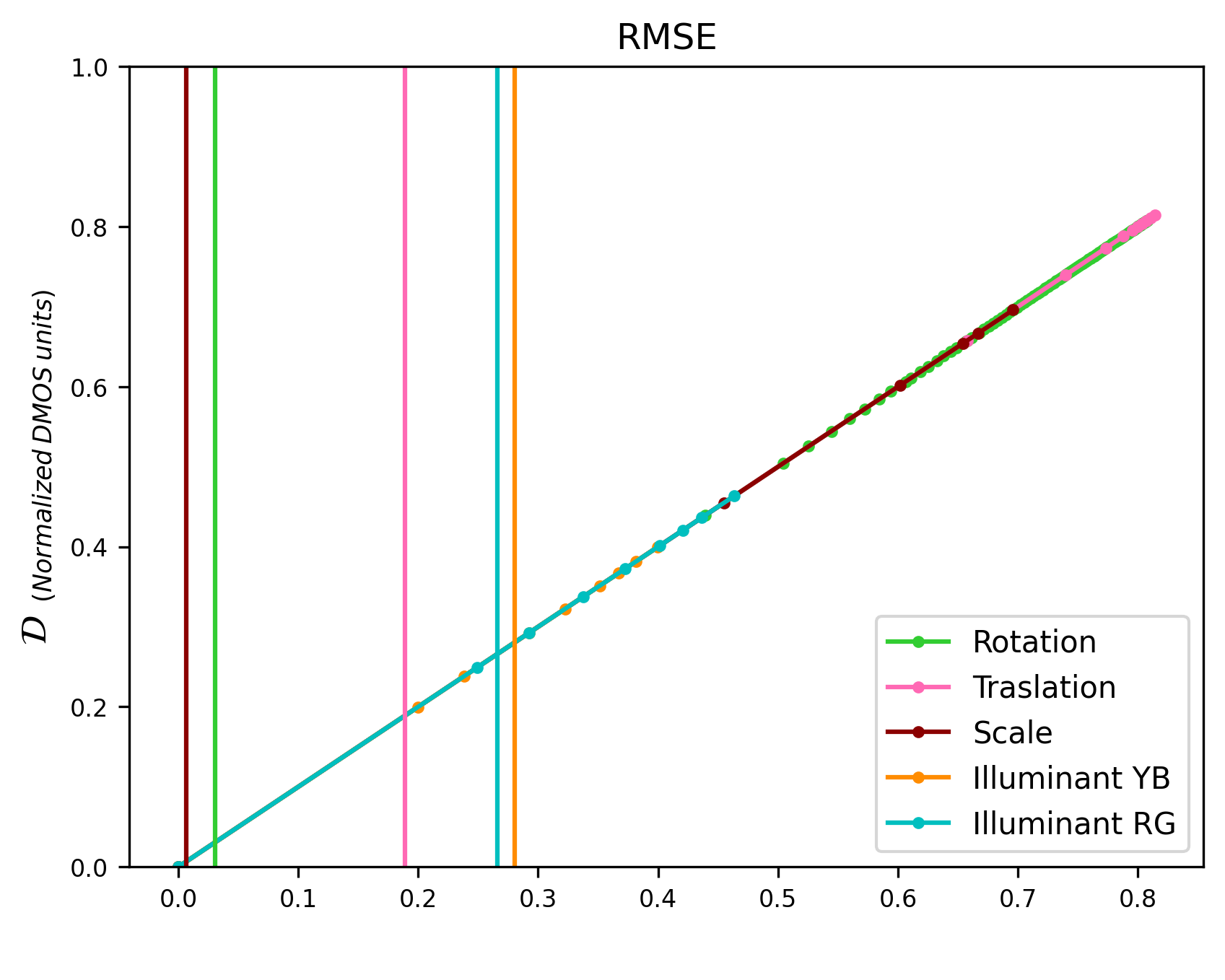}
		\end{subfigure}%
		\begin{subfigure}{0.33\textwidth}
			\centering
			\includegraphics[width=1\linewidth]{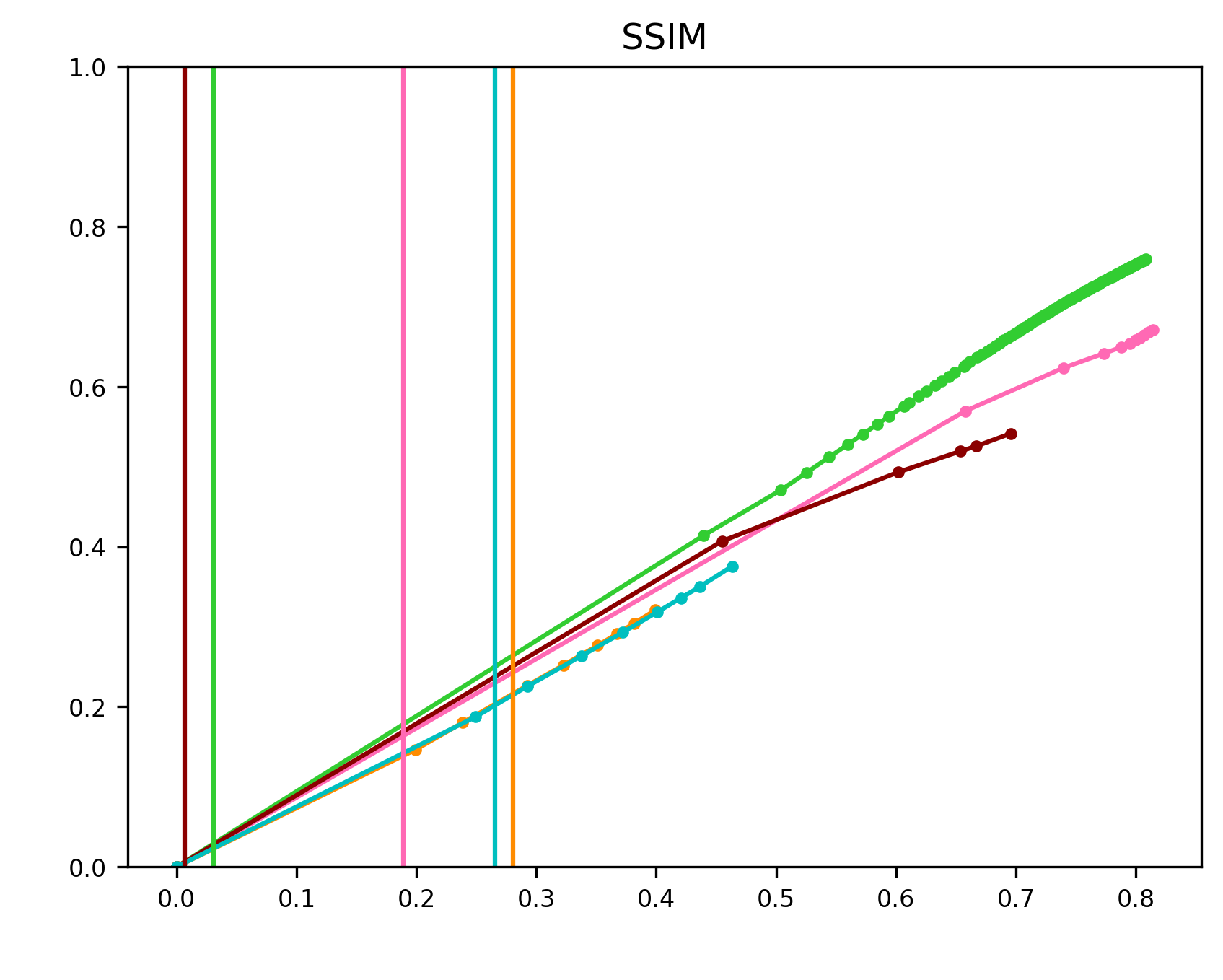}			
		\end{subfigure}
            \begin{subfigure}{0.33\textwidth}
			\centering
			\includegraphics[width=1\linewidth]{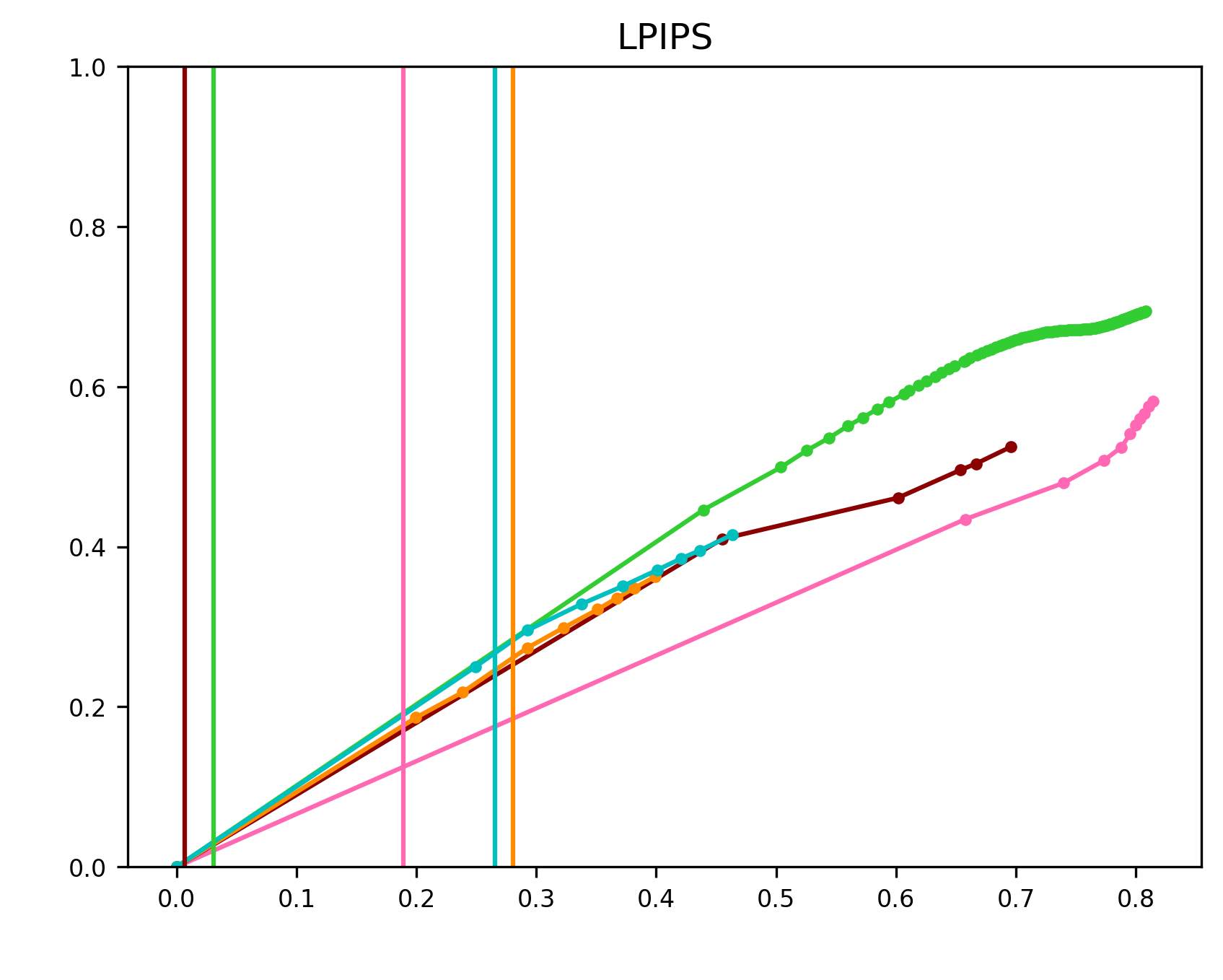}			
		\end{subfigure}
  
            \begin{subfigure}{0.33\textwidth}
			\centering
			\includegraphics[width=1\linewidth]{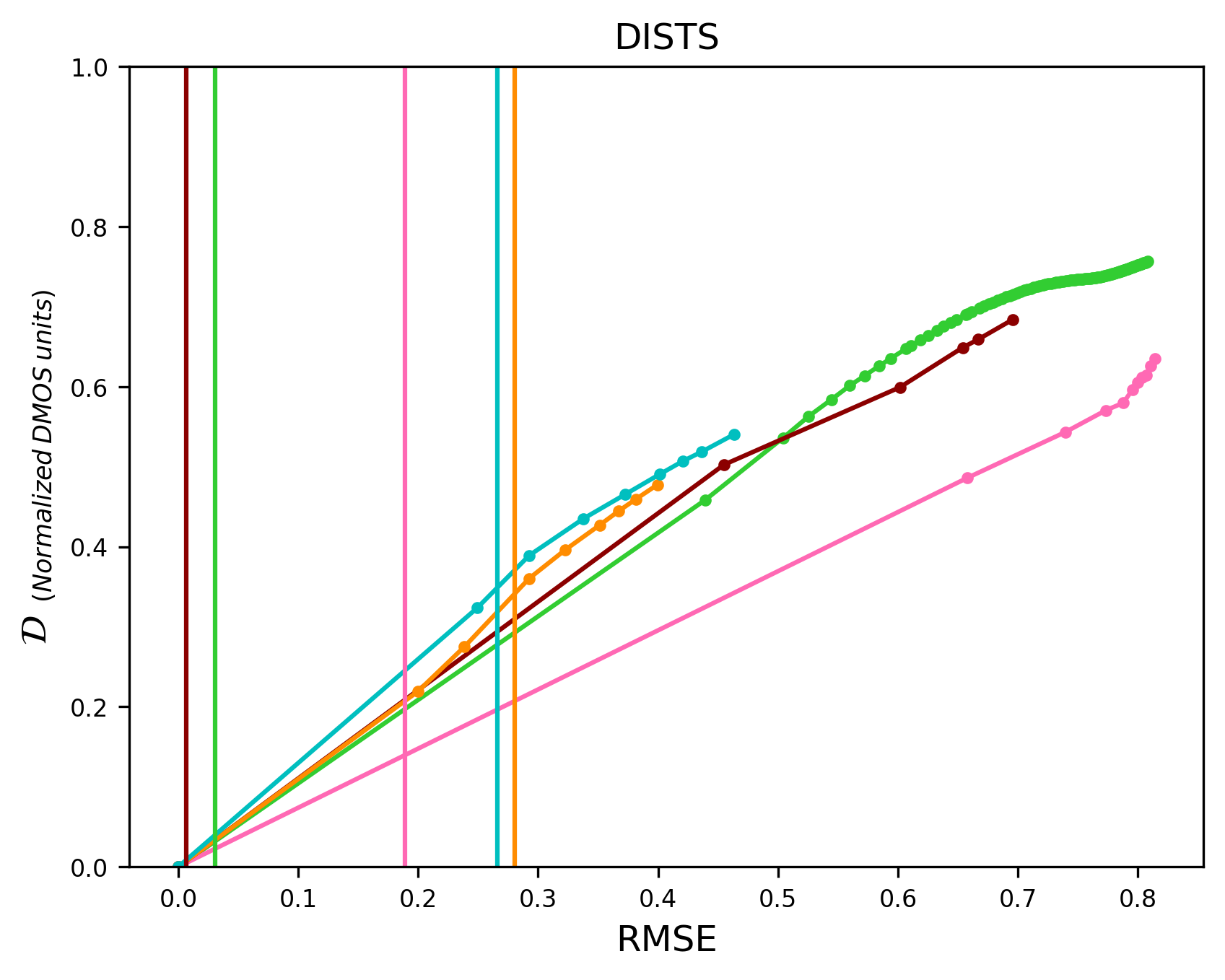}
		\end{subfigure}%
		\begin{subfigure}{0.33\textwidth}
			\centering
			\includegraphics[width=1\linewidth]{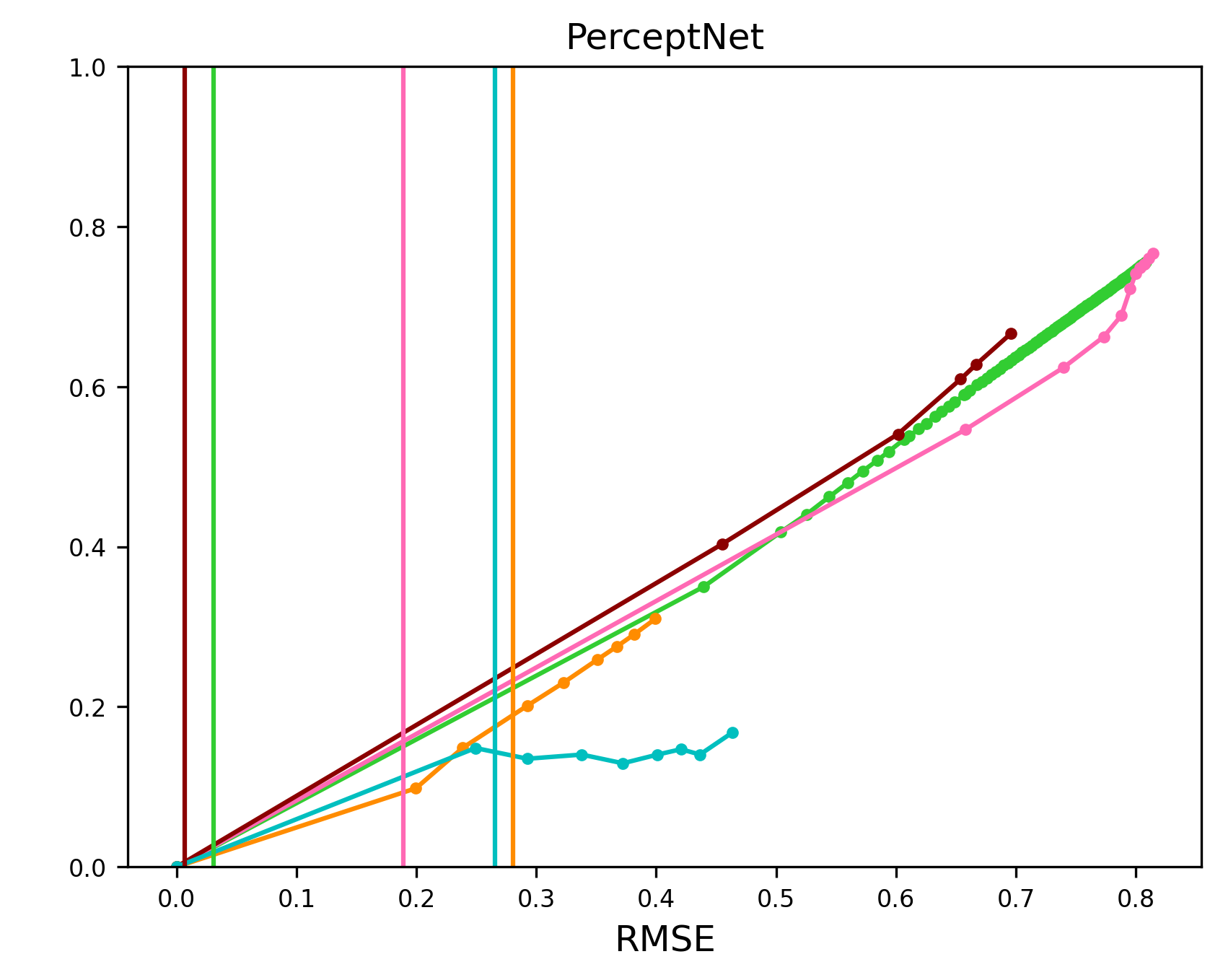}
		\end{subfigure}
            \begin{subfigure}{0.33\textwidth}
			\centering
			\includegraphics[width=1\linewidth]{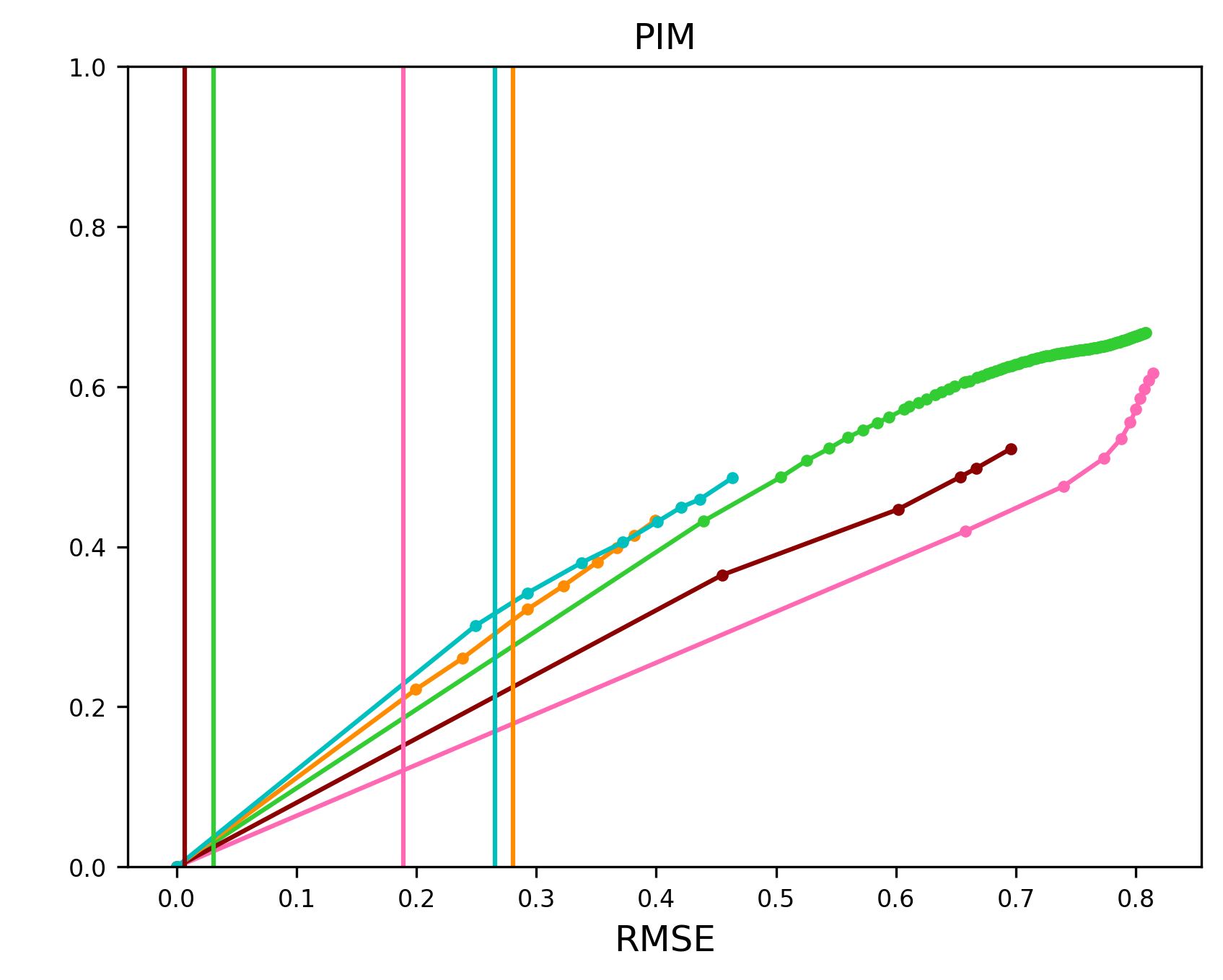}
		\end{subfigure}
            \label{fig:leaky}
		        \caption{Sensitivity test results for the MNIST dataset. Table \ref{tab:resultados2} shows in detail the order of the sensitivities together with the human ordering, where it can be seen that no metric follows the order correctly. However, in all cases the relative order of illuminant changes is respected, i.e. they are more sensitive to the RG direction than to the YB direction.}
		\label{fig:results_ord_mnist}
	\end{figure*}

\begin{figure*}[!h]
		
		\begin{subfigure}{0.33\textwidth}
			\centering
			\includegraphics[width=1\linewidth]{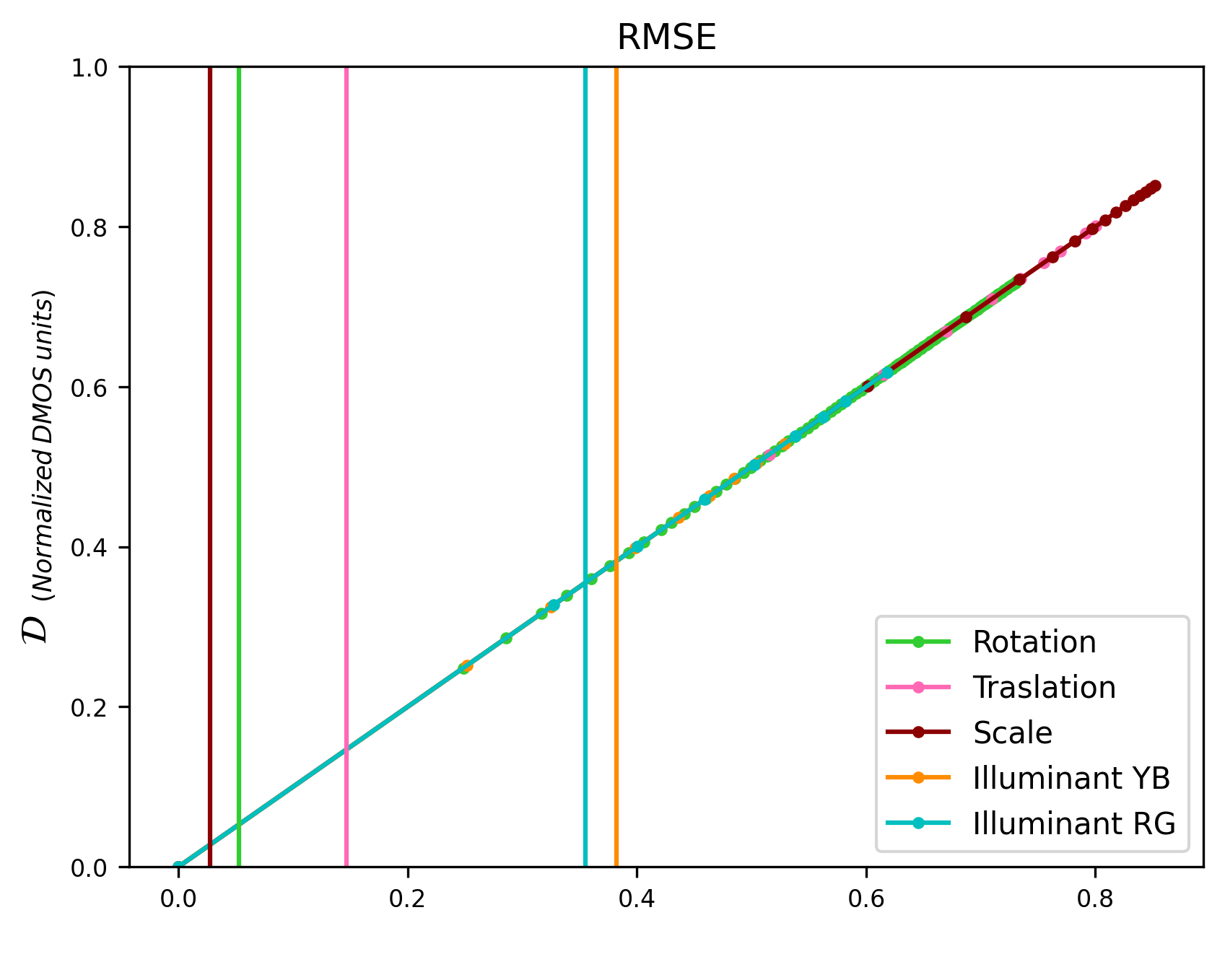}
		\end{subfigure}%
		\begin{subfigure}{0.33\textwidth}
			\centering
			\includegraphics[width=1\linewidth]{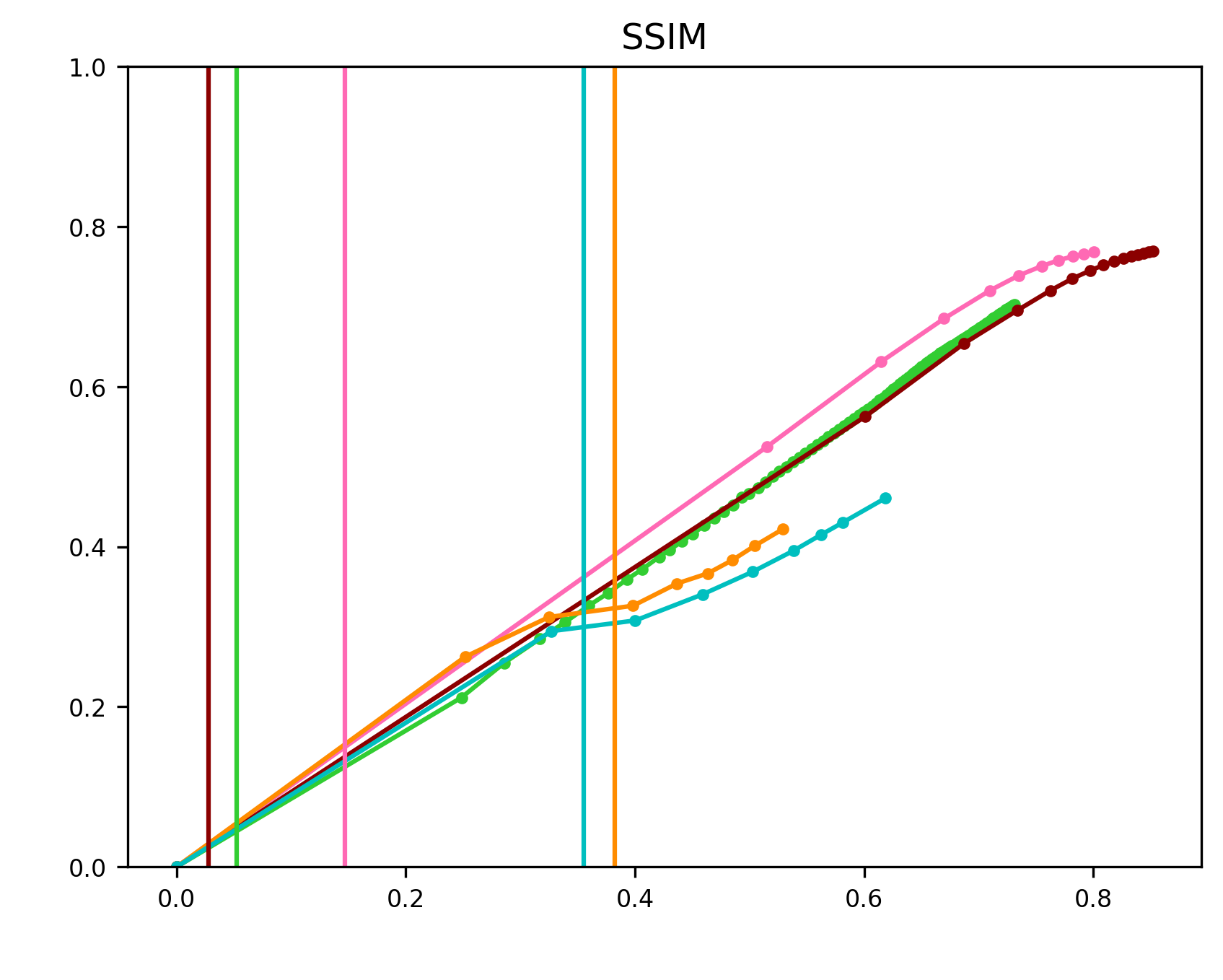}			
		\end{subfigure}
            \begin{subfigure}{0.33\textwidth}
			\centering
			\includegraphics[width=1\linewidth]{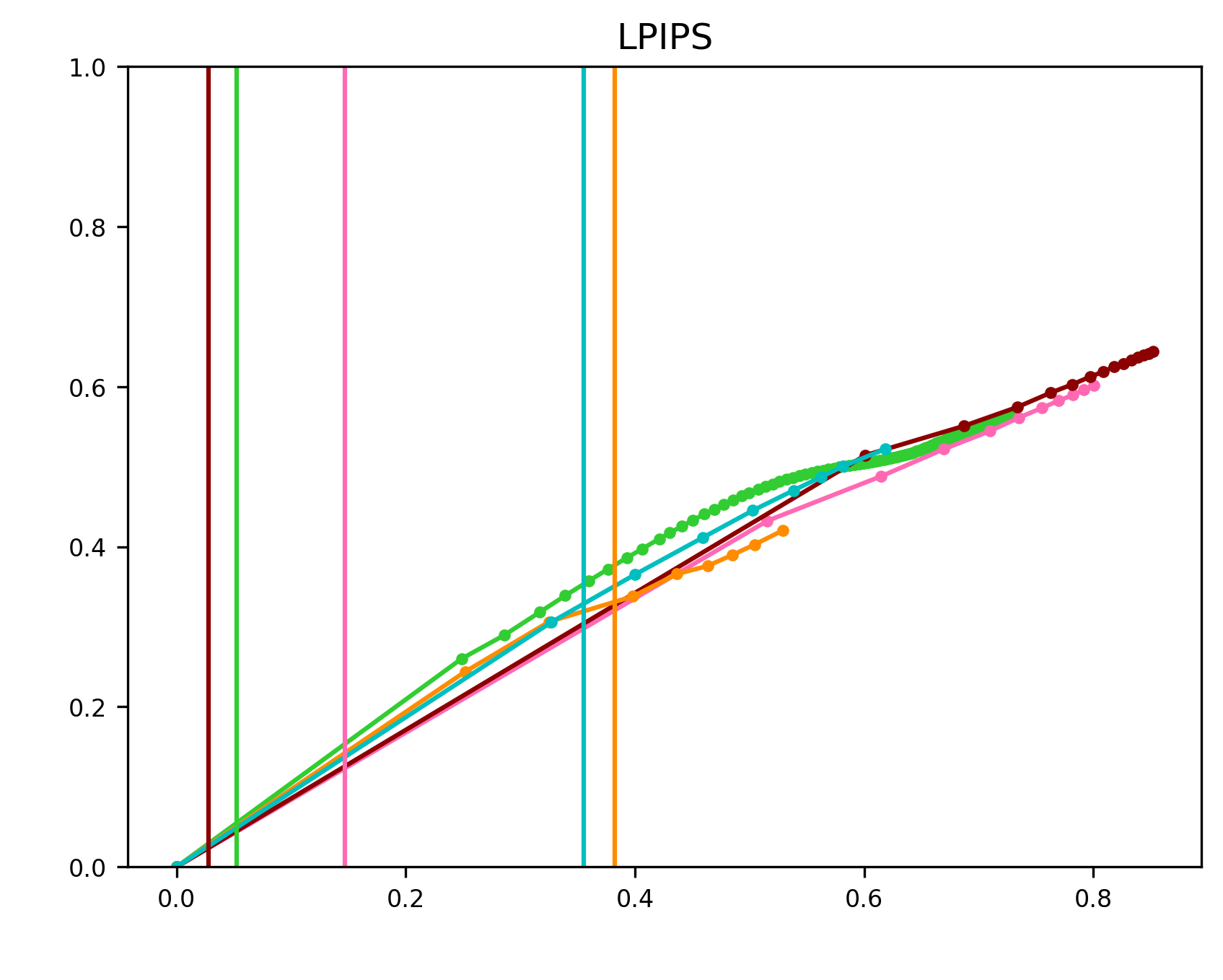}			
		\end{subfigure}
		\label{fig:results_rotation}

            \begin{subfigure}{0.33\textwidth}
			\centering
			\includegraphics[width=1\linewidth]{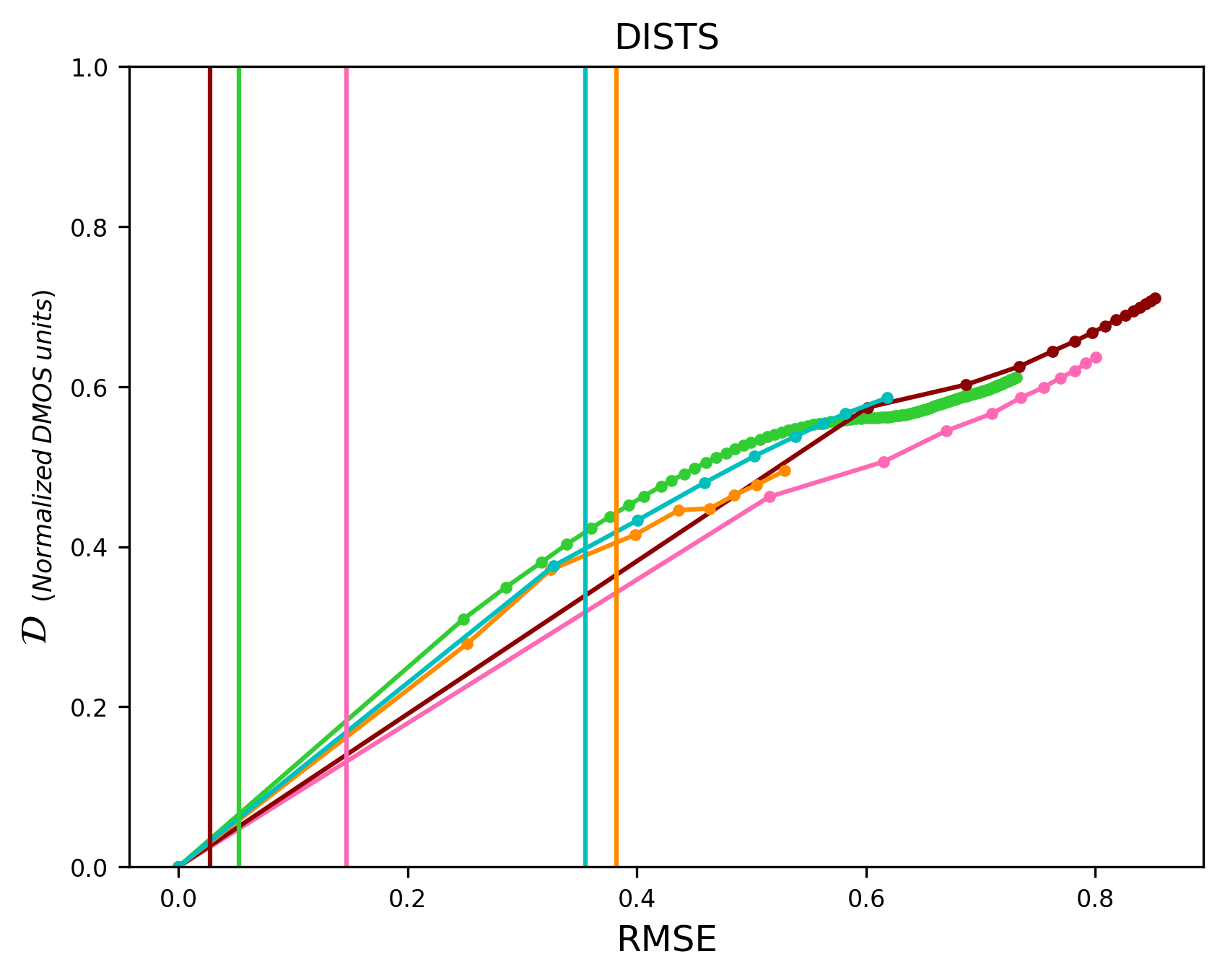}
		\end{subfigure}%
		\begin{subfigure}{0.33\textwidth}
			\centering
			\includegraphics[width=1\linewidth]{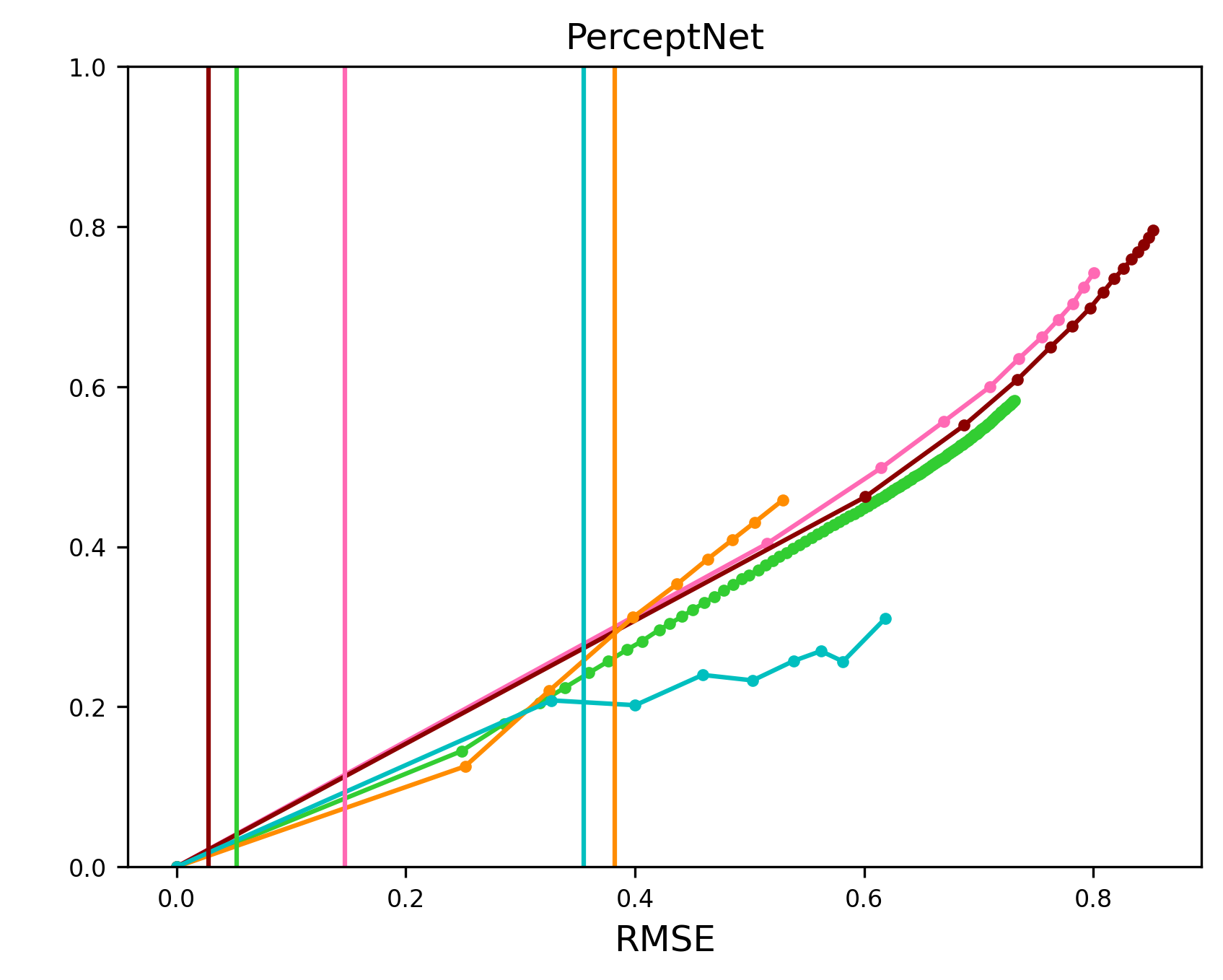}
		\end{subfigure}
            \begin{subfigure}{0.33\textwidth}
			\centering
			\includegraphics[width=1\linewidth]{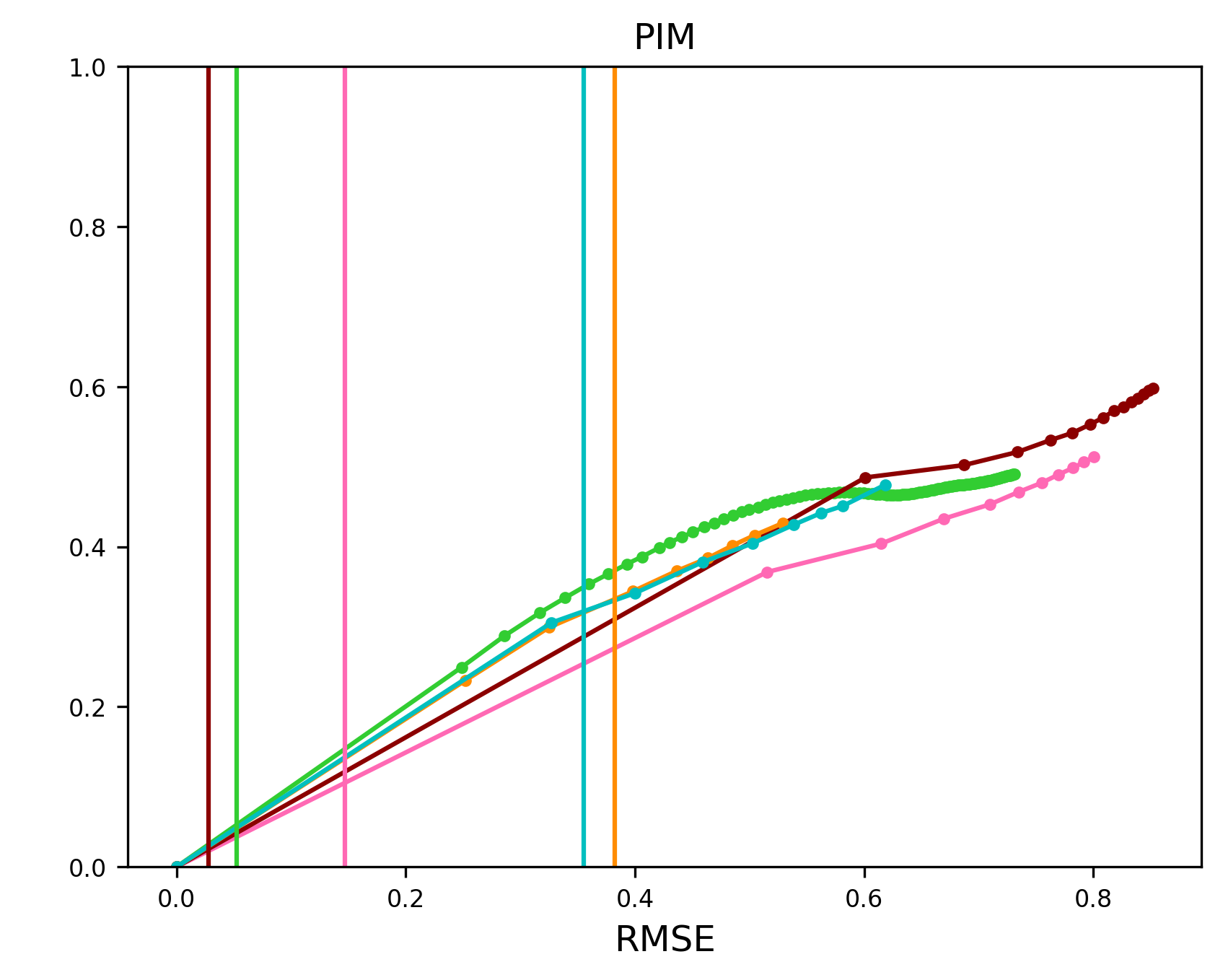}
		\end{subfigure}
            \label{fig:leaky}
		        \caption{Sensitivity test results for the CIFAR10 dataset. Table \ref{tab:resultados2} shows in detail the order of the sensitivities together with the human ordering, where it can be seen that no metric follows the order correctly.}
		\label{fig:results_ord_cifar}
	\end{figure*}

\begin{figure*}[!h]
		
		\begin{subfigure}{0.33\textwidth}
			\centering
			\includegraphics[width=1\linewidth]{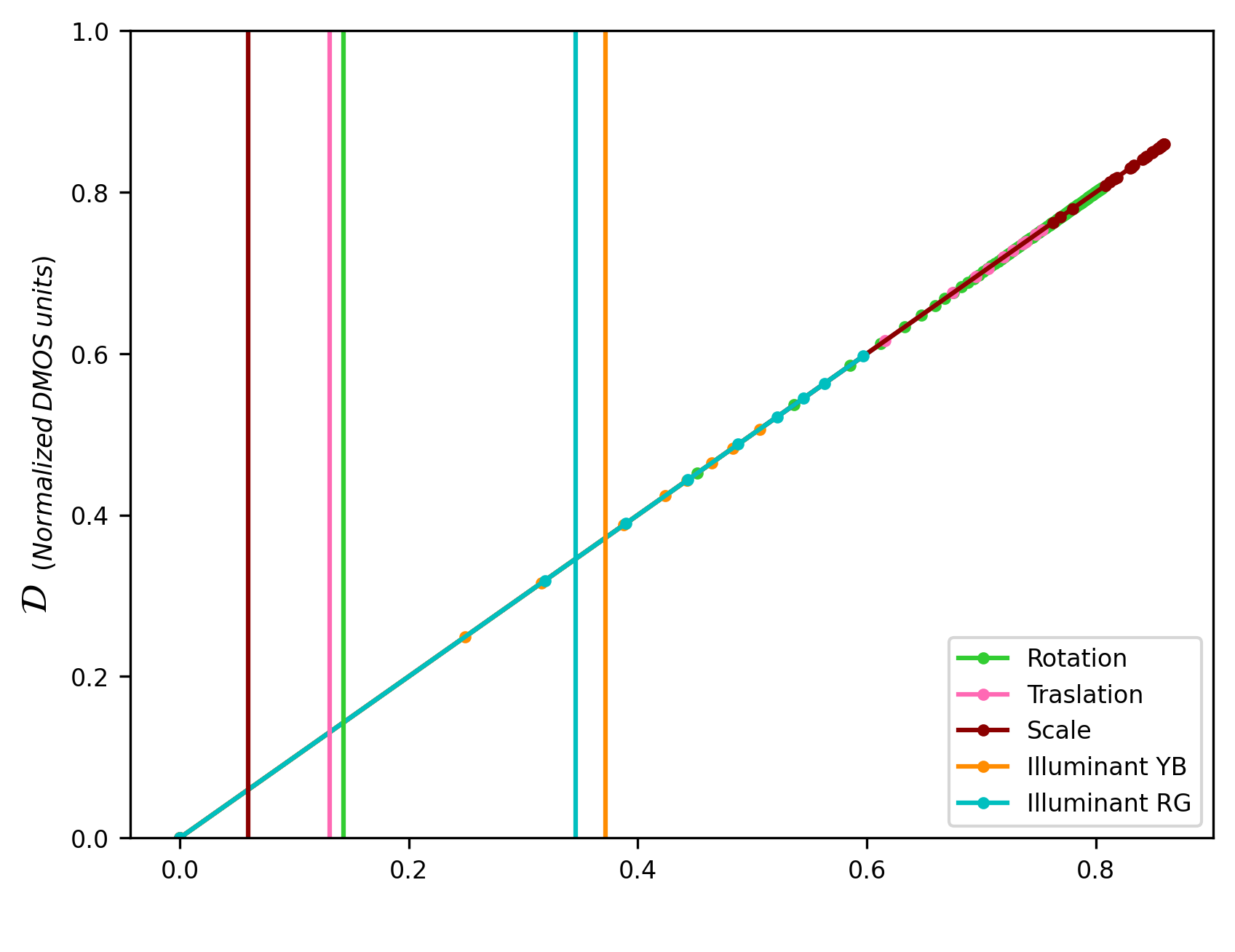}
		\end{subfigure}%
		\begin{subfigure}{0.33\textwidth}
			\centering
			\includegraphics[width=1\linewidth]{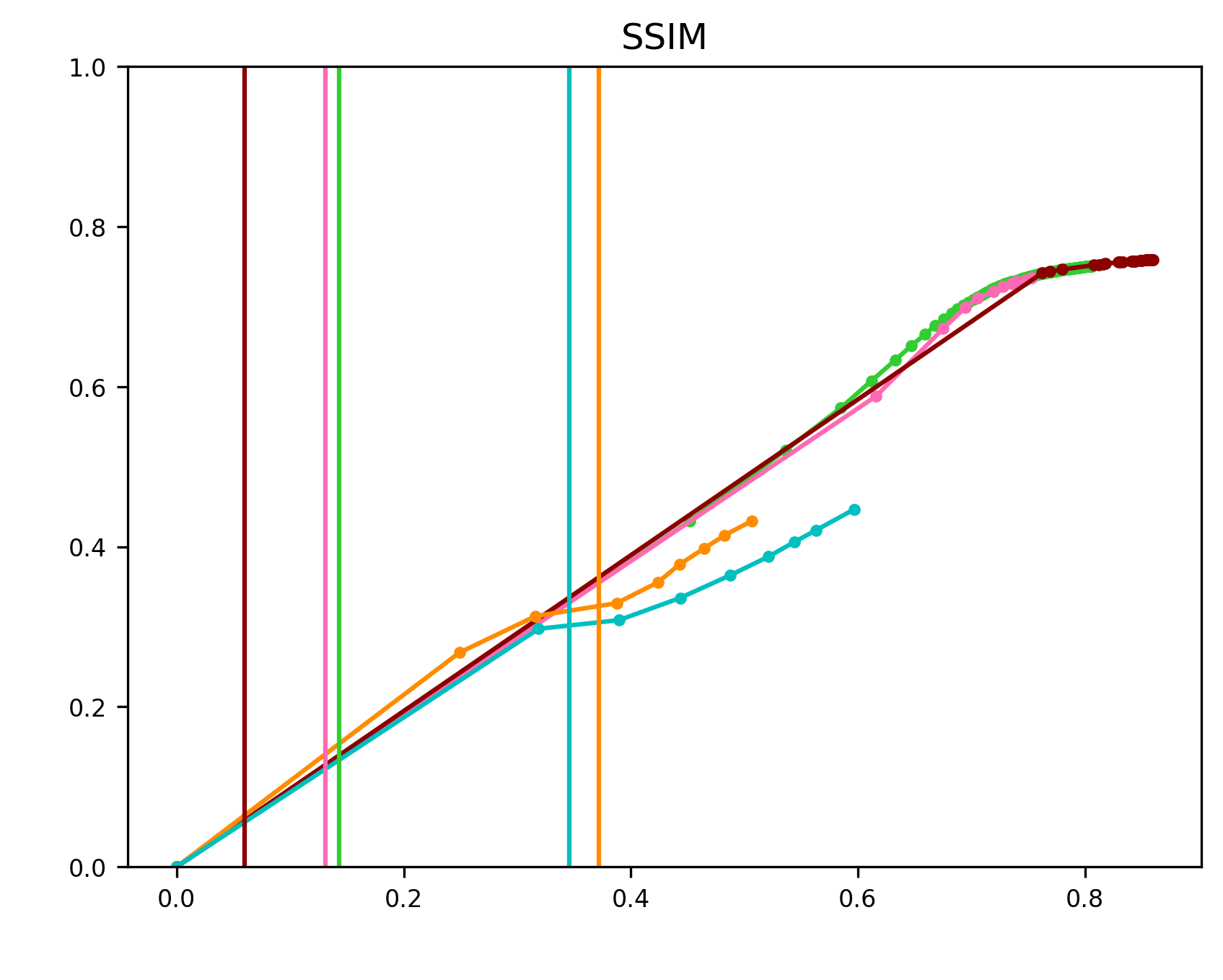}			
		\end{subfigure}
            \begin{subfigure}{0.33\textwidth}
			\centering
			\includegraphics[width=1\linewidth]{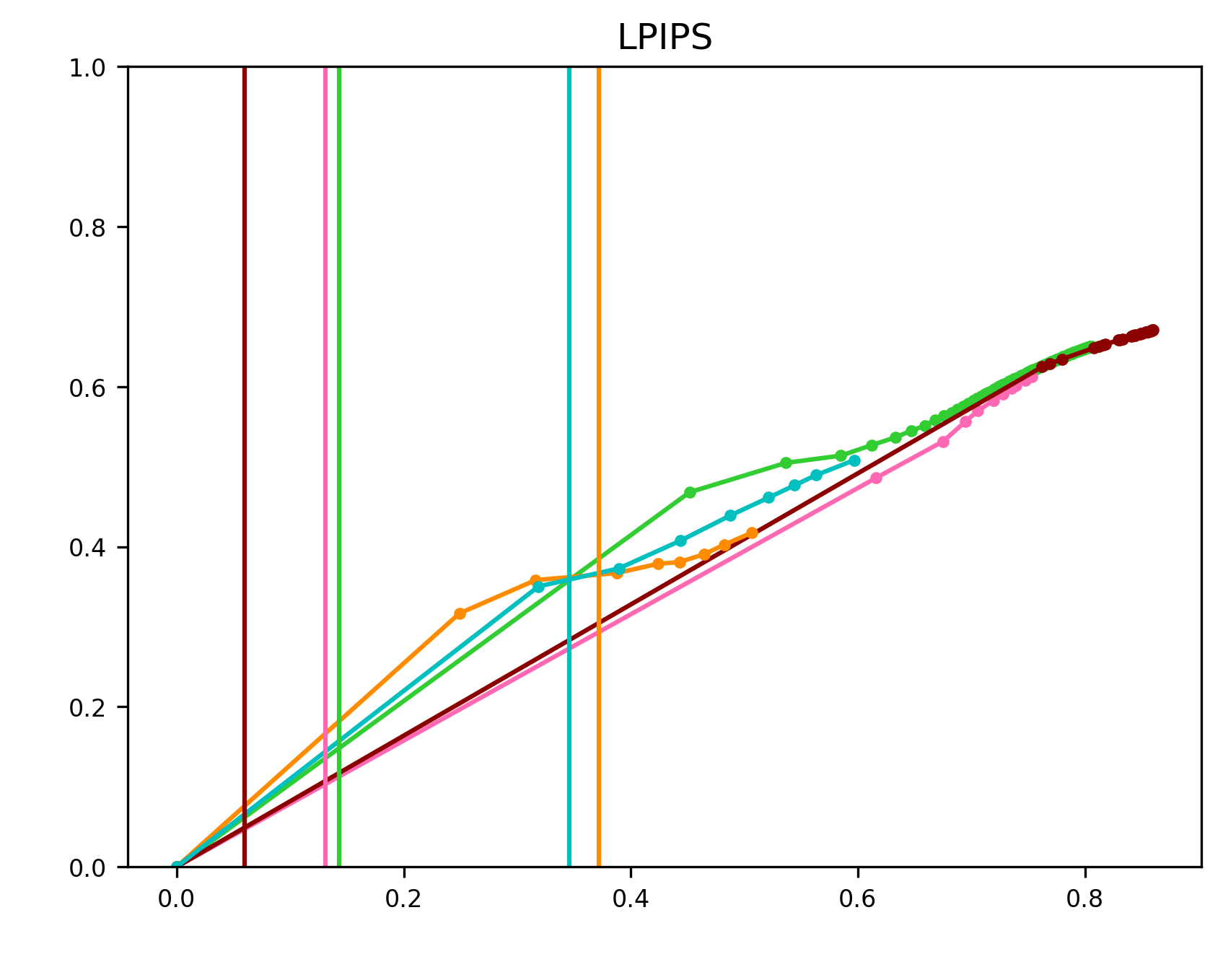}			
		\end{subfigure}
		\label{fig:results_rotation}

            \begin{subfigure}{0.33\textwidth}
			\centering
			\includegraphics[width=1\linewidth]{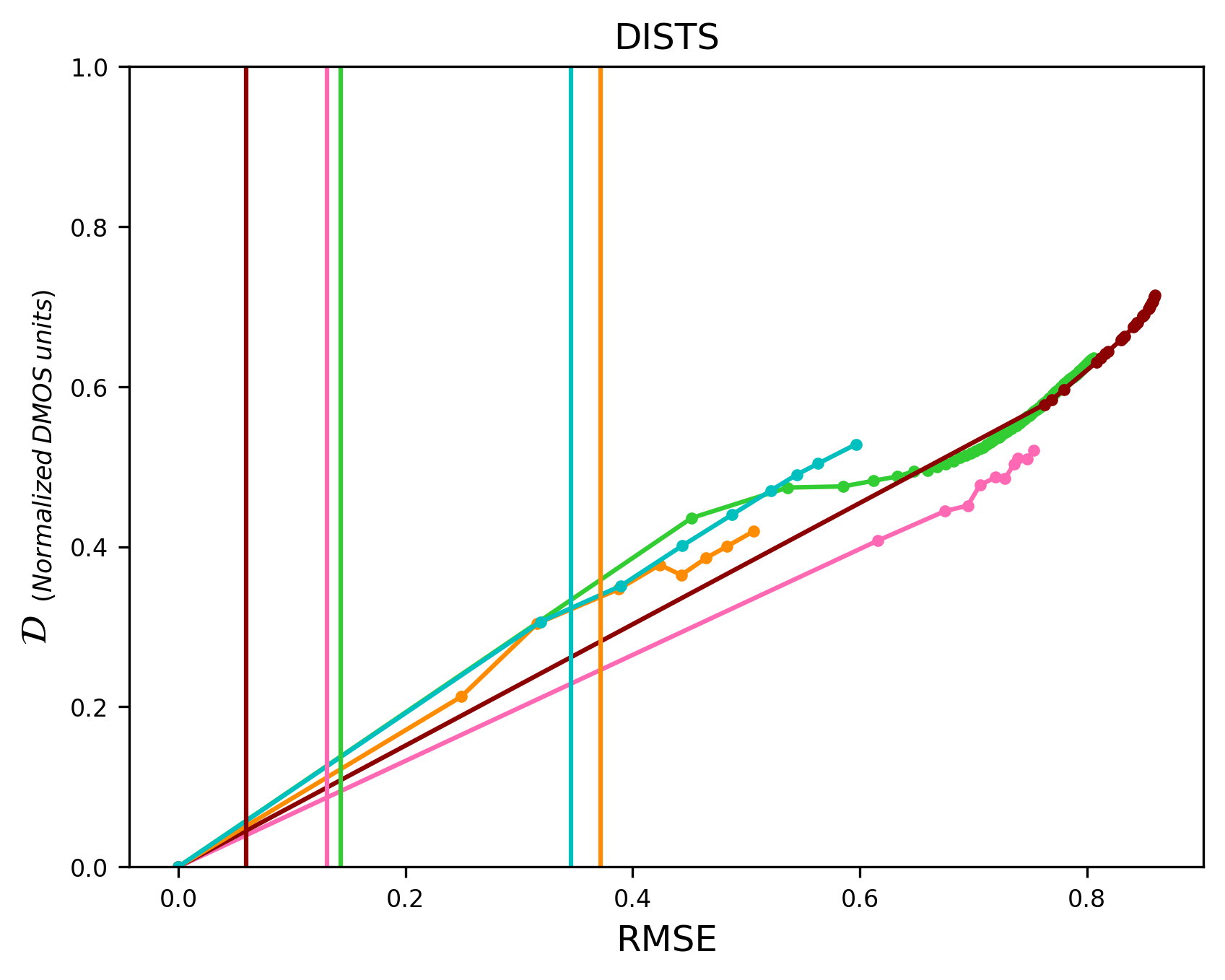}
		\end{subfigure}%
		\begin{subfigure}{0.33\textwidth}
			\centering
			\includegraphics[width=1\linewidth]{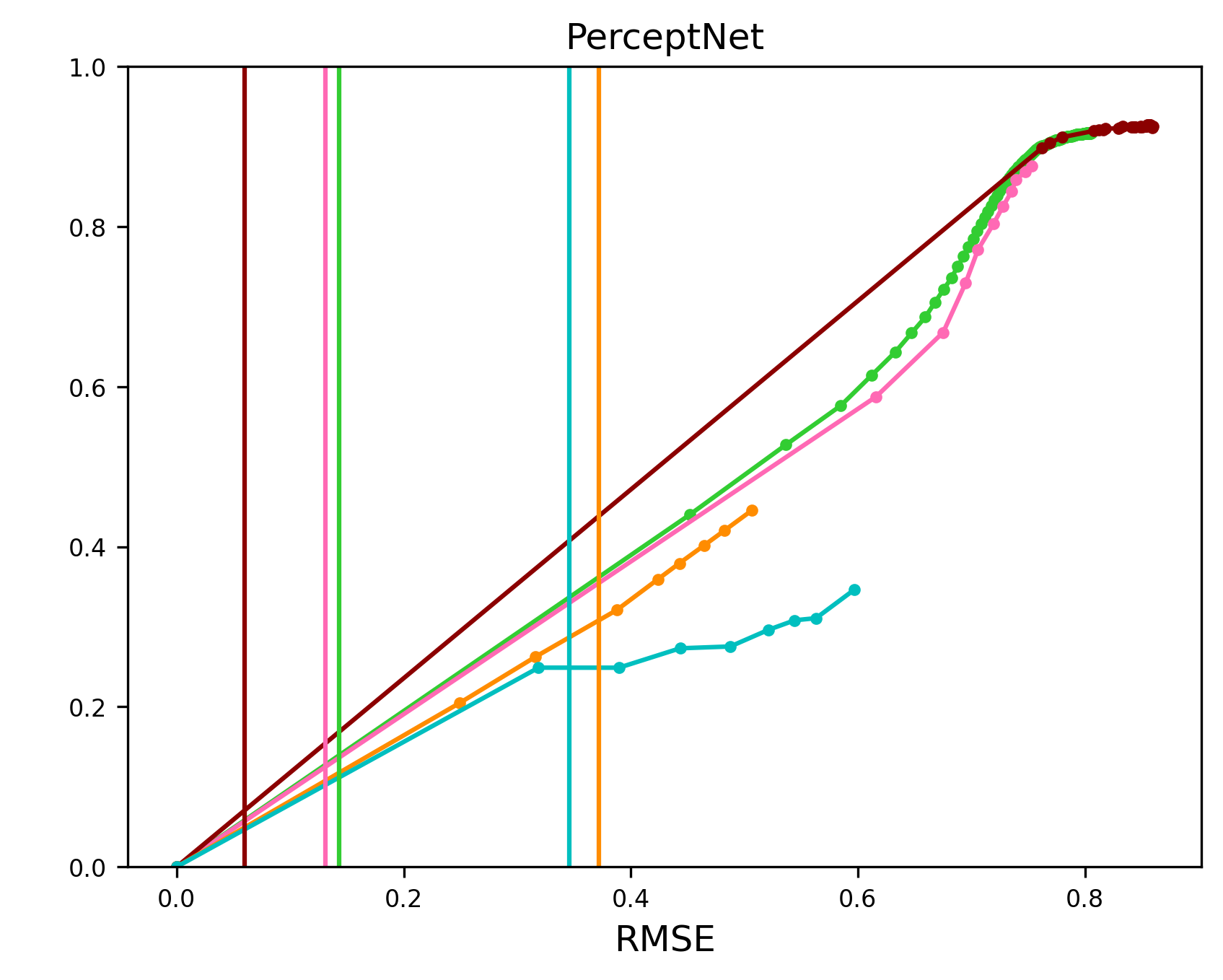}
		\end{subfigure}
            \begin{subfigure}{0.33\textwidth}
			\centering
			\includegraphics[width=1\linewidth]{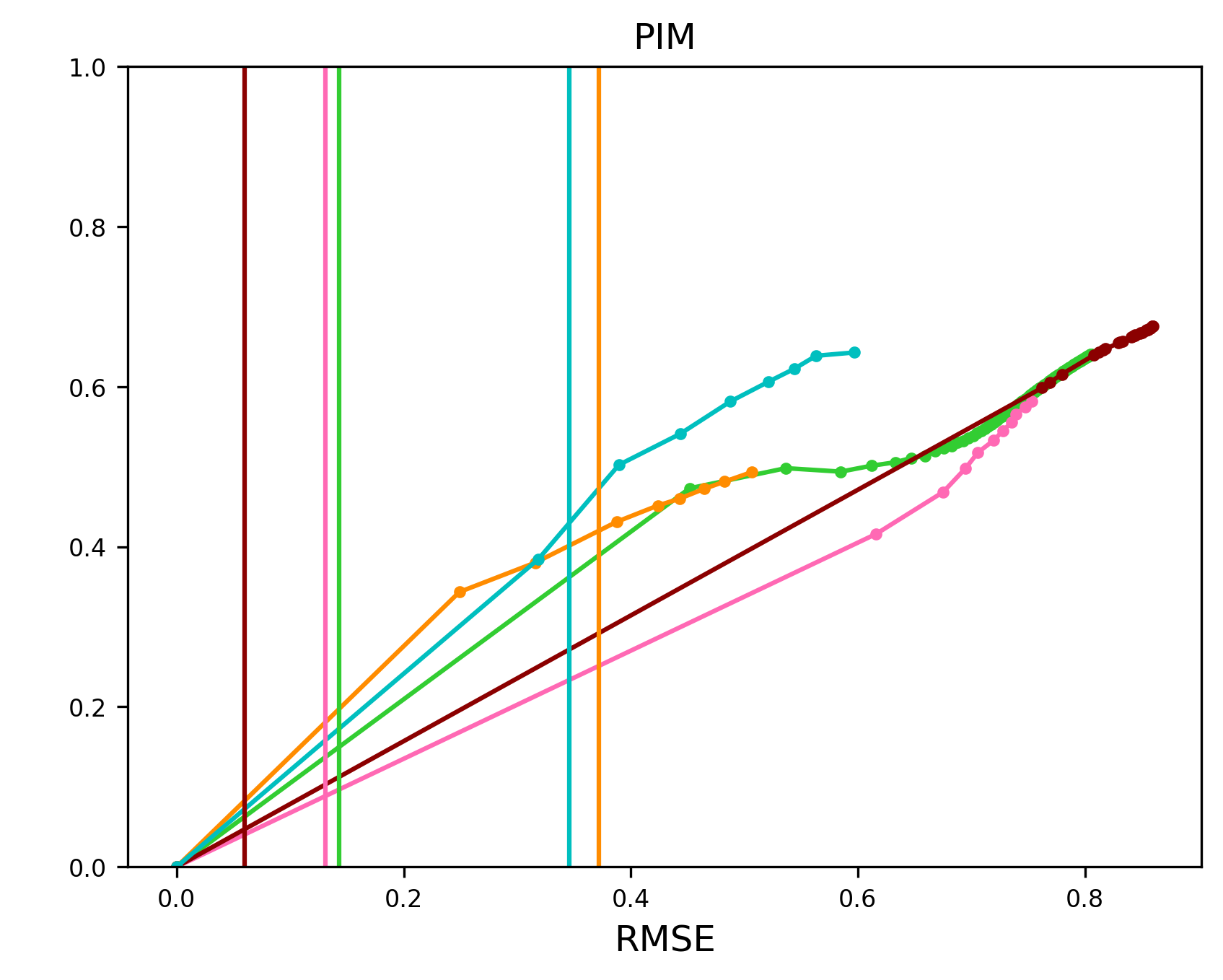}
		\end{subfigure}
            \label{fig:leaky}
		        \caption{Sensitivity test results for the TID13 dataset. Table \ref{tab:resultados2} shows in detail the order of the sensitivities together with the human ordering, where it can be seen that no metric follows the order correctly.}
		\label{fig:results_ord_tid}
	\end{figure*}

\begin{figure*}[!h]
		
		\begin{subfigure}{0.33\textwidth}
			\centering
			\includegraphics[width=1\linewidth]{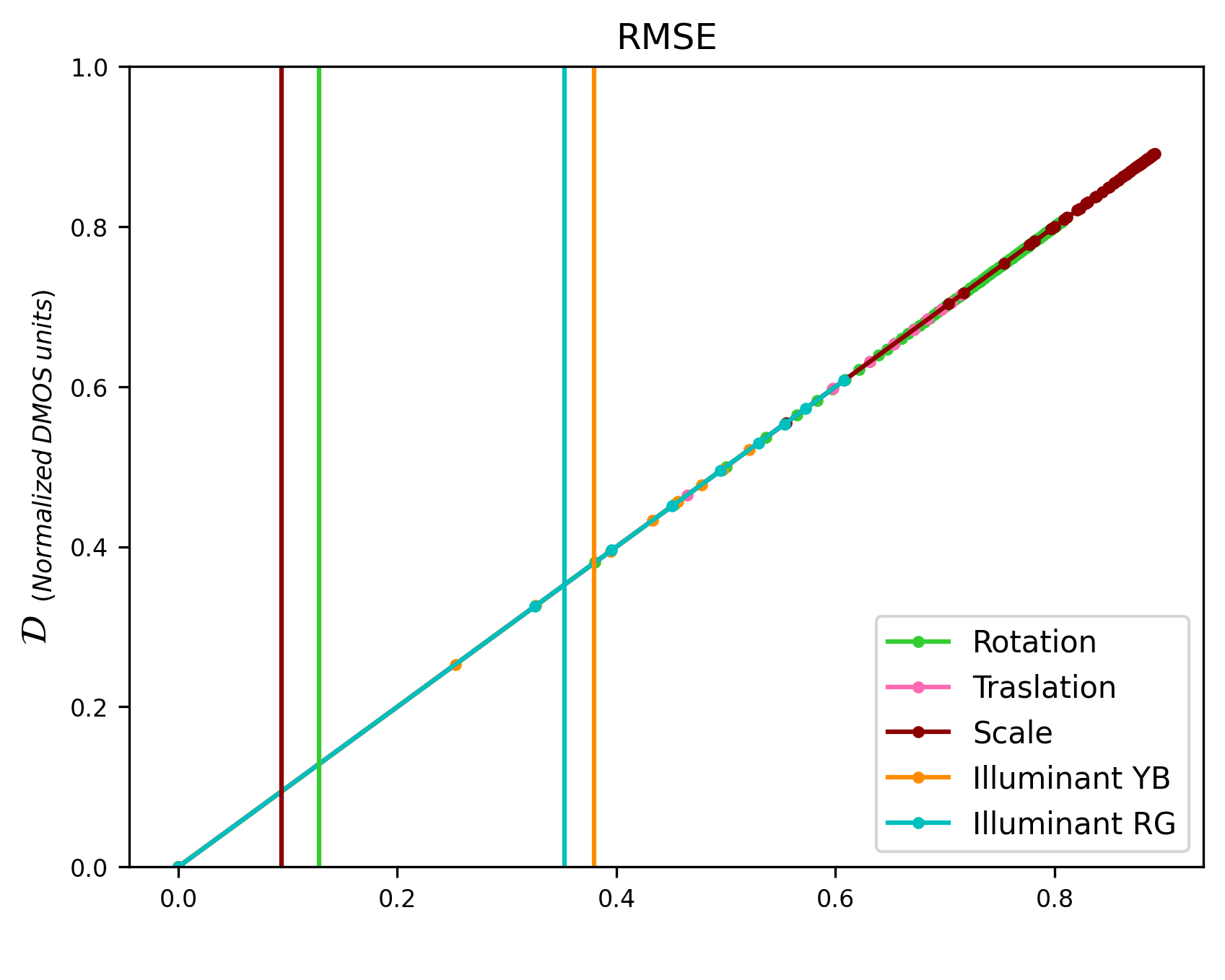}
		\end{subfigure}%
		\begin{subfigure}{0.33\textwidth}
			\centering
			\includegraphics[width=1\linewidth]{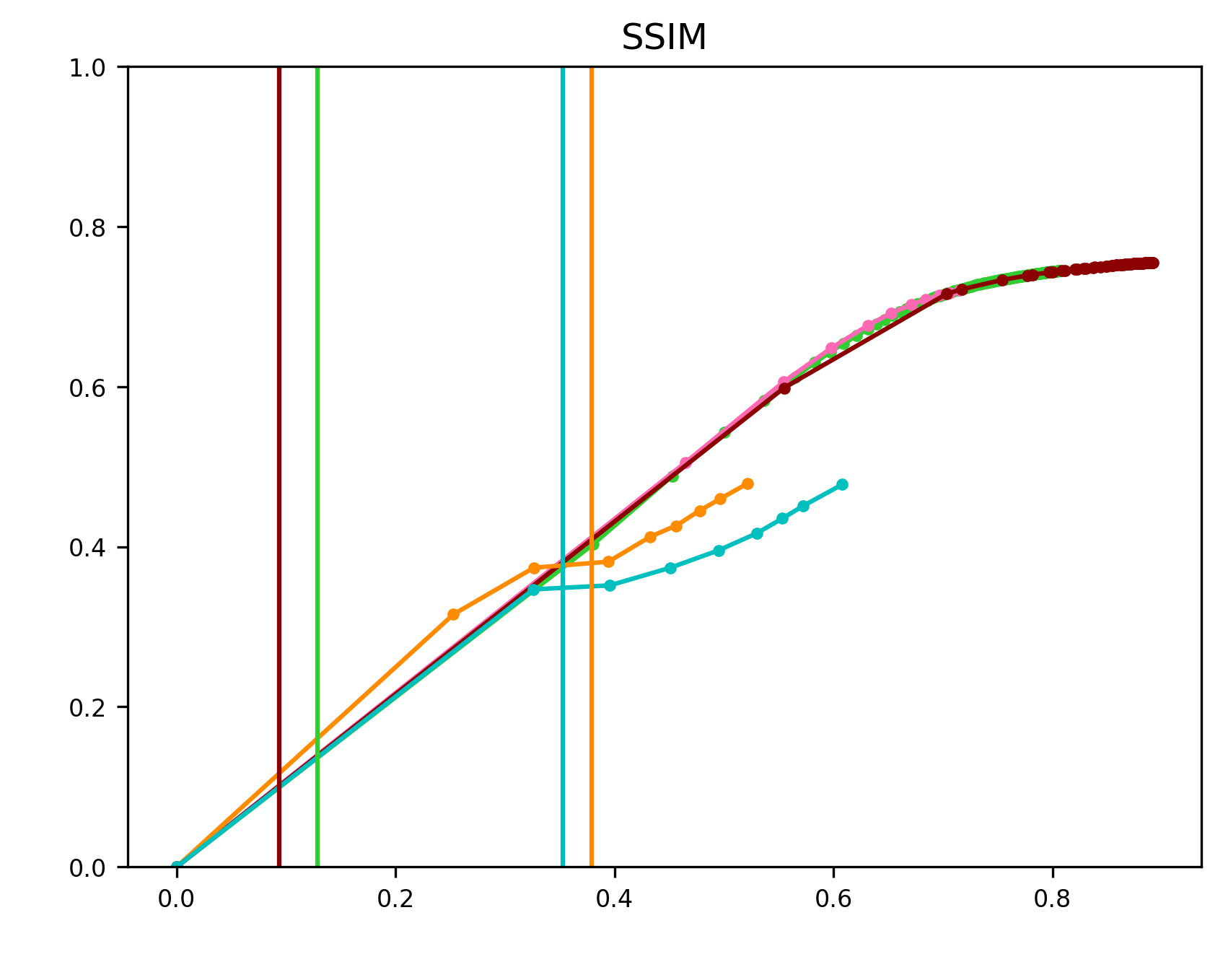}			
		\end{subfigure}
            \begin{subfigure}{0.33\textwidth}
			\centering
			\includegraphics[width=1\linewidth]{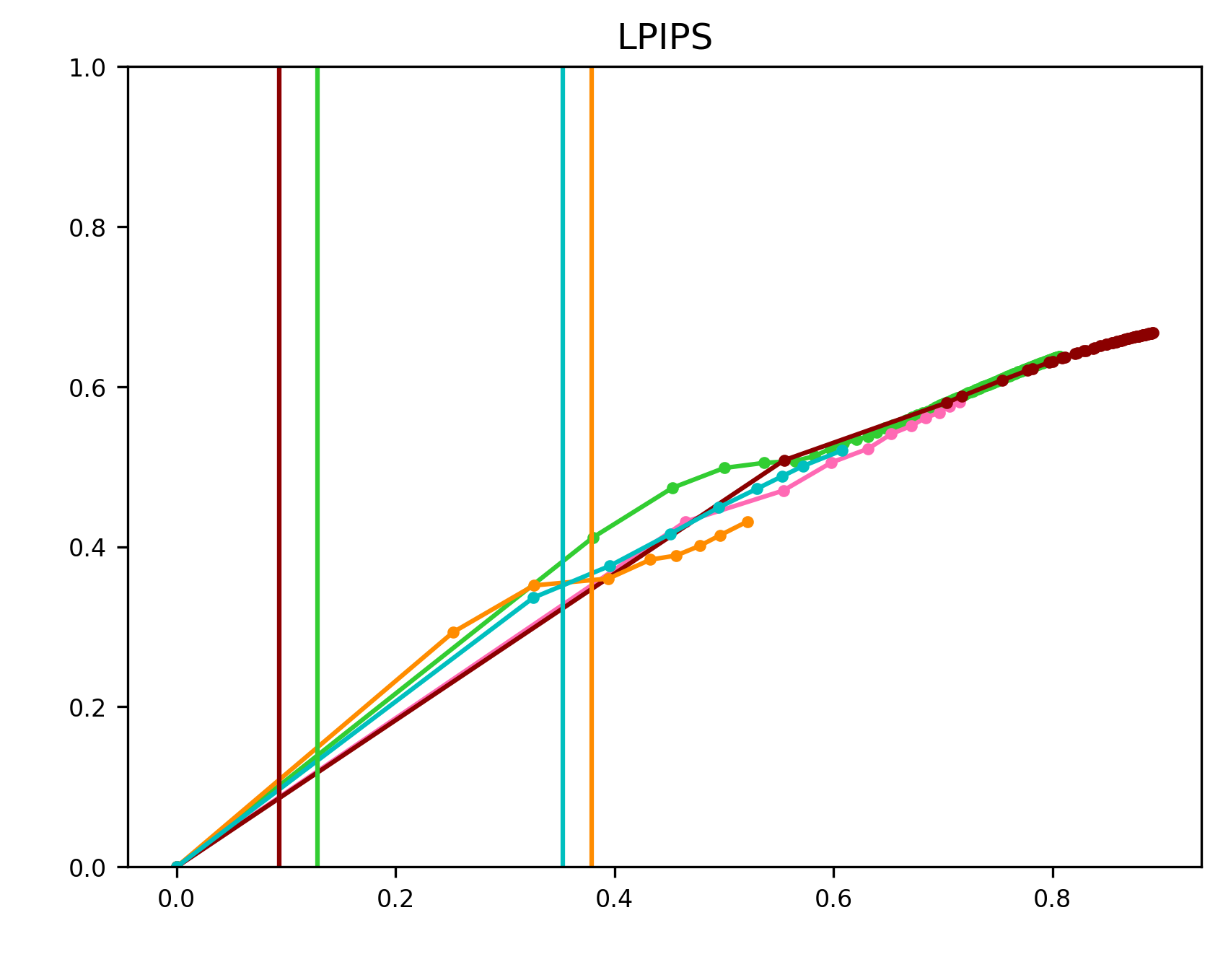}			
		\end{subfigure}
		\label{fig:results_rotation}

            \begin{subfigure}{0.33\textwidth}
			\centering
			\includegraphics[width=1\linewidth]{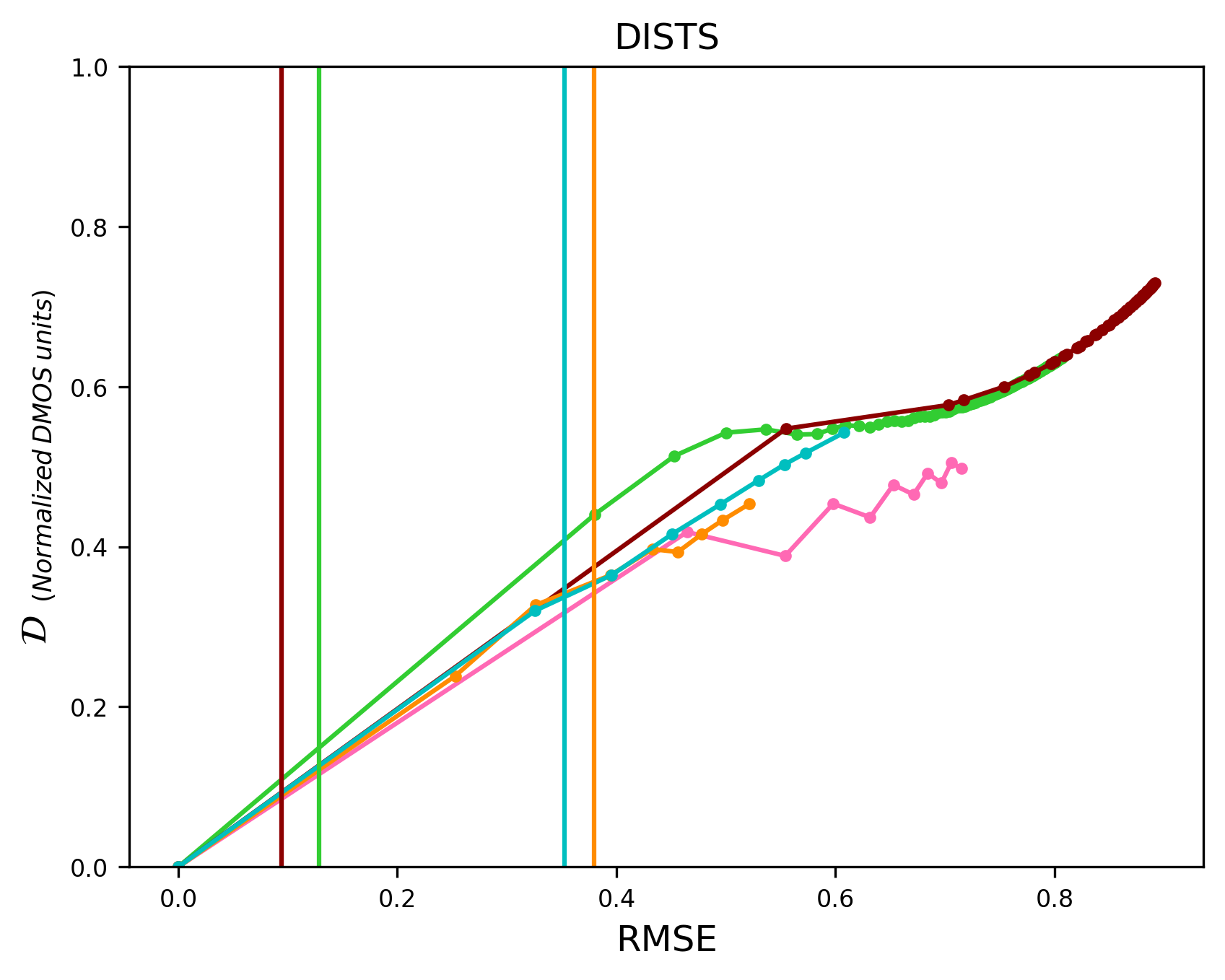}
		\end{subfigure}%
		\begin{subfigure}{0.33\textwidth}
			\centering
			\includegraphics[width=1\linewidth]{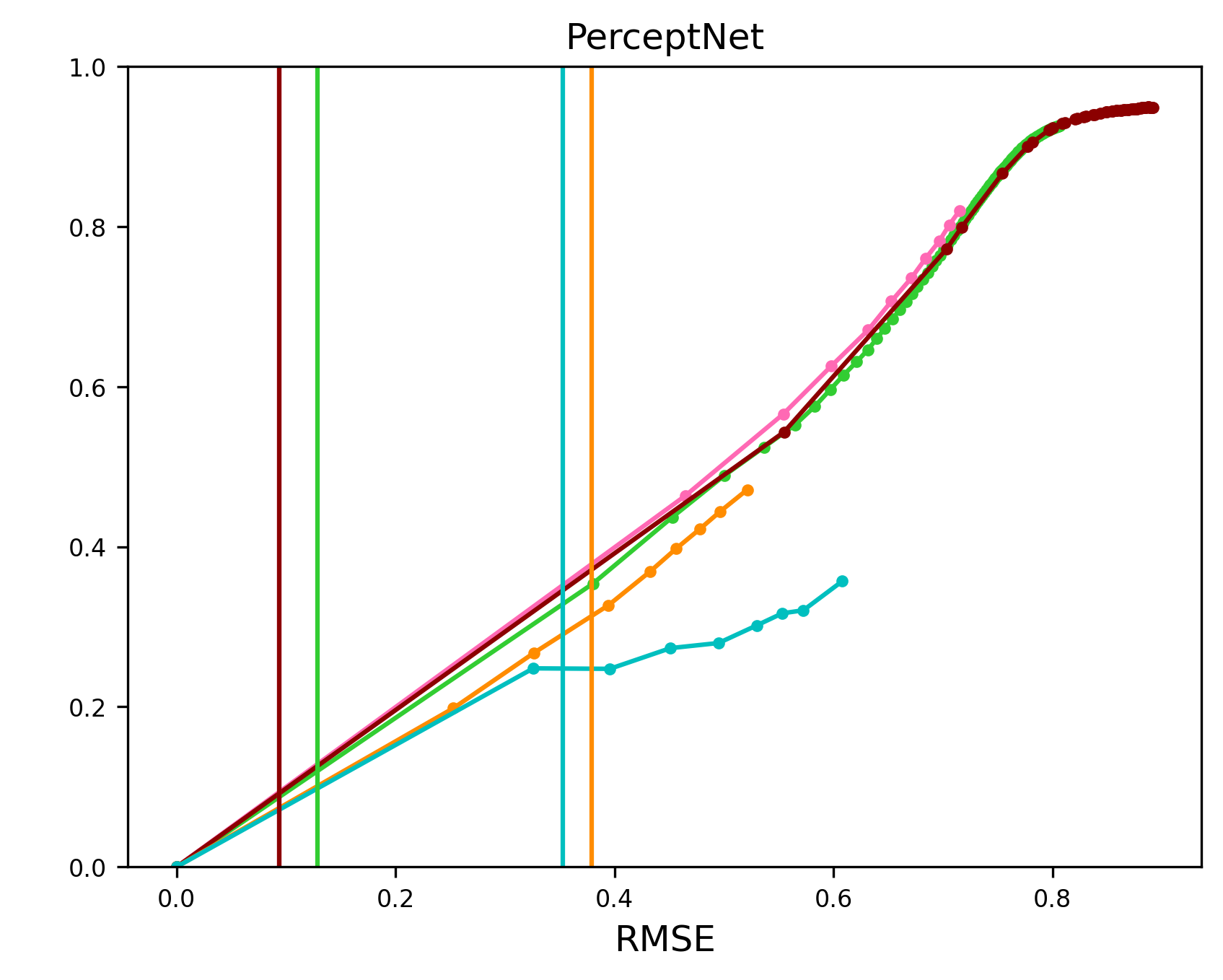}
		\end{subfigure}
            \begin{subfigure}{0.33\textwidth}
			\centering
			\includegraphics[width=1\linewidth]{Images/IMA_RMSE_PerceptNet.png}
		\end{subfigure}
            \label{fig:leaky}
		        \caption{Sensitivity test results for the ImageNet dataset. Table \ref{tab:resultados2} shows in detail the order of the sensitivities together with the human ordering, where it can be seen that no metric follows the order correctly.}
		\label{fig:results_ord_ima}
	\end{figure*}

Figures~\ref{fig:results_ord_mnist}, \ref{fig:results_ord_cifar},\ref{fig:results_ord_tid}, and \ref{fig:results_ord_ima} show the results for the all databases for the comparison of sensibilities. The vertical lines show the human thresholds for each transformation, the different curves, the behaviour of each model. The order of the vertical lines is compared with the value of the slope. A steep slope implies higher sensitivity.

\newpage

\printcredits





\end{document}